\documentclass{article}

\usepackage{microtype}
\usepackage{graphicx}
\usepackage{subcaption}
\usepackage{booktabs} 
\usepackage{amsmath} 
\usepackage{amssymb}  
\usepackage{pifont}   
\usepackage{xcolor} 
\usepackage{multirow, makecell, booktabs}
\usepackage[utf8]{inputenc}
\usepackage[T1]{fontenc}
\newcommand{\xmark}{\textcolor{red}{\ding{55}}}
\newcommand{\cmark}{\textcolor{green!70!black}{\checkmark}}

\usepackage{hyperref}
\usepackage{multirow}
\usepackage[table]{xcolor} 
\usepackage{gradient-text}

\usepackage[preprint]{icml2026}

\usepackage{amsmath}
\usepackage{amssymb}
\usepackage{mathtools}
\usepackage{amsthm}
\usepackage{booktabs} 
\usepackage{multirow} 
\usepackage{wrapfig}
\usepackage[title]{appendix}
\usepackage{titletoc}
\usepackage[capitalize,noabbrev]{cleveref}

\theoremstyle{plain}

\theoremstyle{definition}

\theoremstyle{remark}

\usepackage[textsize=tiny]{todonotes}

\icmltitlerunning{GameVerse: Can Vision-Language Models Learn from Video-based Reflection?}

\begin{document}

\twocolumn[
   \icmltitle{\textit{GameVerse}: {Can Vision-Language Models Learn from Video-based Reflection?}}




  \icmlsetsymbol{equal}{*}
  \icmlsetsymbol{cor}{$\dagger$}

  \begin{icmlauthorlist}
    \icmlauthor{Kuan Zhang}{equal,yyy}
    \icmlauthor{Dongchen Liu}{equal,yyy}
    \icmlauthor{Qiyue Zhao}{yyy}
    \icmlauthor{Jinkun Hou}{yyy}
    \\
    \icmlauthor{Xinran Zhang}{yyy}
    \icmlauthor{Qinlei Xie}{yyy}
    \icmlauthor{Miao Liu}{cor,yyy}
    \icmlauthor{Yiming Li}{cor,yyy}
    
  \end{icmlauthorlist}

  \icmlaffiliation{yyy}{College of AI, Tsinghua University, China}

  \icmlcorrespondingauthor{Yiming Li}{yimingli9702@gmail.com}
  \icmlcorrespondingauthor{Miao Liu}{lmaptx4869@gmail.com}

  \icmlkeywords{Machine Learning, ICML}
  \begin{center}
    {\small \url{https://gameverse-bench.github.io/}}
    \end{center}
  \vskip 0.3in
]



\printAffiliationsAndNotice{\icmlEqualContribution}

\begin{abstract}
Human gameplay is a visually grounded interaction loop in which players act, reflect on failures, and watch tutorials to refine strategies. Can Vision-Language Models (VLMs) also learn from video-based reflection?  We present \textbf{GameVerse}, a comprehensive video game benchmark that enables a \textit{reflective visual interaction loop}. Moving beyond traditional \textit{\textbf{fire-and-forget}} evaluations, it uses a novel \textit{\textbf{reflect-and-retry}} paradigm to assess how VLMs internalize visual experience and improve policies. To facilitate systematic and scalable evaluation, we also introduce a \textit{cognitive hierarchical taxonomy} spanning 15 globally popular games, \textit{dual action space} for both semantic and GUI control, and \textit{milestone evaluation} using advanced VLMs to quantify progress. Our experiments show that VLMs benefit from video-based reflection in varied settings, and perform best by combining failure trajectories and expert tutorials—a \textit{training-free} analogue to reinforcement learning (RL) plus supervised fine-tuning (SFT).

\vspace{-3mm}


\end{abstract}

\section{Introduction}

\begin{figure}[t]
    \centering
    \begin{subfigure}[b]{0.45\textwidth}
        \centering
        \includegraphics[width=\textwidth]{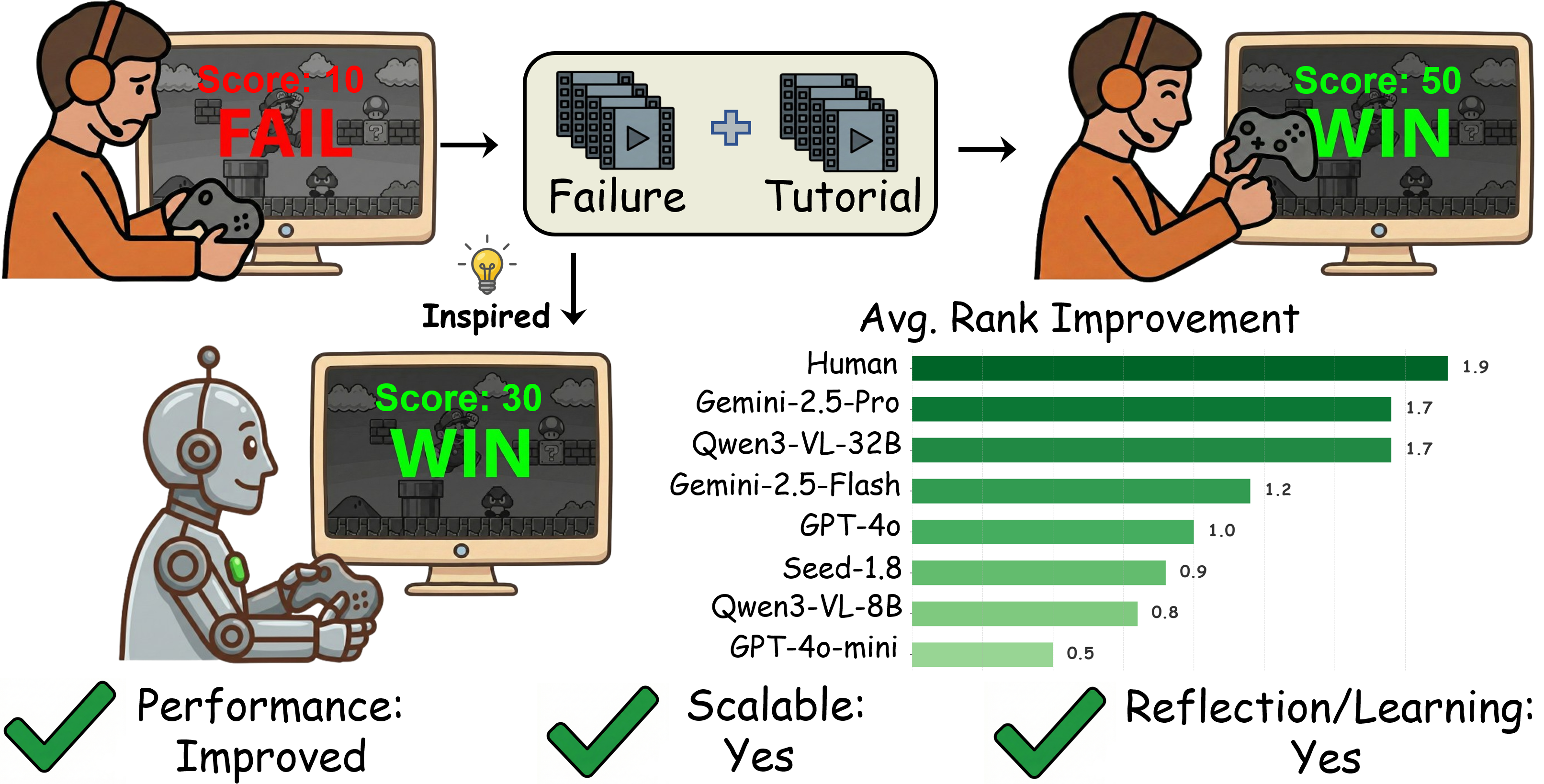}
    \end{subfigure}
    \caption{Humans improve gameplay by reflecting on failures and consulting expert tutorials. Our benchmark mimics this process, enabling agents to learn from video-based reflection. Humans show the largest gains, while all models benefit to varying degrees.}
    \vspace{-3mm}
    \label{png_motivation}
\end{figure}

Video games have been closely linked to artificial intelligence (AI) since its inception~\cite{turing54}. These virtual worlds provide diverse environments with varying dynamics and objectives. Humans adapt to these worlds via a visually grounded interaction loop (see~\cref{png_motivation}): they observe, act, reflect on failures, and consult external resources to refine their strategies. This ability to perceive, act, and learn from mistakes is fundamental to human intelligence, enabling continual adaptation through experience.

Motivated by the richness and diversity of video games, recent work has adopted them as embodied AI benchmarks for Vision-Language Models (VLMs). Unlike static datasets (e.g., VQA~\cite{VQA3}), which risk saturation, data quality~\cite{ido58} and data contamination~\cite{pollution12}, video games provide high quality, dynamic, visually rich, and long-horizon environments that demand a synthesis of perception, planning, reasoning, and motor control. \textit{However, building a comprehensive benchmark capturing human-like gameplay remains non-trivial.}


\begin{table*}[t]
\centering
\caption{\textbf{\textit{G}ame\textit{V}erse} (\textit{GV}) distinguishes itself through a \textbf{Cognitive Taxonomy} rather than commercial genres (\textit{Comm.Genre}), \textbf{Dual Action Space}, and a \textbf{Visually-Grounded Interaction Loop} allowing agents to reflect and retry. \textit{Mixed} features selecting games spanning various eras and genres. Tag $\dagger$ denotes using scaffolds/APIs to extract ground-truth state text, with vision as an auxiliary role.
}
\vspace{-1mm}
\label{tab:bench_comparison}
\resizebox{\textwidth}{!}{%
\begin{tabular}{lccccccccc}
\toprule
\textbf{Benchmarks} & \textbf{Game Env.} & \textbf{Taxonomy} & \textbf{\# Games} & \textbf{Diff. Tiers} & \textbf{Major Modality} & \textbf{Privilege-Free} & \textbf{Paradigm} & \textbf{Action} & \textbf{Scalability} \\ 
\midrule
GameArena~\cite{gamearena27} & 1D Text World & \xmark & 3 & \xmark & Text-Centric & \cmark & Fire-and-Forget & Semantic & \cmark \\
BALROG~\cite{balrog13} & 1D Text + 2D Grid & \xmark & 6 & \cmark & Text-Centric$\dagger$ & Scaffold & Fire-and-Forget & Semantic & \xmark \\
LVLM-Play~\cite{LVLM10} & 2D Grid & \xmark & 6 & \cmark & Vision-Centric & \cmark & Fire-and-Forget & Semantic & \cmark \\
\midrule
VideoGameBench~\cite{videogamebench4} & 2D Retro Arcade & \xmark & 10 & \xmark & Vision-Centric & \cmark & Fire-and-Forget & GUI & \cmark \\
FlashAdventure~\cite{flashadventure6} & 2D Web Flash & \xmark & \textbf{34} & \xmark & Vision-Centric & \cmark &Fire-and-Forget & GUI & \xmark\\
V-MAGE~\cite{vmage7} & 2D Retro Arcade & Comm. Genre & 5 & \cmark & Vision-Centric & \cmark & Fire-and-Forget & Semantic & \cmark \\
LMGAME-BENCH~\cite{lmgbench9} & 2D Mixed & Comm. Genre & 6 & \xmark & Text-Centric$\dagger$ & API + Scaffold & Fire-and-Forget & Semantic & \xmark \\
Orak~\cite{oark5} & 2D Mixed + 3D Mixed & Comm. Genre & 12 & \xmark &  Text-Centric$\dagger$ & API + Scaffold & Fire-and-Forget & Semantic & \xmark \\
\midrule
\rowcolor{gray!10} 
\textbf{GameVerse (Ours)} & \textbf{2D Mixed + 3D Mixed} & \textbf{Cognitive} & \textbf{15} & \cmark & \textbf{Vision-Centric} & \textbf{\cmark} & \textbf{Reflect-and-Retry} & \textbf{Dual} & \cmark \\
\bottomrule
\end{tabular}
}
\vspace{-3mm}
\end{table*}

In this work, we address this gap by establishing the desiderata for a comprehensive video game benchmark for VLM agents. \textbf{(1) Systematic Taxonomy.} Unlike the physical world, video games come in a wide variety of genres and settings. A fine-grained categorization is essential to rigorously evaluate agent performance across various scenarios. \textbf{(2) Vision-Centric Input.} Human gameplay relies solely on visual input. A good benchmark should therefore mimic this setup, allowing agents to learn from pixels without privileged information. \textbf{(3) Failure Reflection.} Most importantly, agents must be able to reflect on past failures and leverage external resources (e.g., tutorials) to improve.

Unfortunately, a comprehensive benchmark unifying all three aspects remains lacking (see~\cref{tab:bench_comparison}). Existing benchmarks often omit a systematic game taxonomy, e.g., relying on coarse classifications such as commercial genres (e.g., RPG, FPS)~\cite{vmage7,lmgbench9,oark5}. Moreover, many depend on scaffolding mechanisms that extract ground-truth state information as text~\cite{balrog13,oark5,lmgbench9}, rather than requiring agents to reason directly from pixels. This setup departs from human-like real-world interaction and limits scalability to closed-source games. Most importantly, all existing benchmarks employ a \textit{fire-and-forget} paradigm, measuring instantaneous performance without accounting for failure reflection and tutorial-driven improvement.

\begin{figure*}[t]
\begin{center}{
\includegraphics[width=0.926\textwidth]{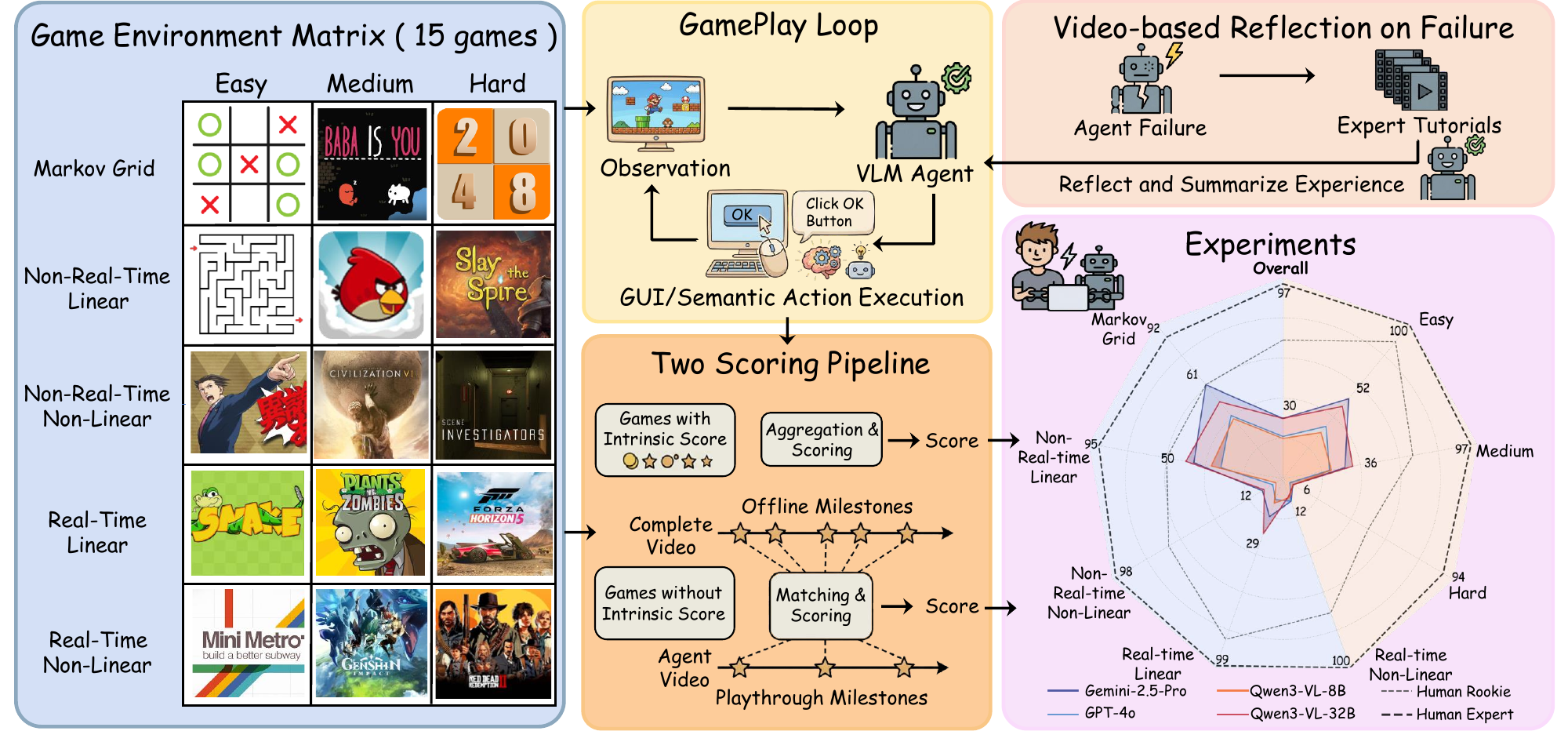}
}
\end{center}
\vspace{-1mm}
\caption{\textbf{Overview of GameVerse}, which effectively probes the capability boundaries of VLMs in diverse video game worlds. GameVerse supports dual action space, enables human-like reflection by integrating failure and tutorial videos, and delivers process score.}
\vspace{-2mm}
\label{Framework}
\end{figure*}

To this end, we develop \textbf{\textit{G}ame\textit{V}erse} (\textit{GV}), a comprehensive benchmark to evaluate VLM agents in a human-like manner. GameVerse distinguishes itself through a \textit{\textbf{diverse environment matrix}} spanning 15 globally popular games, \textit{\textbf{dual action space}} supporting both semantic and GUI control, and a \textit{\textbf{visually grounded interaction loop}} allowing failure reflection and tutorial learning (see \cref{tab:bench_comparison} and \cref{Framework}). Our major contributions are summarized as follows. 
\begin{itemize}
\vspace{-2mm}
    \item \textbf{Cognitive Hierarchical Taxonomy.} We build a brand-new fine-grained taxonomy based on three cognitive axes: \textit{image structure}, \textit{temporal dynamics}, and \textit{causal linearity}. By classifying games into five distinct categories and three difficulty tiers, we can precisely map the capability boundaries of different models.
\vspace{-1.5mm}
    \item \textbf{Video-Based Reflection.} We propose a novel \textit{reflect-and-retry} paradigm to enable visual experience internalization and policy refinement. Unlike conventional \textit{fire-and-forget} evaluation, we allow agents to diagnose their failure trajectories against expert video tutorials and leverage this feedback in subsequent attempts.
    \vspace{-1.5mm}
    \item \textbf{Scalable Evaluation Protocol.} We introduce a milestone scoring pipeline that leverages advanced VLMs to quantify agent progress purely from pixels, without relying on internal APIs or manual annotation. This enables scalable evaluation of closed-source games.
    \vspace{-2mm}
\end{itemize}
We conduct comprehensive experiments and reveal that while VLMs succeed in simple tasks, they struggle to generalize in complex games. Besides, reflection yields improvements, but VLMs are incapable of human-level adaptation from experience. Notably, the largest gains emerge from jointly leveraging failures and tutorials—a \textit{training-free paradigm} that combines \textit{in-context exploration} (failure-as-RL) with \textit{in-context imitation} (tutorial-as-SFT). This synergy outperforms either approach alone, mirroring the complementary benefits of reinforcement plus supervised learning in foundation model post-training~\cite{pmlr-v267-chu25c}.

\section{Related Work}

\textbf{Game Environments as AI Benchmarks.} Early game agents relied on reinforcement learning (RL) to solve simple environments \cite{atari23, gym22}. Recent attention has shifted toward tackling complex video games with general-purpose Vision-Language Models (VLMs). Balrog~\cite{balrog13} and LVLM~\cite{LVLM10} are primarily based on 2D-grid worlds. V-MAGE~\cite{vmage7}, VGB~\cite{videogamebench4} and GameStore~\cite{aigamescore57} assess VLMs with strict visual input in classic video games but suffer from poor performance and struggle to differentiate between models. Recently, Orak~\cite{oark5} and Lmgame-bench~\cite{lmgbench9} have focused on popular video games but rely heavily on internal game states via APIs for scaffolding; however, such privileged access is unrealistic for a human-like play-learning loop and does not scale to closed-source commercial titles.


\textbf{VLM Agents on Games.} Video game agents typically fall into modular or learning-based categories. Modular frameworks leverage frozen models with scaffolding: Voyager~\cite{voyager8} uses a code policy for exploration; Cradle~\cite{cradle14} employs a six-module cognitive architecture; FlashAdventure~\cite{flashadventure6} integrates clue-seeking for puzzles; and Gemini Plays Pokemon utilizes a memory module~\cite{geminipokemon15}. These systems exhibit the power of prompt engineering but often struggle with real-time precision. Learning-based approaches bridge the execution gap via training: CombatVLA \cite{combatvla16} and JARVIS-VLA \cite{jarvis17} improve performance via multi-stage training. UI-TARS, Game-TARS, and NitroGen \cite{UITARS118, uitars119, gametars20, nitrogen34} validate model capability to control games via atomic GUI actions, while Lumine~\cite{lumine21} achieves real-time interaction, showcasing live gameplay. While these agents perform well, they require extensive training on massive datasets of image-(language)-action pairs.


\textbf{Self-Reflection in Agents.} The ability to reflect on past actions is a hallmark of advanced intelligence. Early work in LLMs pioneered text-based self-reflection~\cite{Relexion47, Self-Refine48}. Reflection mechanisms evolved as agents moved into embodied environments: Voyager~\cite{voyager8} used environment feedback within a Minecraft prompt mechanism to refine its skill library; GROOT~\cite{groot51} employed videos as direct instructions for open-world tasks, training agents to guide their behavior; R3V~\cite{R3V49} reflects on Chain-of-Thought rationales to enhance multimodal reasoning; and ROE~\cite{ROE50} utilized predefined expert text experience, updated from past episodes in Starcraft II. Despite these advances, existing methods either learn offline from videos (MineDojo~\cite{minedojo52}, GROOT~\cite{groot51}) or reflect internally with complex workflows (R3V~\cite{R3V49}, ROE~\cite{ROE50}). They do not replicate the direct, purposeful human gameplay loop of watching a tutorial to diagnose a specific past failure.


\section{GameVerse}
In this section, we will detail the proposed hierarchical game taxonomy, the video-based reflection paradigm, and the novel evaluation metrics (see Figure \ref{Framework}).

\subsection{Cognitive Hierarchical Game Taxonomy}
\label{taxonomy}
Recent benchmarks~\cite{videogamebench4, lmgbench9, oark5} usually lack of a systematic taxonomy. They typically categorize games by commercial genres (e.g., RPG). While useful for consumers, they are too coarse for analyzing the bottleneck of VLM agents. For instance, a \textit{Strategy} game can range from a simple grid, turn-based task (e.g., Tic-Tac-Toe) to a complex, real-time, partially observable task (e.g., StarCraft), but they vary significantly in difficulty for the evaluation of VLM agents.

\begin{table*}[t]
    \centering
    \caption{Overview of the 15 games in GameVerse. We detail the number of trials ($N$), average time per round ($T$), supported evaluation modes (Semantic $A_S$ and GUI $A_G$), and a concise summary of the game content and the specific cognitive capabilities evaluated.}
    \label{tab:game_overview}
    \resizebox{1.00\textwidth}{!}{
    \begin{tabular}{l | c | c | >{\raggedright\arraybackslash}p{15cm}}
        \toprule
        \textbf{Game} & \textbf{Trials ($N$) / Time ($T$)} & \textbf{Eval. Mode} & \textbf{Feature \& Capabilities Evaluated} \\
        \midrule
        \rowcolor{gray!10} 
        \multicolumn{4}{l}{\textit{Markov Grid}} \\
        Tic-Tac-Toe & 10 / 40 sec & $A_S, A_G$ & Classic zero-sum grid game; tests \textbf{basic adversarial logical reasoning} and \textbf{turn-taking planning}. \\
        Baba Is You & 5 / 1.5 min & $A_S, A_G$ & Puzzle with manipulable rules; tests \textbf{out-of-the-box thinking} and \textbf{flexible logical inference}. \\
        2048 & 5 / 10 min & $A_S, A_G$ & Stochastic number puzzle; tests \textbf{probabilistic planning} and \textbf{spatial foresight} to manage grid constraints. \\
        \midrule
        \rowcolor{gray!10} 
        \multicolumn{4}{l}{\textit{Non-Real-Time Linear}} \\
        Maze & 10 / 45 sec & $A_S, A_G$ & Navigational puzzle; tests \textbf{virtual perception} and \textbf{spatial pathfinding} algorithms. \\
        Angry Birds & 5 / 2 min & $A_S, A_G$ & Physics-based puzzle; tests \textbf{spatial intelligence} and understanding of \textbf{causal physical interactions}. \\
        Slay the Spire & 3 / 20 min & $A_S, A_G$ & Roguelike deck-builder; tests long-term strategic \textbf{resource management} and \textbf{combinatorial optimization}. \\
        \midrule
        \rowcolor{gray!10} 
        \multicolumn{4}{l}{\textit{Non-Real-Time Non-Linear}} \\
        Ace Attorney & 5 / 15 min & $A_G$ & Narrative-driven adventure; tests high-level \textbf{reading comprehension} and \textbf{contradiction detection}. \\
        Civilization VI & 3 / 30 min & $A_G$ & Grand strategy; tests complex \textbf{multi-step planning}, \textbf{economic management}, and \textbf{long-horizon reasoning}. \\
        Scene Investigators & 3 / 50 min & $A_G$ & Deductive investigation; tests \textbf{visual observation}, \textbf{information synthesis}, and \textbf{causal inference}. \\
        \midrule
        \rowcolor{gray!10} 
        \multicolumn{4}{l}{\textit{Real-Time Linear}} \\
        Snake & 10 / 1 min & $A_S, A_G$ & Real-time arcade; tests \textbf{reflex-based control} and \textbf{spatial path planning} under growing constraints. \\
        Plants vs. Zombies & 5 / 3 min & $A_S, A_G$ & Tower defense; tests \textbf{real-time tactical deployment} and efficient \textbf{resource allocation}. \\
        Forza Horizon 5 & 3 / 3 min & $A_G$ & High-fidelity racing simulation; tests \textbf{fine-grained continuous control} and \textbf{visual-motor coordination}. \\
        \midrule
        \rowcolor{gray!10} 
        \multicolumn{4}{l}{\textit{Real-Time Non-Linear}} \\
        Mini Metro & 5 / 5 min & $A_G$ & Subway simulation; tests \textbf{graph optimization} and efficient \textbf{flow management} in evolving networks. \\
        Genshin Impact & 3 / 30 min & $A_G$ & Cartoon open-world RPG; tests \textbf{embodied exploration}, \textbf{3D navigation}, and \textbf{real-time combat mechanics}. \\
        Red Dead Redemption 2 & 3 / 2 hr & $A_G$ & Real open-world RPG; tests comprehensive \textbf{general intelligence}, \textbf{realistic physics}, and \textbf{social interaction}. \\
        \bottomrule
    \end{tabular}
    }
\end{table*}
Motivated by traditional game design \cite{game2def30, game1def29}, to evaluate the boundaries of VLM capabilities, we move beyond commercial genres taxonomy. We construct a hierarchical taxonomy based on three cognitive axes: \textit{Image Structure} (Grid/2D/3D), \textit{Temporal Dynamics} (Real-time/Non-Real-time) and \textit{Causal Linearity} (Linear/Non-linear), resulting in five distinct categories: \textit{\textbf{Markov Grid}}: Turn-based discrete state transitions with full observability. \textit{\textbf{Non-Real-time Linear}}: Turn-based progression with a fixed narrative path. \textit{\textbf{Non-Real-time Non-linear}}: Turn-based mechanics with open-ended goals. \textit{\textbf{Real-time Linear}}: Continuous time constraints with singular objectives. \textit{\textbf{Real-time Non-linear}}: Continuous dynamics with complex, branching objectives. For each category, we curate games across three difficulty tiers: \textit{easy, medium,} and \textit{hard} by factors like environmental complexity, action space, and etc (see Appendix~\ref{appa}). It probes the capability boundaries of VLM agents. In total, \textit{GV} selects 15 globally popular video games, distributed across these categories: (\textit{Tic-Tac-Toe, Baba Is You, 2048}), (\textit{Maze, Angry Birds, Slay the Spire}), (\textit{Ace Attorney, Civilization VI, Scene Investigator}), (\textit{Snake, Plants vs. Zombies, FORZA HORIZON 5}), and (\textit{Mini Metro, Genshin Impact, Red Dead Redemption 2}).

\textbf{Dual Action Space.} To assess both high-level reasoning and low-level control, we define two action mode. Semantic Actions $A_S$: High-level semantic actions (e.g., "Position(1,3)"), testing the agent's perception and reasoning. GUI Actions $A_G$: Low-level GUI operations (e.g., KeyPress(A)), testing end-to-end precise visual control.

\noindent \textbf{Game Description.} As shown in Table~\ref{tab:game_overview}, for each game, we provide a brief description of the game state, number of trials ($N$) and average estimated time for each person per round ($T$), and supported evaluation mode. Due to cost and time constraints, we limited the evaluation to $3\sim20$ trials, consistent with prior work, 1 in VideoGameBench~\cite{videogamebench4}, 3 in LMGAME-Bench~\cite{lmgbench9}, $3\sim20$ in Orak~\cite{oark5}. More descriptions of environments and tasks are elaborated in Appendix~\ref{appa} and~\ref{appd}.

\subsection{Video-based Reflection and Learning}
\label{vl}
Existing benchmarks predominantly use a \textit{fire-and-forget} workflow, maximizing immediate performance while ignoring reflection—a key dimension of human intelligence. We address this with a video-based reflection paradigm, enabling agents to refine gameplay by observing failures and tutorials.
The paradigm consists of four steps. \textbf{\textit{(1)~Trial and Failure.}} The agent first attempts the game task. If it fails, the system records the sequence of visual observations leading to the negative outcome. \textbf{\textit{(2)~Expert Demonstration Retrieval.}} The system retrieves an expert walkthrough from online gameplay videos. \textbf{\textit{(3)~Visual Reflection.}} The VLM acts as a reflector. It takes both its failure trajectory and the expert demonstration as visual input. The model is prompted to contrast the two, analyzing the divergence in strategy and execution, and generating condensed empirical reflections (e.g., "\textit{I failed because I targeted the wrong support beam; the expert targeted the central pillar}"). \textbf{\textit{(4)~Policy Update.}} These reflections are injected into the agent's system prompt, enabling it to re-attempt the task with the new context.

\subsection{Scoring Metrics}
\label{metrics}
We use a hybrid evaluation strategy tailored to each game’s feedback mechanism across the 15-game suite.

\textbf{Intrinsic Scoring.} For games that provide fine-grained, numerical feedback or clear competitive metrics (8 games, e.g., 2048 and etc.), we utilize the game's native scoring systems as the primary performance metrics.

\textbf{Milestone Scoring.} Evaluation in complex games (7 games, e.g., Genshin and etc.) is challenging due to sparse rewards and the absence of quantifiable scores. We propose a scalable pipeline that utilizes advanced VLMs (\textit{Gemini-3-pro} in this paper) to quantify progress purely from pixels.

\textbf{Phase 1: Offline Milestone Detection.} We employ an advanced VLM to watch expert walkthrough videos of the target task. The model extracts a structured JSON reference, denoted as $M_{ref} = \{M_1, M_2, ..., M_N\}$. Each milestone $M_i$ contains three attributes: (1) Timestamp: The time of occurrence in the video. (2) Title: A concise label for the event. (3) Description: A detailed description of the scene.

\textbf{Phase 2: Online Scoring.} Following gameplay, an advanced VLM analyzes the playthrough video, comparing the agent's trajectory against the reference milestones $M_{ref}$. By matching visual states to these milestones, the system calculates a process-oriented score $S=\frac{|M_{finished}|}{|M_{ref}|}\in[0,1]$. This method scales easily to closed-source commercial games where internal state information is inaccessible.

Notably, all milestones in GameVerse are manually verified to ensure accurate evaluation. We find them to be highly representative and suitable for direct use. Despite occasional hallucinations, they require minimal human verification (see Appendix~\ref{milestone:appendix}). This significantly reduces labor intensity compared to prior works \cite{wukong55, cradle14, flashadventure6} and ensures scalability.


\begin{table*}[t]
    \centering
    \caption{Performance of 7 VLMs on 15 GameVerse games in GUI mode, with and without Video-based Reflection (VR). \colorbox{green!30}{Green}/\colorbox{red!30}{Red} cells indicate performance gain/drop after applying VR. \textbf{Bold} represents the top performance across all models and human baselines.}
    \label{main-exp-merged}
    \newcommand{\std}[1]{{\scriptsize$\pm$#1}}
    
    \resizebox{1\textwidth}{!}{
    \begin{tabular}{ll | ccc | ccc | ccc | ccc | ccc | c }
        \toprule
        \multirow{2}{*}{\textbf{Model}} & \multirow{2}{*}{\textbf{VR}} & \multicolumn{3}{c|}{\textbf{Markov Grid}} & \multicolumn{3}{c|}{\textbf{Non-Real-time Linear}} & \multicolumn{3}{c|}{\textbf{Non-Real-time Non-Linear}} & \multicolumn{3}{c|}{\textbf{Real-time Linear}} & \multicolumn{3}{c|}{\textbf{Real-time Non-Linear}} & {\textbf{Avg}} \\
        \cmidrule(lr){3-11} \cmidrule(lr){12-17} 
        & & \textbf{TicTacToe} & \textbf{Baba} & \textbf{2048} & \textbf{Maze} & \textbf{AngryBird} & \textbf{Slay} & \textbf{Attorney} & \textbf{Civilization} & \textbf{Scene} & \textbf{Snake} & \textbf{PvZ} & \textbf{Horizon} & \textbf{Metro} & \textbf{Genshin} & \textbf{RDR 2} & \textbf{Rank}  \\
        \midrule
        \rowcolor{gray!15} \multicolumn{18}{c}{\textbf{\textit{Three Baselines}}} \\
        \midrule
        Random & --- & 26.3\std{19.3} & 10.8\std{7.6} & 1.2\std{0.9} & 21.3\std{2.7} & 0.0\std{0.0} & 0.0\std{0.0} & 0.0\std{0.0} & 0.0\std{0.0} & 0.0\std{0.0} & 7.0\std{9.1} & 0.0\std{0.0} & 0.0\std{0.0} & 0.0\std{0.0} & 0.0\std{0.0} & 0.0\std{0.0} & 16.5 \\
        \midrule
        Human Expert & --- & 98.9\std{2.6} & \textbf{100.0\std{0.0}} & \textbf{77.1\std{16.4}} & \textbf{100.0\std{0.0}} & \textbf{85.3\std{7.2}} & \textbf{99.4\std{1.2}} & \textbf{100.0\std{0.0}} & \textbf{100.0\std{0.0}} & \textbf{94.2\std{6.3}} & \textbf{100.0\std{0.0}} & \textbf{100.0\std{0.0}} & \textbf{98.1\std{4.4}} & \textbf{100.0\std{0.0}} & \textbf{100.0\std{0.0}} & \textbf{100.0\std{0.0}} & \textbf{1.6}\\
         \midrule
        \multirow{2}{*}{Human Rookie} & No & 85.1\std{19.3} & 83.4\std{15.2} & 15.9\std{4.1} & 99.3\std{1.2} & 46.4\std{10.3} & 34.2\std{7.1} & 77.4\std{27.2} & 54.1\std{17.2} & 70.4\std{5.2} & 93.2\std{10.4} & 89.3\std{10.8} & 73.9\std{9.1} & 91.2\std{15.4} & 63.3\std{18.2} & 58.4\std{21.3} & 4.1\\
        & Yes & \cellcolor{green!40}\textbf{99.4\std{1.9}} & \cellcolor{green!40}\textbf{100.0\std{0.0}} & \cellcolor{green!60}47.3\std{18.2} & \cellcolor{green!15}\textbf{100.0\std{0.0}} & \cellcolor{green!60}74.2\std{5.4} & \cellcolor{green!60}81.3\std{14.1} & \cellcolor{green!60}\textbf{100.0\std{0.0}} & \cellcolor{green!60}96.4\std{8.9} & \cellcolor{green!60}90.1\std{9.3} & \cellcolor{green!40}\textbf{100.0\std{0.0}} & \cellcolor{green!40}97.9\std{4.4} & \cellcolor{green!40}85.3\std{3.1} & \cellcolor{green!40}98.4\std{4.2} & \cellcolor{green!60}88.2\std{18.4} & \cellcolor{green!60}83.1\std{21.2} & \cellcolor{green!60}2.2 \\
        \midrule
        \rowcolor{gray!15} \multicolumn{18}{c}{\textbf{\textit{Seven Vision-Language Models}}} \\
        \midrule
        \multirow{2}{*}{Qwen3-VL-8B} & No & 53.3\std{15.4} & 60.0\std{0.0} & 4.4\std{2.1} & 80.3\std{45.4} & 19.6\std{8.3} & 11.7\std{0.0} & 7.1\std{6.2} & 0.0\std{0.0} & 6.3\std{4.4} & 2.2\std{6.8} & 33.4\std{4.1} & 3.2\std{0.0} & 7.8\std{5.2} & 14.3\std{0.0} & 7.2\std{5.4} & 12.4\\
        & Yes & \cellcolor{red!5}51.8\std{14.6} & \cellcolor{green!40}70.4\std{8.3} & \cellcolor{green!10}5.1\std{1.2} & \cellcolor{green!20}89.4\std{31.1} & \cellcolor{green!40}31.2\std{10.8} & \cellcolor{red!15}7.4\std{8.4} & \cellcolor{green!40}26.1\std{19.2} & 0.0\std{0.0} & 6.3\std{4.4} & \cellcolor{red!5}0.0\std{0.0} & \cellcolor{green!40}51.3\std{21.4} & 3.2\std{0.0} & \cellcolor{green!10}8.3\std{4.2} & 14.3\std{0.0} & 7.2\std{5.4} & \cellcolor{green!25}11.6\\
        \midrule
        \multirow{2}{*}{Qwen3-VL-32B} & No & 70.6\std{17.4} & 73.3\std{9.1} & 6.4\std{3.2} & \textbf{100.0\std{0.0}} & 41.6\std{7.4} & 8.2\std{4.1} & 17.3\std{6.3} & 8.3\std{0.0} & \textbf{8.3\std{0.0}} & 42.6\std{26.2} & 41.2\std{12.6} & 3.2\std{0.0} & 5.4\std{4.2} & 14.3\std{0.0} & 7.2\std{5.4} & 9.5\\
        & Yes & \cellcolor{red!15}63.1\std{7.4} & \cellcolor{green!40}\textbf{80.0\std{0.0}} & \cellcolor{red!5}6.1\std{3.7} & \textbf{100.0\std{0.0}} & \cellcolor{green!60}\textbf{59.4\std{10.9}} & \cellcolor{green!20}11.3\std{1.4} & \cellcolor{green!60}37.2\std{6.4} & 8.3\std{0.0} & \textbf{8.3\std{0.0}} & \cellcolor{red!15}33.4\std{16.9} & \cellcolor{green!40}48.3\std{19.0} & 3.2\std{0.0} & \cellcolor{green!40}11.2\std{6.3} & 14.3\std{0.0} & \cellcolor{green!20}10.0\std{0.0} & \cellcolor{green!60}7.8\\
        \midrule
        \multirow{2}{*}{GPT-4o-mini} & No & 34.4\std{17.2} & 60.0\std{0.0} & 1.3\std{0.4} & 77.1\std{43.4} & 16.7\std{11.3} & 2.0\std{0.0} & 0.0\std{0.0} & 0.0\std{0.0} & 3.2\std{4.1} & 0.0\std{0.0} & 27.6\std{14.0} & 3.2\std{0.0} & 4.2\std{3.4} & 14.3\std{0.0} & 10.0\std{0.0} & 14.1\\
        & Yes & \cellcolor{green!40}43.2\std{12.1} & \cellcolor{green!40}68.3\std{14.9} & \cellcolor{green!10}1.4\std{1.2} & \cellcolor{green!40}91.3\std{26.2} & \cellcolor{red!5}13.3\std{10.2} & 2.0\std{0.0} & 0.0\std{0.0} & 0.0\std{0.0} & 3.2\std{4.1} & 0.0\std{0.0} & \cellcolor{red!15}22.4\std{10.4} & 3.2\std{0.0} & \cellcolor{green!20}7.1\std{2.4} & 14.3\std{0.0} & 10.0\std{0.0} & \cellcolor{green!15}13.6\\
        \midrule
        \multirow{2}{*}{GPT-4o} & No & 58.6\std{4.2} & 60.9\std{1.4} & 3.8\std{0.9} & 80.2\std{44.1} & 14.3\std{9.0} & 2.0\std{0.0} & 18.2\std{6.1} & 0.0\std{0.0} & 0.0\std{0.0} & 0.0\std{0.0} & 32.2\std{12.4} & 2.4\std{2.2} & 11.2\std{2.1} & 14.3\std{0.0} & 10.0\std{0.0} & 13.3\\
        & Yes & \cellcolor{green!40}64.2\std{9.8} & \cellcolor{red!5}60.0\std{0.0} & \cellcolor{red!5}3.4\std{1.1} & \cellcolor{green!30}85.1\std{39.4} & \cellcolor{green!10}15.4\std{12.3} & 2.0\std{0.0} & \cellcolor{green!60}33.3\std{11.1} & 0.0\std{0.0} & 0.0\std{0.0} & 0.0\std{0.0} & \cellcolor{green!30}36.4\std{15.2} & \cellcolor{green!10}3.2\std{0.0} & \cellcolor{green!25}14.3\std{3.4} & 14.3\std{0.0} & 10.0\std{0.0} & \cellcolor{green!25}12.3\\
        \midrule
        \multirow{2}{*}{Seed-1.8} & No & 92.3\std{3.4} & \textbf{80.0\std{0.0}} & 9.1\std{5.2} & 87.8\std{20.9} & 29.4\std{7.1} & \textbf{39.6\std{21.2}} & 33.3\std{0.0} & \textbf{22.2\std{10.4}} & 3.2\std{4.1} & 7.2\std{10.1} & 26.1\std{8.2} & 5.4\std{2.1} & 1.3\std{2.4} & 14.3\std{0.0} & 10.0\std{0.0} & 9.3\\
        & Yes & \cellcolor{green!40}\textbf{100.0\std{0.0}} & \textbf{80.0\std{0.0}} & \cellcolor{green!60}18.2\std{8.4} & \cellcolor{green!40}\textbf{100.0\std{0.0}} & \cellcolor{green!60}57.2\std{12.7} & \cellcolor{red!15}28.4\std{8.1} & \cellcolor{green!60}\textbf{59.3\std{6.2}} & \cellcolor{red!5}19.5\std{7.8} & \cellcolor{green!25}\textbf{8.3\std{0.0}} & \cellcolor{green!25}10.4\std{10.8} & \cellcolor{red!5}22.1\std{10.3} & \cellcolor{red!5}2.4\std{2.2} & \cellcolor{green!10}1.4\std{1.2} & 14.3\std{0.0} & \cellcolor{red!5}7.2\std{5.4} & \cellcolor{green!45}8.4\\
        \midrule
        \multirow{2}{*}{Gemini-2.5-Flash} & No & 88.2\std{16.4} & \textbf{80.0\std{0.0}} & 7.8\std{2.1} & \textbf{100.0\std{0.0}} & 9.4\std{8.2} & 2.0\std{0.0} & 33.3\std{11.1} & 0.0\std{0.0} & 3.2\std{4.1} & 20.4\std{17.9} & 51.1\std{13.2} & 5.4\std{2.1} & 14.6\std{2.7} & 14.3\std{0.0} & 10.0\std{0.0} & 9.3\\
        & Yes & \cellcolor{green!40}95.1\std{12.4} & \textbf{80.0\std{0.0}} & \cellcolor{green!35}11.4\std{3.8} & \textbf{100.0\std{0.0}} & \cellcolor{green!10}10.2\std{6.4} & 2.0\std{0.0} & \cellcolor{red!5}29.3\std{25.1} & 0.0\std{0.0} & \cellcolor{green!20}\textbf{8.3\std{0.0}} & \cellcolor{red!15}10.4\std{8.1} & \cellcolor{green!25}\textbf{55.3\std{10.4}} & 5.4\std{2.1} & \cellcolor{green!25}\textbf{19.1\std{9.6}} & 14.3\std{0.0} & \cellcolor{green!25}\textbf{13.4\std{4.6}} & \cellcolor{green!45}8.1\\
        \midrule
        \multirow{2}{*}{Gemini-2.5-Pro} & No & 90.0\std{9.0} & \textbf{80.0\std{0.0}} & 13.2\std{3.1} & \textbf{100.0\std{0.0}} & 32.7\std{12.8} & 4.4\std{3.4} & 37.2\std{6.4} & 0.0\std{0.0} & 0.0\std{0.0} & 24.2\std{30.1} & 33.4\std{17.2} & 3.2\std{0.0} & 11.3\std{3.1} & 14.3\std{0.0} & 10.0\std{0.0} & 9.2\\
        & Yes & \cellcolor{green!40}\textbf{100.0\std{0.0}} & \textbf{80.0\std{0.0}} & \cellcolor{green!60}\textbf{26.4\std{6.8}} & \textbf{100.0\std{0.0}} & \cellcolor{green!45}42.2\std{11.3} & \cellcolor{green!25}7.9\std{0.0} & \cellcolor{green!40}48.3\std{6.1} & 0.0\std{0.0} & 0.0\std{0.0} & \cellcolor{green!60}\textbf{68.4\std{32.2}} & \cellcolor{red!5}30.3\std{16.4} & \cellcolor{green!25}\textbf{5.4\std{2.1}} & \cellcolor{green!40}18.2\std{8.1} & 14.3\std{0.0} & 10.0\std{0.0} & \cellcolor{green!60}7.5\\
        \bottomrule
    \end{tabular}
    }
\end{table*}

\section{Experiment}
\subsection{Experimental Setup}
\label{sec:exp_setup}

\textbf{Models.} We evaluate the performance of 7 VLMs to assess the current landscape of capabilities. The models include 2 open-source VLMs: Qwen3-VL-8B/32B \cite{Qwen3vl24}, and 5 proprietary VLMs: GPT-4o/4o-mini \cite{GPT4o25}, Seed-1.8 \cite{Seed1.8-33} and Gemini-2.5-Pro/Flash \cite{gemini2.526}.

\textbf{Setup.} We conduct experiments in two modes to verify model's \textit{reflect-and-retry} ability in \ref{vl}. W/o video-based reflection ($VR=No$): The agent plays the game directly. W/ video-based reflection ($VR=Yes$): The agent utilizes failure and expert videos to refine its policy in next attempt. We retrieve the top-ranked videos on public platforms. Prompts, FPS, and video details are in Appendix~\ref{reflection:appendix}.

\textbf{Baselines.} To contextualize model performance, we compare them against three baselines. \textbf{\textit{(1) Random.}} Selects valid actions uniformly at random. \textbf{\textit{(2) Human Rookie.}} Users who are computer literate but unfamiliar with the target games. \textbf{\textit{(3) Human Expert.}} Users who are computer literate and possess significant prior experience with the target games. We recruited 37 participants, conducting a total of 458 gameplay sessions. Further details are in Appendix~\ref{appc}.

\textbf{Evaluation Metrics.} For each game, we report the average normalized scores on \textbf{a 100-point scale} with standard deviation. All performance metrics were manually verified. For 10 games, we use a zero-shot agent, where only the current screenshot and the vision-action pair from the last step are provided as context. For 5 games, we employ a memory-aware agent, which utilizes a memory module to store long-term and short-term game history. The comparison between zero-shot and memory agents is detailed in Appendix~\ref{appd}. More detailed configurations are in Appendix~\ref{appb}.

\subsection{Performance Overview}
\label{performance}

Table~\ref{main-exp-merged} presents the performance of 7 VLMs across different game categories in GameVerse. Specifically, \textbf{Gemini-2.5-Pro} achieves the highest average ranking (7.5/9.2) among all VLMs. It is closely followed by \textbf{Gemini-2.5-Flash} (8.1/9.3), \textbf{Seed-1.8} (8.4/9.3) and the open-weights model \textbf{Qwen3-VL-32B} (7.8/9.5), establishing a clear performance hierarchy.

\textbf{Surpassing the Random Baseline.} Our results establish that current VLMs have acquired basic gameplay capabilities under the vision-centric condition. Across almost all games, models consistently outperform the random baseline, indicating that they are not merely acting stochastically but engaging in meaningful visual reasoning and execution.

\begin{figure}[t]
\begin{center}{
\includegraphics[width=1\columnwidth]{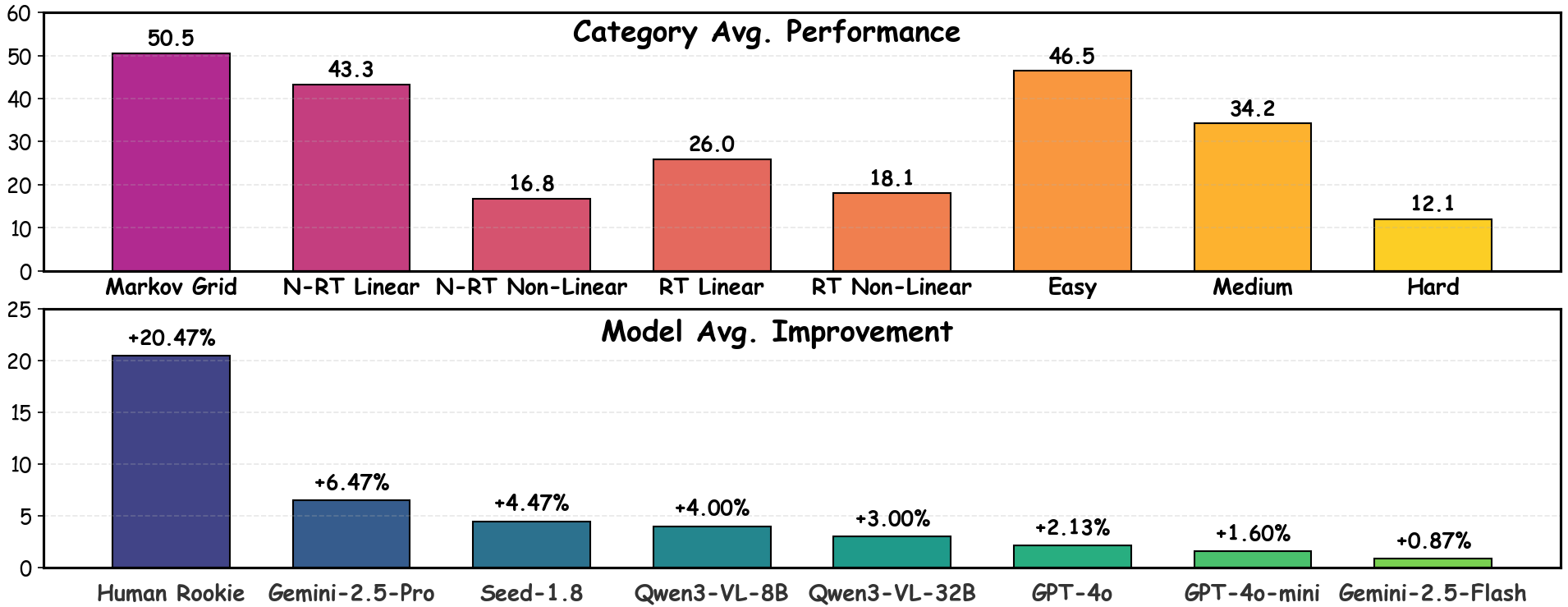}
}
\end{center}
\vspace{-1mm}
\caption{\textbf{Top:} Average performance across 5 cognitive and 3 difficulty categories. \textbf{Bottom:} Average improvement across models.}
\vspace{-1mm}
\label{png:vlstatusa}
\end{figure}

\begin{figure}[t]
\begin{center}{
\includegraphics[width=0.9\columnwidth]{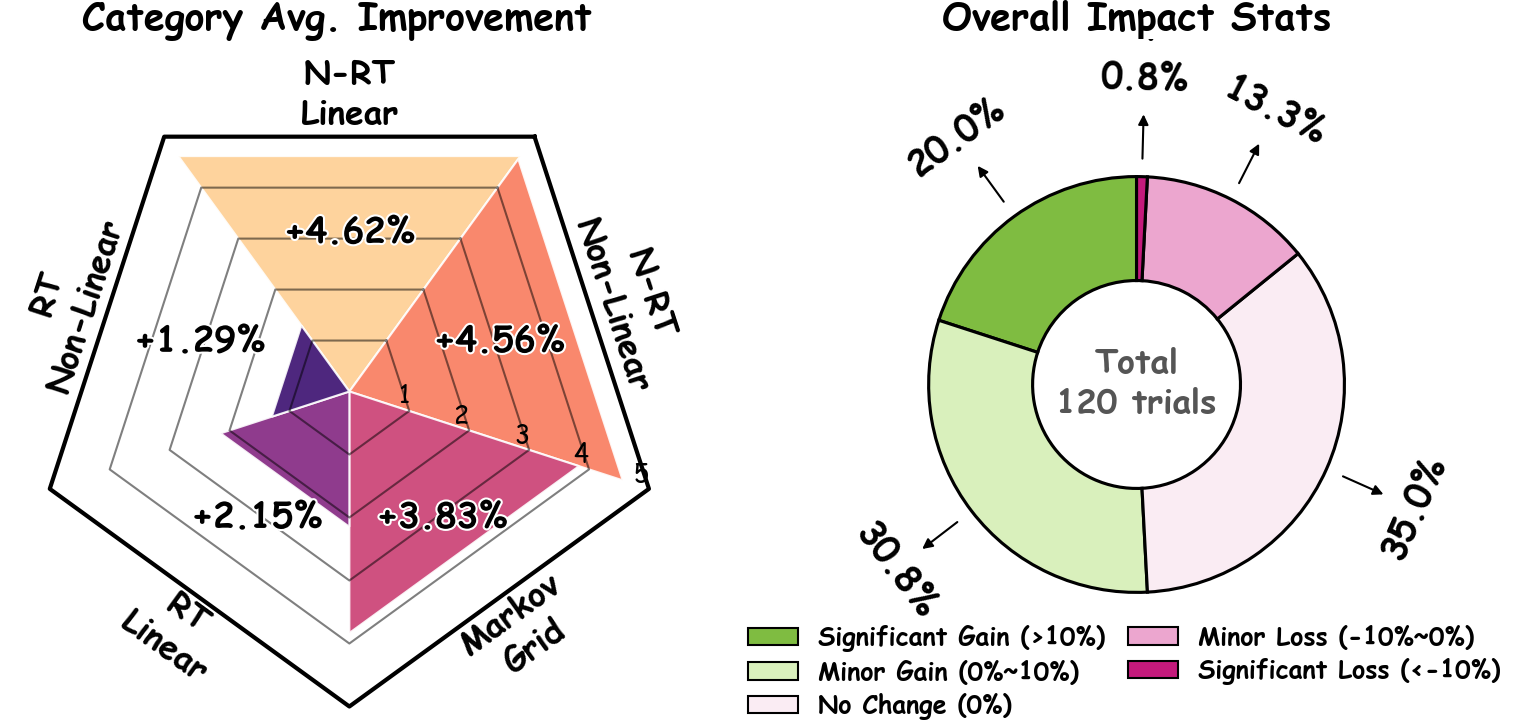}
}
\end{center}
\vspace{-1mm}
\caption{\textbf{Left:} Average improvement across 5 cognitive category tasks. \textbf{Right:} Overall improvement statistics of 120 trials.}
\vspace{-1mm}
\label{png:vlstatusb}
\end{figure}

\textbf{The Rich-Get-Richer Effect in Video-Based Reflection.} We observe that reflection generally yields improvements, but its efficacy is non-uniform. Data reveals two opposing trends. As shown in Figure~\ref{png:vlstatusa} (Bottom), gains scale positively with model capability. Gemini-2.5-Pro improves $6.47\%$ significantly more compared to GPT-4o-mini with $1.60\%$. As shown in Figure~\ref{png:vlstatusb} (Left), gains scale negatively with cognitive complexity and difficulty, dropping from $4.4\%$ in Non-Real-time to $1.7\%$ in Real-time. This mirrors a \textbf{\textit{Rich-Get-Richer}} phenomenon: models require a basic threshold of reasoning capital to effectively convert reflection into policy updates. Strong models possess sufficient capacity to internalize reflection, whereas weak models or even strong ones facing overwhelming complexity, fail to ground these insights into perception, reasoning and execution, resulting in diminishing returns.


\textbf{The Generalization Gap: Robust Humans vs. Brittle Agents.} Human players demonstrate remarkable generalization: they maintain high competitiveness and adaptivity regardless of whether the environment is a simple grid game or a complex, open-world simulation. In contrast, VLM agents exhibit severe degradation as complexity increases, lacking generalization. In \textit{Easy} games like \textit{Tic-Tac-Toe}, Gemini-2.5-Pro achieve perfect scores ($100$), effectively matching human expert baselines ($98.9$). In \textit{Hard} games that demand long-horizon reasoning or real-time control, such as \textit{Scene} or \textit{RDR 2}, model performance collapses to $0$, falling short of human rookie ($58.4\sim70.4$). While models have mastered distinct, simple tasks, they lack the generalization capability of humans to transfer knowledge and skills to more challenging and dynamic environments.

\textbf{The Knowing-Doing Gap: Semantic vs. GUI Control.} Although these models have demonstrated impressive performance on static GUI benchmarks \cite{OSWorld32, screenspot-pro31}, the precise visual grounding still remains a bottleneck in challenging video games. As shown in Table~\ref{mode-comp}, model averages $50.5$ in semantic mode, consistently outperforming the GUI mode average of $33.5$. While current VLMs possess strong reasoning capabilities for high-level planning, they still struggle with the visual grounding required to translate these plans into precise pixel coordinates. Besides, this gap suppresses the gains from learning. The gains $\sim8.7\%$ in semantic mode are consistently higher than $\sim3.75\%$ in GUI mode. While agents successfully learn the strategies, the strategies fail to translate into effective gameplay, highlighting a disconnect between high-level reasoning and low-level execution.

\begin{table}[t]
    \centering
    \caption{Performance of 4 VLMs on 4 games of GameVerse, evaluated with and without Video-base Reflection(\textit{VR}) in two modes.}
    \label{mode-comp}
    \newcommand{\std}[1]{{\scriptsize$\pm$#1}}
    
    \resizebox{1\columnwidth}{!}{
    \begin{tabular}{ll | cc | cc | cc | cc}
        \toprule
        \multirow{2}{*}{\textbf{Model}} & \multirow{2}{*}{\textbf{VR}} & \multicolumn{2}{c|}{\textbf{TicTacToe}} & \multicolumn{2}{c|}{\textbf{AngryBird}} & \multicolumn{2}{c|}{\textbf{Slay the Spire}} & \multicolumn{2}{c}{\textbf{Plants vs. Zombies}} \\
        \cmidrule(lr){3-4} \cmidrule(lr){5-6} \cmidrule(lr){7-8} \cmidrule(lr){9-10}
        & & \textbf{GUI} & \textbf{Semantic} & \textbf{GUI} & \textbf{Semantic} & \textbf{GUI} & \textbf{Semantic} & \textbf{GUI} & \textbf{Semantic} \\
        \midrule
        Random & --- & 26.3\std{19.3} & 30.4\std{24.1} & 0.0\std{0.0} & 7.4\std{6.0} & 0.0\std{0.0} & 0.0\std{0.0} & 0.0\std{0.0} & 0.0\std{0.0} \\
        \midrule
        \multirow{2}{*}{Qwen3-VL-8B} & No & 53.3\std{15.4} & 51.7\std{13.6} & 19.6\std{8.3} & 37.2\std{11.7} & \textbf{11.6\std{0.0}} & 11.7\std{0.0} & 33.4\std{4.1} & 54.2\std{5.8} \\
        & Yes & \cellcolor{red!5}51.8\std{14.6} & \cellcolor{green!60}67.0\std{12.9} & \cellcolor{green!40}31.2\std{10.8} & \cellcolor{green!60}63.2\std{15.1} & \cellcolor{red!15}7.4\std{8.4} & \cellcolor{red!15}5.4\std{5.2} & \cellcolor{green!60}51.3\std{21.4} & \cellcolor{green!40}65.2\std{11.3} \\
        \midrule
        \multirow{2}{*}{GPT-4o} & No & 58.6\std{4.2} & 66.8\std{10.2} & 14.3\std{9.0} & 49.3\std{13.0} & 2.0\std{0.0} & 25.3\std{10.2} & 32.2\std{12.4} & 42.7\std{16.0} \\
        & Yes & \cellcolor{green!40}64.2\std{9.8} & \cellcolor{red!5}64.3\std{8.0} & \cellcolor{green!10}15.4\std{12.3} & \cellcolor{green!60}61.5\std{17.6} & 2.0\std{0.0} & \cellcolor{green!20}29.4\std{8.1} & \cellcolor{green!20}36.4\std{15.2} & \cellcolor{green!20}53.4\std{7.8} \\
        \midrule
        \multirow{2}{*}{Gemini-2.5-Flash} & No & 88.2\std{16.4} & 93.8\std{9.1} & 9.4\std{8.2} & 45.8\std{9.2} & 2.0\std{0.0} & 11.6\std{0.0} & 51.1\std{13.2} & 70.2\std{8.5} \\
        & Yes & \cellcolor{green!30}95.1\std{12.4} & \cellcolor{green!30}\textbf{100.0\std{0.0}} & \cellcolor{green!10}10.2\std{6.4} & \cellcolor{green!60}68.4\std{15.4} & 2.0\std{0.0} & \cellcolor{green!25}16.4\std{0.0} & \cellcolor{green!25}\textbf{55.3\std{10.4}} & \cellcolor{green!40}\textbf{81.8\std{17.0}} \\
        \midrule
        \multirow{2}{*}{Gemini-2.5-Pro} & No & 90.0\std{9.0} & \textbf{100.0\std{0.0}} & 32.7\std{12.8} & 64.2\std{13.3} & 4.4\std{3.4} & \textbf{37.6\std{23.2}} & 33.4\std{17.2} & 45.4\std{3.3} \\
        & Yes & \cellcolor{green!25}\textbf{100.0\std{0.0}} & \textbf{100.0\std{0.0}} & \cellcolor{green!25}\textbf{42.2\std{11.3}} & \cellcolor{green!60}\textbf{81.0\std{15.8}} & \cellcolor{green!20}7.9\std{0.0} & \cellcolor{red!15}28.5\std{8.0} & \cellcolor{red!5}30.3\std{16.4} & \cellcolor{green!25}52.7\std{15.2} \\
        \midrule
        \multirow{2}{*}{Human Rookie} & No & 85.1\std{19.3} & 85.1\std{19.3} & 46.4\std{10.3} & 46.2\std{9.8} & 34.2\std{7.1} & 34.2\std{7.1} & 89.3\std{10.8} & 89.3\std{10.8} \\
        & Yes & \cellcolor{green!45}\textbf{99.4\std{1.9}} & \cellcolor{green!45}\textbf{99.4\std{1.9}} & \cellcolor{green!60}74.2\std{5.4} & \cellcolor{green!60}74.0\std{5.2} & \cellcolor{green!60}81.3\std{14.1} & \cellcolor{green!60}81.3\std{14.1} & \cellcolor{green!25}97.9\std{4.4} & \cellcolor{green!25}97.9\std{4.4} \\
        \midrule
        Human Expert & --- & 98.9\std{2.6} & 98.9\std{2.6} & \textbf{85.3\std{7.2}} & \textbf{85.4\std{7.3}} & \textbf{99.4\std{1.2}} & \textbf{99.4\std{1.2}} & \textbf{100.0\std{0.0}} & \textbf{100.0\std{0.0}}\\
        \bottomrule
    \end{tabular}
    }
\end{table}

\begin{figure}[t]
\begin{center}{
\includegraphics[width=1\columnwidth]{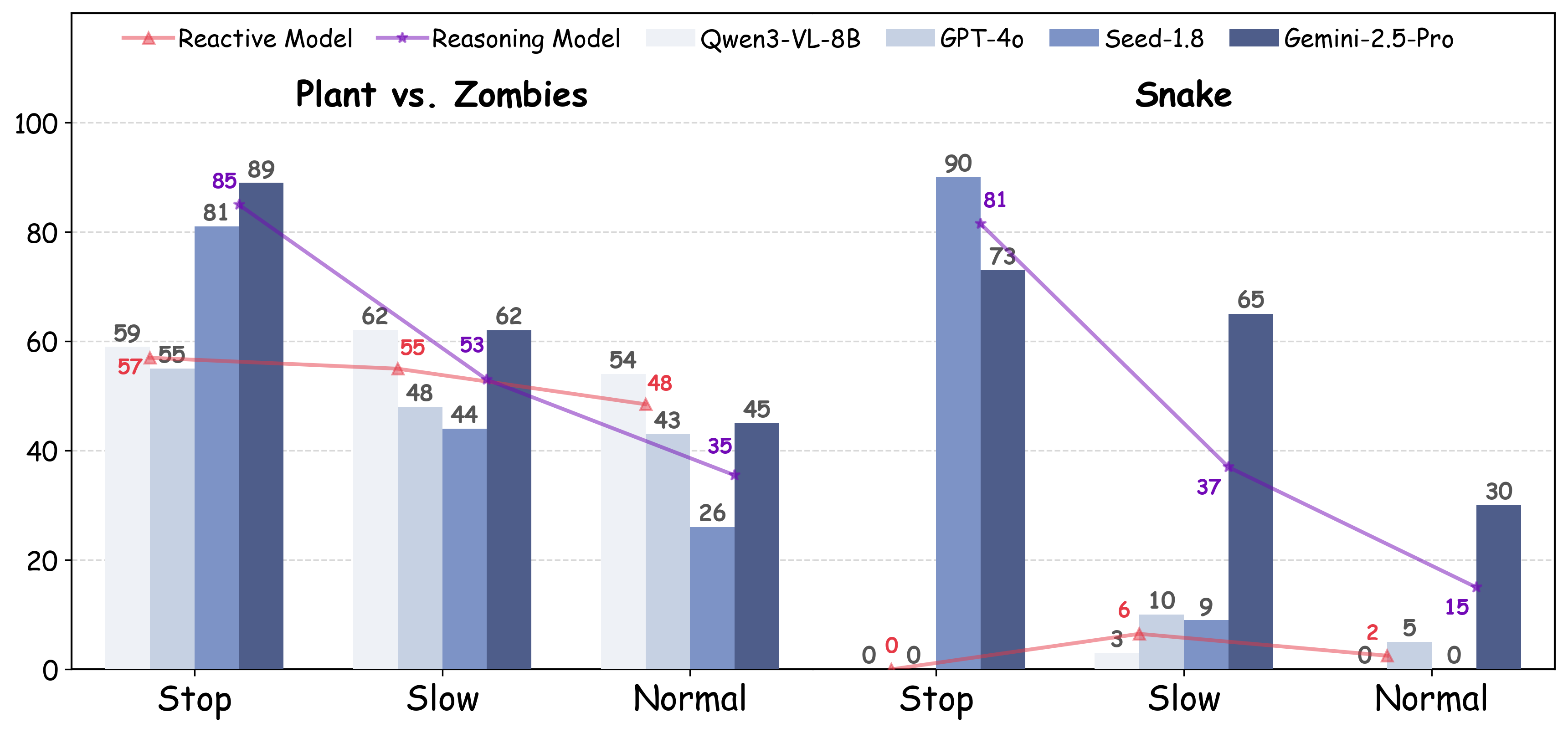}
}
\end{center}
\vspace{-1mm}
\caption{Performance of 4 VLMs on 2 games of GameVerse in semantic action mode with three latency settings. Reactive model is average performance of GPT-4o and Qwen3-VL-8B. Reasoning model is average performance of Seed-1.8 and Gemini-2.5-Pro.}
\vspace{-1mm}
\label{png:latency}
\end{figure}

\textbf{The Latency-aware Evaluation.} Figure~\ref{png:latency} reveals a performance divergence between reasoning and reactive models under three latency constraints. Reasoning models, Gemini-2.5-Pro and Seed-1.8, demonstrate high sensitivity to time constraints. In \textit{PvZ.}, their performance peaks ($\sim85$) in the \textit{Stop} setting but degrades sharply ($\sim35)$ in the \textit{Normal} real-time setting. This indicates that inference delays desynchronize the agent's planning from the evolving game state, nullifying their cognitive advantage. Conversely, reactive models, GPT-4o and Qwen3-VL-8B, exhibit stability across settings ($57, 55, 49$ in \textit{PvZ.}), suggesting their performance is bounded by reasoning capacity rather than response speed. Consequently, reducing latency is critical for deploying reasoning models in real-time environments, whereas it offers limited gains for reactive architectures.

Due to space limitations, Appendix~\ref{appd} presents a comprehensive analysis of game prompt, per-game performance, detailed error attribution, and reflection improvements.

\begin{table}[t]
    \centering
    \caption{Performance comparison of 2 VLMs on 3 games in semantic mode. The \textit{VR} settings are categorized into: \textbf{No}, \textbf{Self-F}, \textbf{Self-T}, \textbf{Self}, and \textbf{Other} (the alternative VLM).}
    \label{tab:vl-source-comp}
    
    \newcommand{\std}[1]{{\scriptsize$\pm$#1}}
    
    \resizebox{0.95\columnwidth}{!}{
    \begin{tabular}{ll | c | c | c | c} 
        \toprule
        \textbf{Model} & \textbf{VR} & \textbf{TicTacToe} & \textbf{AngryBird} & \textbf{Plants vs. Zombies} & \textbf{Avg} \\
        \midrule
        
        \multirow{5}{*}{\makecell[l]{GPT-4o-mini \\ \textit{weaker model}}} 
        & No    & 29.2\std{15.1} & 54.6\std{6.2} & 53.2\std{25.6} & 45.6 \\
        & Self-F & 43.1\std{4.3} & 57.4\std{13.4} & 52.1\std{18.0} & 50.8 \\
        & Self-T & 49.2\std{13.7} & 73.2\std{8.2} & 63.3\std{13.1} & 61.9 \\
        & Self & \textbf{52.3\std{16.0}} & 75.8\std{10.6} & \textbf{68.6\std{14.6}} & \textbf{65.5}\\
        & Other  & 45.0\std{7.2} & \textbf{77.1\std{7.0}} & 55.4\std{7.2} & 59.1 \\
        \midrule
        
        \multirow{5}{*}{\makecell[l]{Qwen3-VL-32B \\ \textit{stronger model}}} 
        & No    & 67.0\std{15.3} & 48.5\std{11.3} & 67.6\std{26.0} & 61.0 \\
        & Self-F & 84.5\std{16.8} & 54.4\std{12.9} & 77.3\std{13.2} & 72.0 \\
        & Self-T & 67.1\std{15.6} & 48.3\std{12.1} & 73.5\std{17.3} & 62.9 \\
        & Self & \textbf{84.5\std{13.2}} & 66.8\std{15.8} & \textbf{78.8\std{9.1}} & \textbf{76.7} \\
        & Other  & 69.2\std{16.4} & \textbf{74.5\std{10.2}} & 74.3\std{1.6} & 72.6 \\
        
        \bottomrule
    \end{tabular}
    }
\end{table}
\subsection{Unpacking Video-based Reflection}
\textbf{On the Roles of Failures and Tutorials}. To understand how models learn, we decompose the reflection process in Table~\ref{tab:vl-source-comp}. It shows that the integration of failure and tutorials (\textit{Self}) consistently outperforms relying on either source by at least $3.6\%$ for GPT-4o-mini and $4.7\%$ for Qwen3-VL-32B. Conceptually, this mirrors the paradigm of foundation model post-training \cite{pmlr-v267-chu25c}. Reflecting on failure (\textit{Self-F}) acts as a negative constraint mechanism, analogous to \textbf{Reinforcement Learning (RL)}, better with strong Qwen3-VL-32B $+11.0\%$. By identifying erroneous branches, the agent learns "what not to do," effectively refining its policy space. While learning from tutorials (\textit{Self-T}) provides positive behaviors, parallel to \textbf{Supervised Fine-Tuning (SFT)}, better with weak GPT-4o-mini $+16.4\%$. This supplies the agent with optimal trajectories that are difficult to discover through sparse-reward exploration alone. The integration of two mechanisms yields the most robust improvement, validating that a \textit{reflect-and-retry} loop can serve as a training-free proxy for combining SFT and RL in post-training. 

\textbf{On the Roles of Reflection Models}. Regarding the transferability of reflection, we employ a stronger Qwen3-VL-32B and a weaker GPT-4o-mini to exchange reflections. We observe that reflections generated by Qwen3-VL-32B do not enhance GPT-4o-mini (\textit{Other} $59.1$ $<$ \textit{Self} $65.5$). This suggests that the primary bottleneck is internalization rather than extraction; while extracting insights from video is relatively accessible, the capability to ground these insights into precise actions represents a higher intelligence barrier that cannot be bridged solely by better reflections.

\subsection{Qualitative Evaluation of Video-based Reflection}
\label{analysis_vl}
To vividly illustrate how reflection alters agent behavior, we visualize the trajectories before and after the \textit{VR} protocol.

\begin{figure}[t]
    \centering
    \begin{subfigure}[b]{0.23\textwidth}
        \centering
        \includegraphics[width=0.95\textwidth]{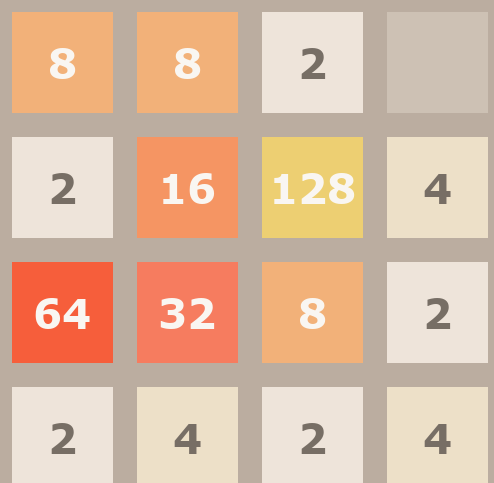}
        \caption{2048, \textit{VR}=No, Score=1930}
        \label{fig:top_left}
    \end{subfigure}
    \begin{subfigure}[b]{0.23\textwidth}
        \centering
        \includegraphics[width=0.95\textwidth]{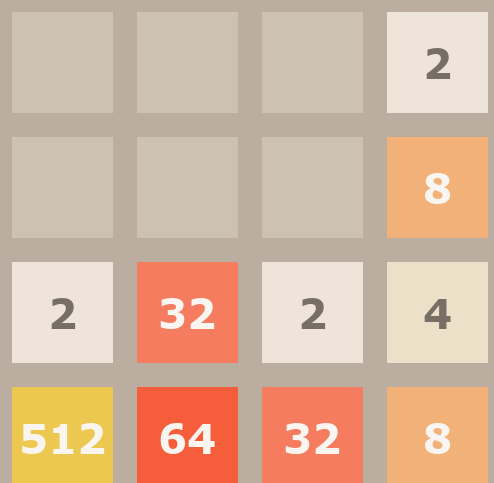}
        \caption{2048, \textit{VR}=Yes, Score=5960}
        \label{fig:top_right}
    \end{subfigure}

    \begin{subfigure}[b]{0.23\textwidth}
        \centering
        \includegraphics[width=0.95\textwidth]{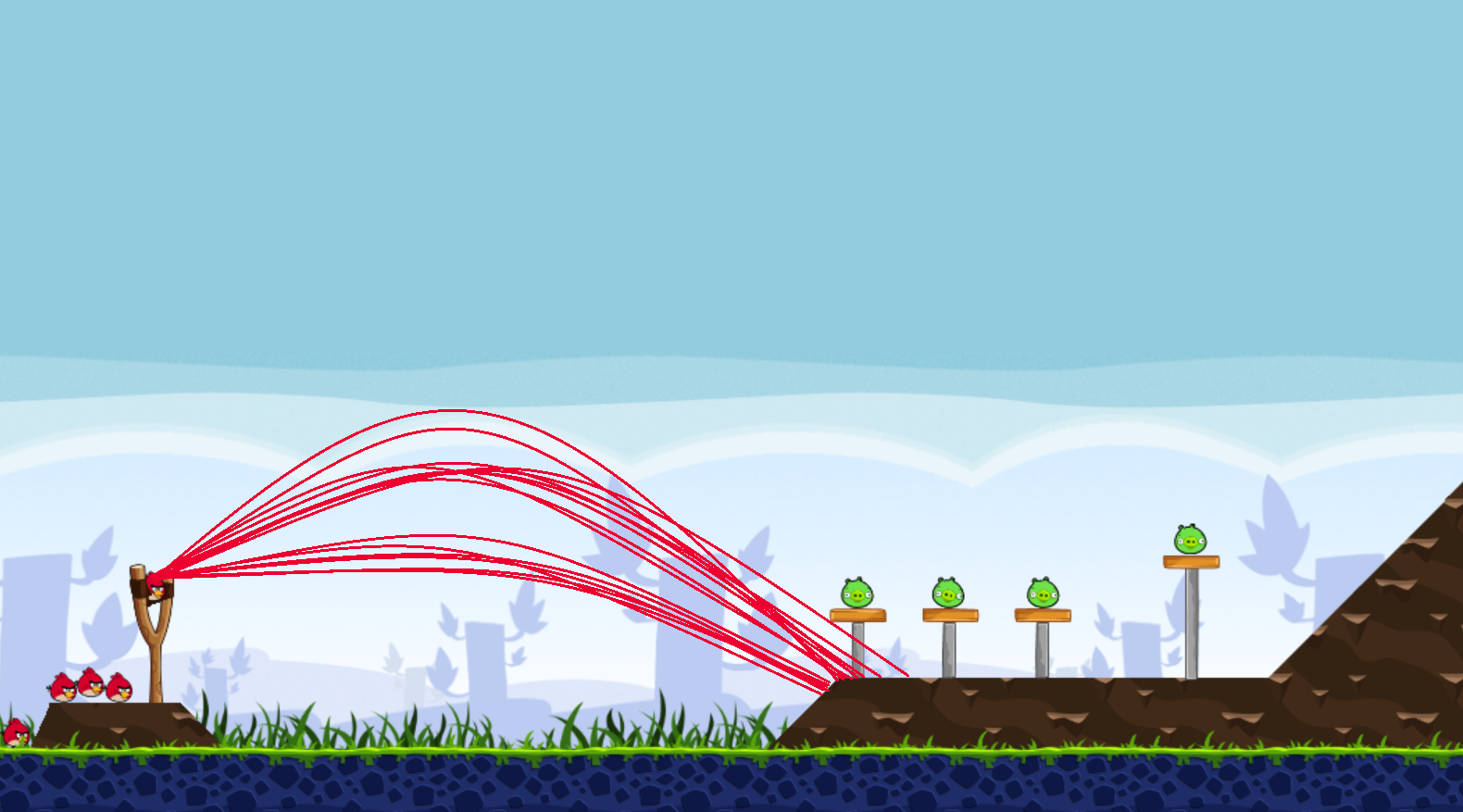}
        \caption{bird, \textit{VR}=No, Score=13250}
        \label{fig:bottom_left}
    \end{subfigure}
    \begin{subfigure}[b]{0.23\textwidth}
        \centering
        \includegraphics[width=0.95\textwidth]{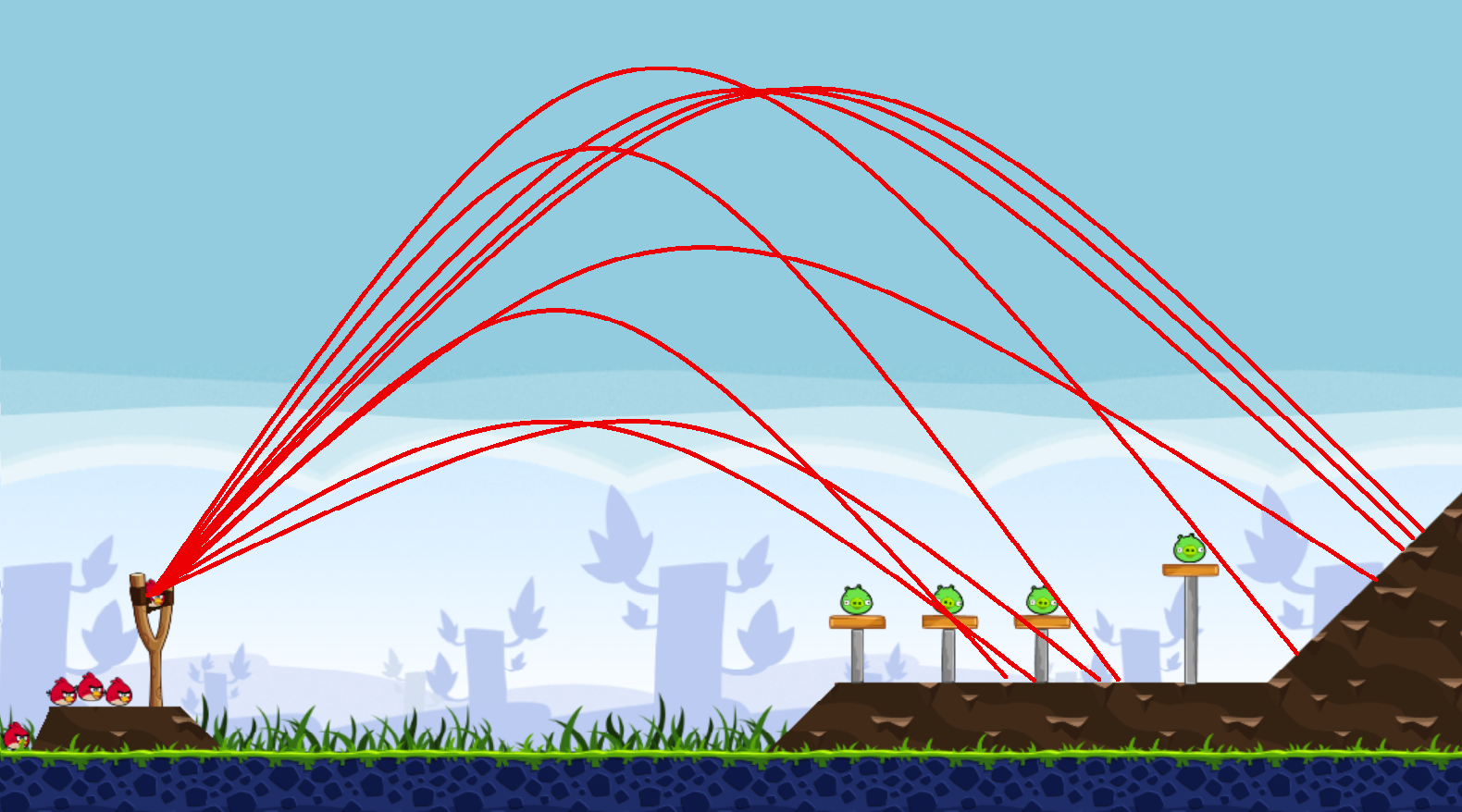}
        \caption{bird, \textit{VR}=Yes, Score=42580}
        \label{fig:bottom_right}
    \end{subfigure}

    \caption{Screenshots of Gemini-2.5-Pro playing 2048 and trajectories of Qwen3-VL-8B playing angry birds level 2.}
    \label{vl_analysis}
\end{figure}

\textbf{Case Study 1: Strategic Shift in 2048.} 
In \textit{2048}, the agent initially often made greedy moves. After reflection, it synthesized high-level strategies. In the reflection experiences, the agent explicitly formulated rules: 
(1) \textit{"Anchor the Largest Tile in a Corner"} to prevent board clutter; 
(2) \textit{"Strictly Limit Swipes to Three Directions"} to lock the anchor row; 
(3) \textit{"Build a Descending Snake Chain"} to facilitate merging. This strategic shift resulted in a score increase from 1920 to 5960, shown in Figure~\ref{fig:top_left} and \ref{fig:top_right}.

\textbf{Case Study 2: Physics Adaptation in Angry Birds.} 
For physics-based puzzles, the \textit{VR} protocol helped correct causal misunderstandings. In Level 2, the agent initially attempted to hit the top plank directly. After reflection, it refined its plan: \textit{"Aim for the central support pillar... Shoot the bird at the right-hand platform... ensure the falling blocks roll down the slope for a second wave of splash damage."} This physics adaptation led to a score improvement from 13,250 to 42,580, shown in Figure \ref{fig:bottom_left} and \ref{fig:bottom_right}.

\begin{figure*}[t]
    \centering
    \begin{subfigure}[b]{0.935\textwidth} 
        \centering
        \includegraphics[width=\linewidth]{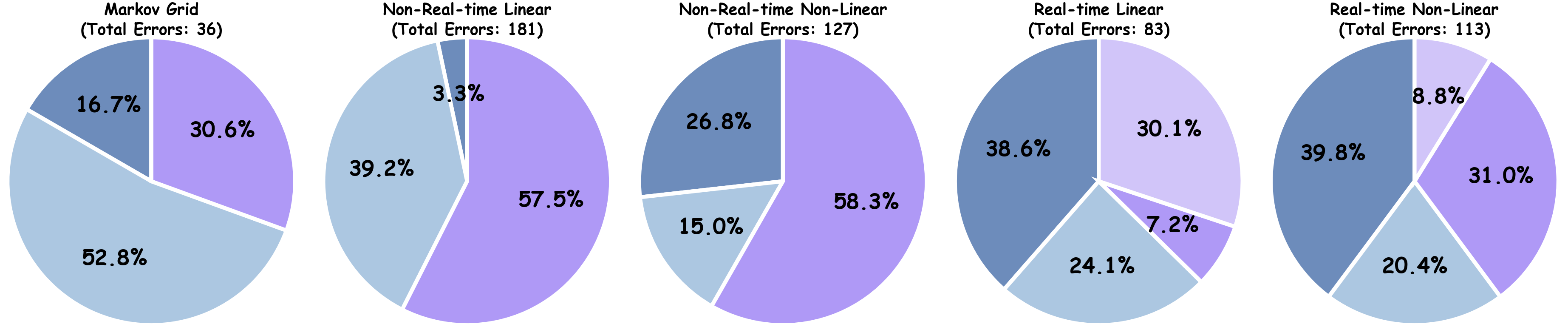}
        \phantomcaption 
        \label{png:error_analysis}
    \end{subfigure}

    \begin{subfigure}[b]{0.935\textwidth}
        \centering
        \includegraphics[width=\linewidth]{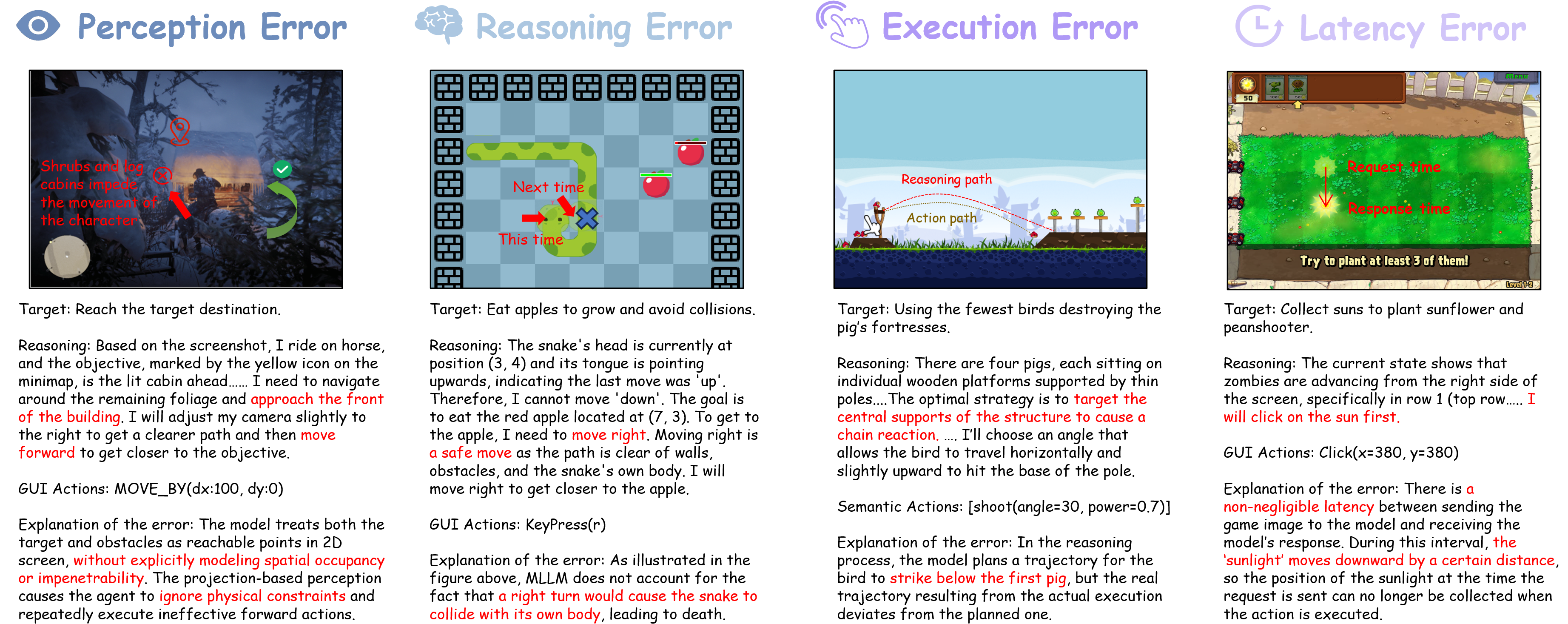}
        \phantomcaption
        \label{png:error_statistics}
    \end{subfigure}
    
    \caption{\textbf{Top:} GameVerse results breakdown of Qwen3-VL-32B. \textbf{Bot}: Failure cases about 4 types of errors in GameVerse.}
    \label{png:error}
\end{figure*}

\subsection{Additional Insights of Video-based Reflection}
Unlike humans who derive stable and substantial gains, model improvements are sometimes inconsistent and constrained. Why does reflection work, and where does it fail? We identify one core advantage countered by two barriers:

\textbf{Advantage: Dual-Stream Capability Acquisition.}
The improvements stem from two distinct cognitive gains extracted from failure and tutorials: \textbf{\textit{(1)~Abstract Principle Comprehension}}, where the agent distills high-level logic (e.g., "Build a snake chain" in 2048) to enhance global planning; and \textbf{\textit{(2)~Specific Behavioral Mimicry}}, where the agent learns concrete, context-specific maneuvers (e.g., "Aim at the central pillar" in Angry Birds). This indicates that current VLMs can effectively parse multimodal demonstrations into both strategic and tactical knowledge.

\textbf{Barrier I: The Cognitive Bottleneck.}
We identify a \textit{Cognitive Bottleneck} that fundamentally restricts the agent performance. Embodied environments demand robust spatial perception and long-horizon reasoning. We observe that even when agents explicitly retrieve the correct insight, they often fail to \textit{internalize} it into the next attempt. This is due to inherent limitations in reasoning and planning capabilities. Unlike humans who seamlessly integrate new lessons into their mental models, models struggle to "compile" textual reflections into a robust, updated policy, leading to repeated strategic errors despite correct theoretical knowledge.

\textbf{Barrier II: The Physical Disconnect.}
Complementary to the cognitive constraint, we identify a \textit{Physical Disconnect} that hinders the agent's interaction with the environment. Embodied environments demand real time reaction and precise control. We observed that even when the strategy is sound, the agent is hindered by inference latency and knowing-doing gap. The high latency desynchronizes the agent's reasoning from the rapidly evolving world, while the knowing-doing gap severs the link between high-level semantic planning and precise low-level execution. Consequently, the agent remains too slow and clumsy, preventing the success of reasoning models in \textit{GameVerse}.

\subsection{Error Analysis}
To diagnose the capability boundaries of VLMs, we classify gameplay failures into four error types based on the agent's capability: \textit{\textbf{Perception Error:}} Fail to interpret visual information correctly. \textit{\textbf{Reasoning Error:}} Fail in logical deduction or future state prediction, where the agent misinterprets the causal consequences of an action. \textit{\textbf{Execution Error:}} There is a misalignment between a correct high-level semantic plan and its low-level motor implementation. \textit{\textbf{Latency Error:}} This is from the temporal desynchronization between observation and action due to inference time. 

We give the breakdown of Qwen3-VL-32B and typical failure cases in Figure~\ref{png:error}. The total error count escalates as tasks shift from static grids to dynamic, non-linear worlds. As visual fidelity increases from abstract grids to high-fidelity 3D environments, the proportion of \textit{perception errors} rises sharply ($16.7\%\to 39.8\%$), indicating that processing complex visual semantics remains a bottleneck. Regardless of the category, \textit{reasoning and execution errors} consistently maintain a substantial proportion ($> 15\%$), suggesting that the reasoning barrier and physical disconnect is a systemic issue. In Real-time settings, \textit{latency errors} emerge as a critical failure mode, confirming that for dynamic agents, inference speed is as vital as cognitive accuracy.

\section{Discussion}

\textbf{Conclusion.} In this work, we introduced \textbf{GameVerse}, benchmarking VLM agents through \textit{reflect-and-retry} loop. Through a cognitive hierarchical taxonomy across 15 games, we revealed that while current VLMs show promise in simple environments, they struggle with generalization in challenging settings. Reflection yields improvements, but models still lack the human-like robustness to consistently internalize experience. Moreover, the integration of failures and tutorials combines negative error-pruning with imitating behaviors, mirroring the complementary benefits of combining reinforcement and supervised learning in post-training~\cite{pmlr-v267-chu25c}. Beyond a benchmark, \textit{GameVerse} serves as an arena where agents mimic human reflection and generalize across virtual worlds toward general intelligence.

\textbf{Limitations and Future Work.} Currently, our agent passively receives failures and tutorials in a single turn. Future systems could employ active, multi-turn interactions, allowing agents to dynamically query specific video segments or diagnose failures through iterative dialogue. While our scalable milestone evaluation pipeline reduces the labor intensity in previous work~\cite{cradle14, videogamebench4, flashadventure6}, it relies on a single VLM. Future iterations could enhance robustness and accuracy by incorporating multi-model voting or hybrid visual-text retrieval. 

\section*{Impact Statement}
This paper presents work whose goal is to advance the field of Vision-Language Models. There are many potential societal consequences of our work, none which we feel must be specifically highlighted here.

\bibliography{Reference}
\bibliographystyle{icml2026}

\newpage
\appendix
\onecolumn

\startcontents[appendix]
\setcounter{tocdepth}{2} 

\begin{center}
    \hrule 
    \textbf{\large \textit{GameVerse}: Can Vision-Language Models Learn from Video-based Reflection?\\} 
    \large (Supplementary materials) 
    \vspace{1mm}
    \hrule    
\end{center}

This appendix provides supplementary materials to support the findings presented in the main paper, ensuring providing detailed information and offering deeper insights into the GameVerse. 

First of all, for the readers’ better understanding, we describe the main content of the appendix. 

\textbf{Appendix A: Implementation Details.} This section detailed the architecture of zeroshot and memory agent, the pipeline of milestone detection and scoring and the universal GUI action space.

\textbf{Appendix B: Human Baselines.} This section detailed the recruitment methodology, and testing protocols used to establish the Human Rookie and Human Expert baselines.

\textbf{Appendix C: Game Selection Criteria.} This section details the rigorous hierarchical taxonomy, ensuring a diverse coverage of cognitive demands and difficulty tiers. 

\textbf{Appendix D: Game Environment Details and Extended Analysis.} This section provides detailed descriptions of the task objectives and mechanics for each of the 15 games. Furthermore, it presents an extended qualitative analysis of the Video-Based Reflection efficacy and a comprehensive case study of agent failures.

\vspace{1em}
\hrule
\vspace{0.5em}
\textbf{\large Contents} 
\printcontents[appendix]{}{1}{\setcounter{tocdepth}{2}}
\vspace{1em}
\hrule
\vspace{2em}

\section{Implementation Details}
\label{appb}
\subsection{Agent Architecture}
In GameVerse, we initially use two kinds of agents, zeroshot agent and memory agent. For \textit{Tic-Tac-Toe, Baba Is You, 2048, Maze, Angry Birds, Slay the Spire, Snake, Plant vs. Zombie, Forza Horizon 5} and \textit{mini metro}, we use zeroshot agent because the current visual input includes most of the information and the goal of the game remains the same. For \textit{Ace Attorney, Civilization VI, Scene Investigators Demo, Genshin Impact} and \textit{Red Dead Redemption 2}, we use memory agent because these games include long story arcs and the memory of key information is needed. We conduct the comparison experiments of zero-shot and memory agent for these games in Appendix~\ref{appd}.

For zeroshot agent, only the current screenshot and the screenshot-action pair from the last one step are provided as context for VLM agents. For memory agent, the screenshot-action pair from the last five step are provided as context for VLM agents to update a short-term memory. The agent can choose to store the key information as long-term memory into vector database and retrieve the long-term memory based on relevance to the current game state. 

\textbf{Zero-shot Agent Details.} Zeroshot agent is designed to handle the situation where the screenshot covers most information of the current game state, and the game goal is obvious and stable. The agent operates through only one module:

1. \textit{Action Inference Module}.This module performs direct action inference from the current game state without maintaining any form of historical context. 

\textbf{Memory Agent Details.} Memory agent is designed to enable long-term memory capabilities for game-playing agents, addressing the challenge of maintaining contextual awareness across extended gameplay sessions. The agent operates through a sequential pipeline of four interconnected modules:

1. \textit{History Review and Reasoning Module}. This module processes short-term history by reviewing the most recent 5 image-action pairs alongside the previous history summary. It employs an importance-based replacement strategy with four priority levels: CRITICAL (confirmed contradictions, successful strategies), HIGH (potential contradictions, important testimonies), MEDIUM (dialogue progression, routine actions), and LOW (outdated information, minor observations). The module maintains a fixed-length summary (500-600 words) through selective compression and merging of similar information, ensuring that the most valuable context is preserved while staying within token limits.

2. \textit{Long-term Memory Retrieval Module}. This module implements semantic retrieval from a persistent vector database. We use ChromaDB as the vector store, supporting both OpenAI and Qwen embedding models for flexibility across different deployment scenarios. The retrieval query is constructed from three components: (1) the current observation state, (2) the compressed history summary, and (3) the reasoning from the previous module. Retrieved memories are filtered using a similarity threshold (default 0.4) to ensure relevance, with top-k (default k=3) most similar memories returned.

3. \textit{Action Inference with Memory Module}. This module performs action selection by integrating multiple information sources: the current game screenshot, the compressed history summary, reasoning context, and retrieved long-term memories. The module also implements an optional memory storage mechanism where the agent decides whether to save new information to long-term memory based on predefined importance criteria (e.g., strategic decisions, puzzle solutions, important character information). A deduplication mechanism prevents storing highly similar memories using a similarity threshold (default 0.8).

4. \textit{Short-term History Update Module}. This module maintains the temporal consistency of the agent's memory by recording the current step's state, image, action, and reasoning. It ensures that image-action pairs are properly indexed for subsequent history review iterations, implementing a sliding window approach over the agent's trajectory.

GameVerse is designed with a modular architecture that enables easy integration of different components and facilitates ablation studies. Different modules such as self-reflection, subtask planning and etc., can be easily added for future studies.

\subsection{GUI Action Space}
Following the GUI action settings in UI-TARS~\cite{UITARS118} and FlashAdvanture~\cite{flashadventure6}, we design our GUI action space as shown in Table~\ref{tab:gui_actions}.
\begin{table}[htbp]
    \centering
    \caption{GUI Action Types Summary}
    \label{tab:gui_actions}
    \resizebox{\textwidth}{!}{
        \begin{tabular}{l l p{10cm}} 
            \toprule
            \textbf{Action Type} & \textbf{Description} & \textbf{Parameters} \\
            \midrule
            
            \multicolumn{3}{l}{\textit{\textbf{Mouse Movement}}} \\
            \midrule
            MOVE\_TO & Move to position & \texttt{x}, \texttt{y} (int, req): Target coordinates (window relative) \\
            MOVE\_BY & Move relatively & \texttt{dx}, \texttt{dy} (int, req): Offset; \texttt{duration} (float, opt): Seconds \\
            
            \midrule
            \multicolumn{3}{l}{\textit{\textbf{Mouse Click}}} \\
            \midrule
            CLICK & Click mouse & \texttt{x}, \texttt{y} (opt); \texttt{button} (left/right/middle); \texttt{num\_clicks} \\
            RIGHT\_CLICK & Right click & \texttt{x}, \texttt{y} (int, opt): Click coordinates \\
            DOUBLE\_CLICK & Double click & \texttt{x}, \texttt{y} (int, opt): Click coordinates \\
            
            \midrule
            \multicolumn{3}{l}{\textit{\textbf{Mouse Drag}}} \\
            \midrule
            MOUSE\_DOWN & Press button & \texttt{button} (str, opt): "left", "right", or "middle"; \texttt{duration} (float, opt) \\
            MOUSE\_UP & Release button & \texttt{button} (str, opt): "left", "right", or "middle";  \\
            DRAG\_TO & Drag to target & \texttt{x}, \texttt{y} (int, req): Target coordinates \\
            
            \midrule
            \multicolumn{3}{l}{\textit{\textbf{Mouse Scroll}}} \\
            \midrule
            SCROLL & Scroll wheel & \texttt{dx}, \texttt{dy} (int, req): Scroll amount (+/-); \texttt{duration} (float, opt) \\
            
            \midrule
            \multicolumn{3}{l}{\textit{\textbf{Keyboard Input}}} \\
            \midrule
            TYPING & Type text & \texttt{text} (str, req); \texttt{interval} (float, opt) \\
            PRESS & Press key & \texttt{key} (str, req); \texttt{duration} (float, opt) \\
            KEY\_DOWN & Key down & \texttt{key} (str, req): Key name; \texttt{duration} (float, opt) \\
            KEY\_UP & Key up & \texttt{key} (str, req): Key name;  \\
            HOTKEY & Key combo & \texttt{keys} (list, req): e.g., \texttt{['ctrl', 'c']}; \texttt{duration} (float, opt) \\
            
            \midrule
            \multicolumn{3}{l}{\textit{\textbf{Control Flow}}} \\
            \midrule
            WAIT & Wait time & \texttt{duration} (float, req): Seconds to wait \\
            DONE & Task success & No parameters \\
            FAIL & Task failure & No parameters \\
            \bottomrule
        \end{tabular}
    }
\end{table}

\subsection{Video-based Reflection Pipeline}
\label{reflection:appendix}
\begin{figure*}[htbp]
\begin{center}{
\includegraphics[width=0.8\textwidth]{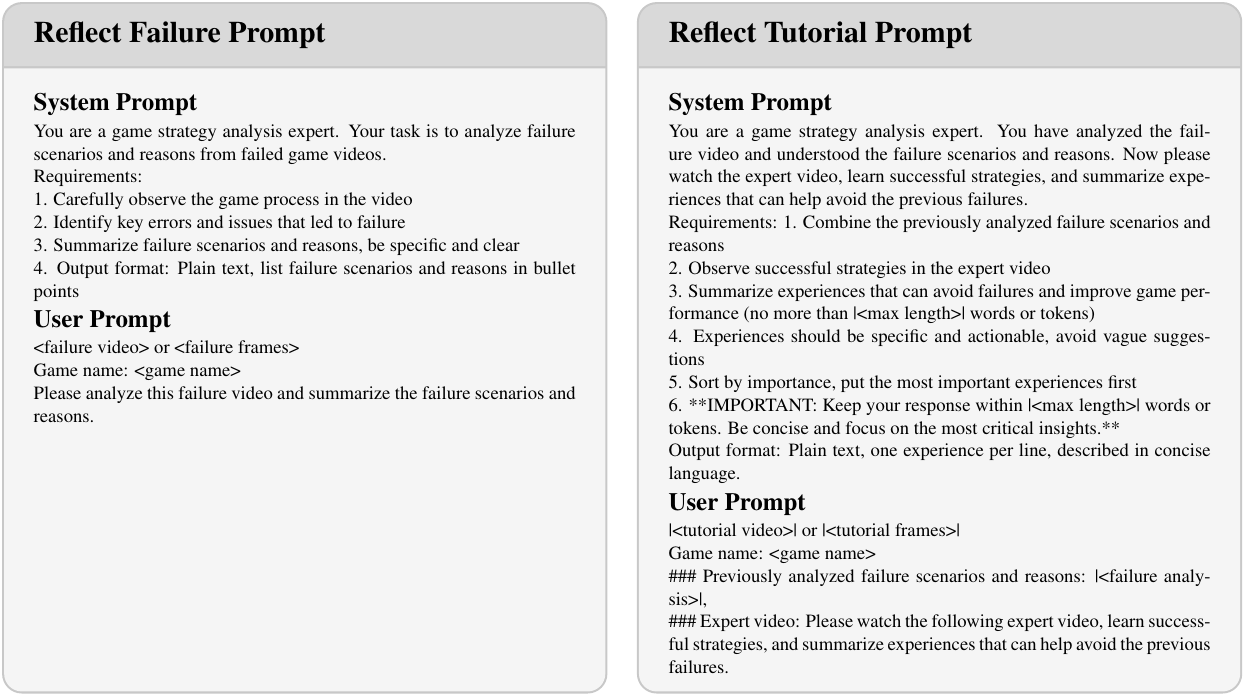}
}
\end{center}
\vspace{-1mm}
\caption{Prompt of reflecting experience from failure and tutorial}
\vspace{-1mm}
\label{png:vl_prompt}
\end{figure*}

Given the limited context window of current VLMs and the complexity of processing multiple videos concurrently, we adopt a sequential analysis strategy. We first query the model to diagnose the failure video—identifying the specific scenario and root cause. This diagnostic context then guides the analysis of expert demonstrations, facilitating the extraction of core insights that are injected into the subsequent prompt. The prompt for extracting experience of failure and tutorials is in Figure~\ref{png:vl_prompt}. These extracted experience will be injected into the next attempt, taking place of |<learned\_experience>| in the system prompt. 

For Qwen3-VL-8B/32B, Gemini-2.5-Pro/flash and seed-1.8, whose API facilitates direct video ingestion, GPT-4o and GPT-4o-mini accept only image-based inputs. To address this, we implement uniform sampling at 1 fps. When the extracted frames exceed the context window due to video length, we iteratively downsample the sequence until it accommodates the model's context limits.

We provide links in Table~\ref{tab:gameverse_links} to all walkthroughs used for reflection and milestone extraction, as well as estimating game lengths for human expert. Each of these walkthroughs assumes full knowledge of the game and does not consider time spent exploring. The videos were not subjectively selected; rather, they were identified as the top-ranked search results on mainstream video platforms corresponding to the target game tasks. The similar videos are abundant online, ensuring our scalability to more video games.

\begin{table*}[htbp]
    \centering
    \caption{List of longplay walkthrough links for games in GameVerse.}
    \label{tab:gameverse_links}
    \resizebox{0.8\textwidth}{!}{
    \begin{tabular}{l | c| c | c}
    \toprule
\textbf{Game} & \textbf{Video Type} & \textbf{Link} & \textbf{\textit{Selected} Time} \\ \midrule
Tic-Tac-Toe & Skill Tutorial & \url{https://www.youtube.com/shorts/ZBZcnImNmhk} & 00:00:59 \\
Baba Is You & Walkthrough & \url{https://www.bilibili.com/video/BV1wb411p7ja/} & 00:00:38\\
2048 & Skill Tutorial & \url{https://www.bilibili.com/video/BV1VH4y1K7Gr/} & 00:03:02 \\ \midrule
Maze & Skill Tutorial & \url{https://www.bilibili.com/video/BV1pp4y197fY/} & 00:03:37 \\
Angry Birds & Walkthrough & \url{https://www.bilibili.com/video/BV1b54y1s7pe/} & 00:01:31\\
Slay the Spire & Skill Tutorial & \url{https://www.bilibili.com/video/BV1RVdLYyEFf/} & 00:09:36\\ \midrule
Ace Attorney & Walkthrough & \url{https://www.bilibili.com/video/BV1oUpEeaEyU/} & 00:21:50\\
Civilization VI & Walkthrough & \url{https://www.bilibili.com/video/BV1Ax41197bu/} & 00:19:23\\
Scene Investigators & Walkthrough & \url{https://www.bilibili.com/video/BV1NA411U7qj/}& 00:27:26 \\\midrule
Snake & Walkthrough & \url{https://www.youtube.com/shorts/CHhnAhoO_ak}& 00:00:44 \\ 
Plants vs. Zombies & Walkthrough & \url{https://www.bilibili.com/video/BV12b41167b7/} & 00:05:51\\
Forza Horizon 5 & Skill Tutorial & \url{https://www.bilibili.com/video/BV1Pi4we3EdF/} & 00:04:58\\ \midrule
Mini Metro & Skill Tutorial & \url{https://www.bilibili.com/video/BV1cY4y1d7vS/} & 00:04:41\\
Genshin Impact & Walkthrough & \url{https://www.bilibili.com/video/BV1H64y1e7YJ/} & 00:25:51\\
Red Dead Redemption 2 & Walkthrough & \url{https://www.bilibili.com/video/BV1hb411N7Rk/}& 00:44:07 \\ \bottomrule
\end{tabular}%
}
\end{table*}

\subsection{Milestone Scoring Pipeline}
\label{milestone:appendix}
\begin{figure*}[htbp]
\begin{center}{
\includegraphics[width=1\textwidth]{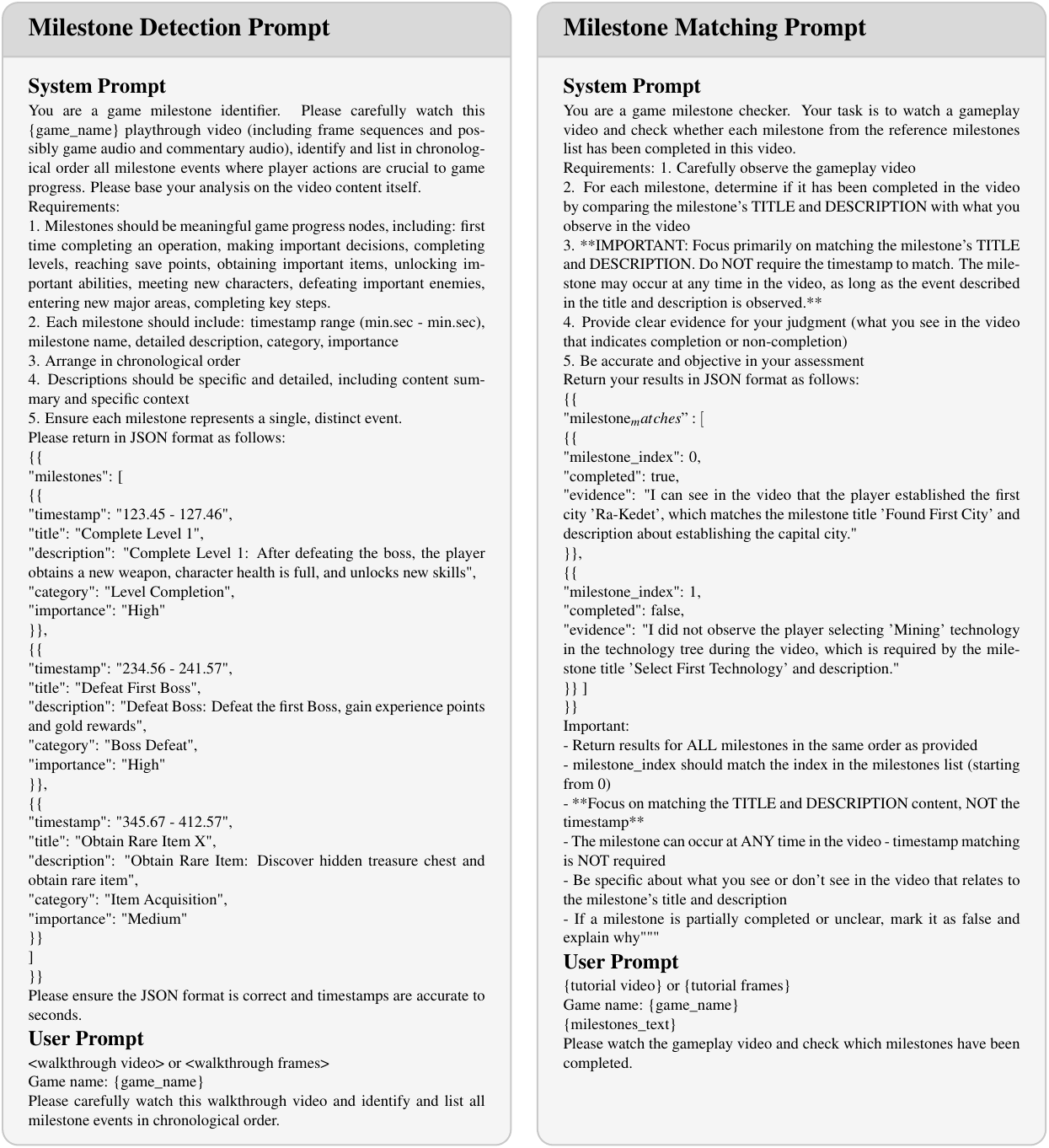}
}
\end{center}
\vspace{-1mm}
\caption{Prompt of milestone extraction and milestone matching}
\vspace{-1mm}
\label{png:milestone_prompt}
\end{figure*}

There are 7 games lacking intrinsic scoring system in \textit{GameVerse}, \textit{Baba is you, Ace Attorney, Civilization VI, Scene Investigator, Plants vs. Zombies, Genshin Impact, Red Dead Redemption 2}. Facing the absence of intrinsic scoring systems in long-horizon games (e.g. \textit{Genshin Impact, RDR 2}, previous methods necessitated either brittle hash-matching~\cite{videogamebench4} or labor-intensive manual milestone annotation~\cite{wukong55, cradle14, videogamebench4, flashadventure6}. We introduce an milestone scoring pipeline utilizing advanced VLMs (Gemini-3-pro used in this paper), which offer extensive knowledge (knowing these famous games), long-context video comprehension (understanding the gameplay video), and web-search capabilities. This system extracts standardized milestone JSON profiles $M_{ref}$ from expert walkthrough videos and employs the model's reasoning capabilities to align these milestones with agent playthrough videos, thereby generating procedural scores without human intervention. The milestone detection and matching prompts are shown in Figure~\ref{png:milestone_prompt} 

\subsubsection*{A.4.1 Milestone Detection Pipeline}
\begin{figure*}[htbp]
\begin{center}{
\includegraphics[width=1\textwidth]{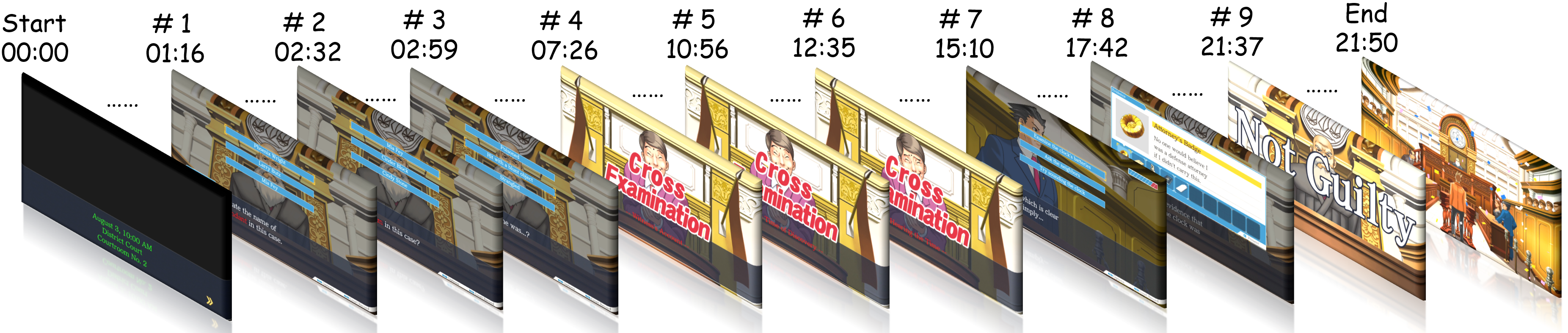}
}
\end{center}
\vspace{-1mm}
\caption{9 milestones are extracted by Gemini-3-pro in the walkthrough video of Ace Attorney.}
\vspace{-1mm}
\label{png:milestone_example}

\end{figure*}

We take expert walkthrough videos as input to extract representative milestones, guided by specific prompts. The videos of these games are the walkthrough videos mentioned in Table~\ref{tab:gameverse_links}. An example of extracting milestone is shown Figure~\ref{png:milestone_example}.  We provide a detailed breakdown of the milestones extracted from the seven games in Table~\ref{tab:milestone_detect}, covering metrics such as count, hallucination rate, and representativeness as evaluated by 10 human experts. Experimental results demonstrate high expert consensus regarding the validity of the extracted milestones, confirming them as significant, stage-defining nodes within the gameplay. Furthermore, the hallucination rate remains within a controllable range. We check the detailed hallucination and find that inaccuracies are primarily confined to fine-grained details—such as entity descriptions and specific perceptions—while macro-level scene depictions remain accurate. Consequently, the proposed milestone extraction pipeline exhibits fundamental robustness, scalability, and reusability.

\begin{table*}[htbp]
    \centering
    \caption{The breakdown of the extracted milestones. \textit{L} means level. Hallucination rate is defined as  $\frac{\# inaccurate~~milestones}{\# total~~milestones}$. Representation score is defined as how representative the milestones are to significant game nodes, ranging in $[0,1]$. The score is evaluated by 10 human video game experts}
    \label{tab:milestone_detect}
    \resizebox{0.8\textwidth}{!}{
    \begin{tabular}{l | c| c | c| c}
    \toprule
\textbf{Game} & \textbf{Video Length} & \textbf{\# Milestone} & \textbf{Hallucination Rate} & \textbf{Representation Score} \\ \midrule
Baba Is You & 00:00:38 & 3 (\textit{L}0), 5 (\textit{L}1)& 12.5\% &0.92\\
Ace Attorney & 00:21:50 &9 &0\% &1.00 \\
Civilization VI & 00:19:23 &12 &17\% &0.81\\
Scene Investigators & 00:27:26 &17 &11\% &0.95 \\
Plants vs. Zombies & 00:05:51 &7 (\textit{L}1), 6(\textit{L}2), 6(\textit{L}4) &5\% &0.78 \\
Genshin Impact & 00:25:51 &7 &0\% &0.95\\
Red Dead Redemption 2 & 00:44:07 &10 &10\% &0.87\\ \bottomrule
\end{tabular}%
}
\end{table*}

\subsubsection*{A.4.2 Milestone Matching Pipeline}
Following extracting the offline milestone JSON files, we establish a reusable framework for evaluating the agent's gameplay process score via video matching. Specifically, the 'gameplay video' consists of the \textit{discrete frame} sequences received by the agent at each step. We adopt this discrete approach because recording the continuous gameplay process would yield excessively long videos that exceed context windows, given the inference latency which ranges from seconds (GPT-4o, GPT-4o-mini) to minutes (Gemini-2.5-Pro, Seed-1.8). Finally, we employ Gemini-3-Pro, guided by specific prompts, to compute matching scores between the milestones and the agent’s gameplay. The results will include whether the milestone is completed and the evidence to prove for further human verified. In Table~\ref{tab:milestone_score}, we present the mean VLM-evaluated and human-evaluated matching scores on agent play video and human play video with percentage difference across seven games. The comparison distinguishes between human gameplay videos (continuous frame sequences) and agent gameplay frames (discrete input frames). The results demonstrate that milestone matching via advanced VLMs exhibits robustness and stability. However, continuous video matching yields higher accuracy compared to discrete environmental frames. We observe that matching scores for agent videos tend to be lower; this is attributed to the discrete sampling nature, where specific milestones occurring in the temporal gaps between frames may fail to be visually captured. Consequently, enhancing the continuity of agent gameplay videos without exceeding context window limits remains a meaningful research problem.

\begin{table*}[htbp]
    \centering
    \caption{The average VLM-evaluated and human-evaluated matching scores on agent play video and human play video with percentage difference across seven games with VL.}
    \label{tab:milestone_score}
    \resizebox{0.9\textwidth}{!}{
    \begin{tabular}{l | c c c| ccc}
    \toprule
    \multirow{2}{*}{\textbf{Game}} & \multicolumn{3}{|c|}{\textbf{Agent gameplay video (discrete input frames)}} & \multicolumn{3}{c}{\textbf{Human gameplay video (continuous frame sequences)}}  \\
    \cmidrule(lr){2-4} \cmidrule(lr){5-7} 
    & \textbf{VLM eval.} & \textbf{human eval.} & \textbf{Percentage Diff. $\downarrow$} & \textbf{VLM eval.} & \textbf{human eval.} & \textbf{Percentage Diff. $\downarrow$} \\ \midrule
Baba Is You &0.68$\pm$0.15 &0.74$\pm$0.10 &8.1\% & 0.98$\pm$0.06 &1.00$\pm$0.00 &2.0\%\\
Ace Attorney &0.29$\pm$0.17 &0.33$\pm$0.19 &12.1\% & 1.00$\pm$0.00 &1.00$\pm$0.00 &0\% \\
Civilization VI &0.04$\pm$0.08 &0.04$\pm$0.08 &0.0\% & 0.92$\pm$0.11 &0.96$\pm$0.09 &4.2\%\\
Scene Investigators &0.05$\pm$0.04 &0.05$\pm$0.04 &0.0\%  & 0.90$\pm$0.09 &0.90$\pm$0.09 &0\% \\
Plants vs. Zombies &0.34$\pm$0.15 &0.38$\pm$0.13 &10.5\% & 1.00$\pm$0.00 &0.98$\pm$0.04 &2.0\% \\
Genshin Impact &0.12$\pm$0.05 &0.14$\pm$0.00 &14.2\% & 0.83$\pm$0.15 &0.88$\pm$0.18 &5.6\%\\
Red Dead Redemption 2 &0.10$\pm$0.02  &0.10$\pm$0.02  & 0.0\% & 0.86$\pm$0.15 &0.83$\pm$0.21 &3.4\%\\ \bottomrule
\end{tabular}%
}
\end{table*}

\subsection{Latency Setting}
We conducted latency ablation studies on two representative real-time games: Plants vs. Zombies and Snake. Given the inherent variability and uncontrollability of API inference and network transmission latencies, we instead modulated the environmental latency (i.e., game simulation speed) to systematically assess its impact on agent performance. Specifically, we established three latency levels, \textbf{Stop}, \textbf{Slow}, and \textbf{Normal}. 

In \textit{Plants vs. Zombies}, the \textbf{Normal} setting follows standard game dynamics without any artificial delays. The \textbf{Slow} setting introduces a 5-second pause after input ingestion to buffer the model's inference latency (which typically exceeds 5 seconds), after which the environment resumes. The \textbf{Stop} setting enforces a synchronous protocol: the environment is fully paused during inference; upon action execution, the game advances for a fixed 2-second interval before the next observation is captured.

In \textit{Snake}, the settings modulate the autonomous movement mechanism. Under \textbf{Normal} and \textbf{Slow} conditions, a time-out mechanism forces the snake to advance one step in its current direction if no action is received within 10 seconds and 20 seconds, respectively. Conversely, the \textbf{Stop} setting renders the environment completely static, where the snake moves solely in response to the model's generated actions.

\section{Human Baselines}
\label{appc}
To evaluate human performance, we recruited 37 participants over the age of 18 who are fluent in English, have at least an undergraduate-level education and are computer literate. Participants were compensated fairly at or above the national minimum wage for each game played, regardless of completion. All procedures, including data collection via screen recording, were conducted with respect for participant privacy and autonomy. Participants were free to withdraw at any time without penalty. We designed this process to minimize any risks, such as fatigue, and ensure ethical treatment throughout the game play sessions.

"Human rookie" proxies the zero-shot performance of most humans to test generalization. "Human expert" refers to players in a trained state with prior game experience. Before playing, the rookie participants confirmed that they had never played the target games before and the expert participants confirmed that they are already familiar with the target games. 

We controlled the gameplay flow to align human players with the LLM as closely as possible. All participants first read our GameVerse human baseline guidance, confirming the game to play and the character to act (rookie/expert). Rookie participants reviewed the same game prompts, played once, reflected using the identical tutorial video and play again while expert participants reviewed the same prompts and play once, allowing for result comparison under maximally similar conditions. Table~\ref{tab:human-trials} shows the statistics of human trials for each game. Due to labor and time constraints, the number of trials was relatively limited for games with extensive content.

\begin{table*}[htbp]
    \centering
    \caption{The statistics of human baselines for each game.}
    \label{tab:human-trials}
    \resizebox{\textwidth}{!}{
    \begin{tabular}{l | ccc | ccc | ccc | ccc | ccc | c }
        \toprule
        \multirow{2}{*}{\textbf{Human}} & \multicolumn{3}{c|}{\textbf{Markov Grid}} & \multicolumn{3}{c}{\textbf{Non-Real-time Linear}} & \multicolumn{3}{c}{\textbf{Non-Real-time Non-Linear}} & \multicolumn{3}{c|}{\textbf{Real-time Linear}} & \multicolumn{3}{c|}{\textbf{Real-time Non-Linear}} & \multirow{2}{*}{\textbf{Total}} \\
        \cmidrule(lr){2-4} \cmidrule(lr){5-7} \cmidrule(lr){8-10} \cmidrule(lr){11-13} \cmidrule(lr){14-16} 
         & \textbf{TicTacToe} & \textbf{Baba} & \textbf{2048} & \textbf{Maze} & \textbf{AngryBird} & \textbf{Slay} & \textbf{Attorney} & \textbf{Civ VI} & \textbf{Scene} & \textbf{Snake} & \textbf{PvZ} & \textbf{Horizon} & \textbf{Metro} & \textbf{Genshin} & \textbf{RDR 2} & \\
        \midrule
        Human Rookie &10 &3 &10 &10 &5 &4 &4 &6 &4 &4 &3 &3 &8 &2 &4 & 80 \\
        Total trials &100 &3 &50 &30 &15 &4 &4 &6 &4 &12 &9 &9 &24 &2 &4 & 276 \\
        \midrule
        Human Expert &6 &6 &5 &10 &4 &7 &2 &3 &3 &3 &3 &2 &2 &2 &2 & 60 \\
        Total trials &60 &6 &25 &30 &12 &7 &2 &3 &3 &9 &9 &6 &6 &2 &2 & 182 \\
        \bottomrule
    \end{tabular}
    }
\end{table*}

\section{Game Selection Criteria}
\label{appa}

\subsection{\textit{GameVerse} Surpasses Traditional Static Benchmarks}
Compared to traditional static benchmarks, video games offer a superior testbed for evaluating VLM agents, primarily due to the following advantages. 

\textbf{1. Dynamic Interactivity and Long-Horizon Reasoning}: unlike static Q\&A tasks where the context is fixed, video games require agents to interact with a changing environment over long horizons. Agents must perceive, plan, and execute actions continuously, where current decisions significantly impact future states, mimicking real-world causal chains.

\textbf{2. Multimodal Synthesis}: Games demand a seamless synthesis of visual perception, language understanding (for instructions or narratives), and motor control. This provides a holistic evaluation of an agent's capabilities rather than testing isolated modalities.

\textbf{3. Active Learning from Failure}: Static benchmarks often use a "fire-and-forget" metric. In contrast, video games allow for an iterative learning loop where agents can observe the consequences of their actions (e.g., game over), reflect on failures, and consult tutorials to improve, mirroring human cognitive learning processes.

\textbf{4. Scalability and Diversity}: The vast array of commercial games spans diverse genres and difficulty levels, offering an infinitely scalable data source without the saturation or contamination issues common in static datasets.

\subsection{Detailed Game Selection Criteria}

To scientifically categorize the cognitive complexity of game environments, we analyze them across three core dynamic dimensions:

\textbf{Image Structure (Spatial Complexity)}: This dimension measures the complexity of the visual state space.
    \begin{itemize}
        \item \textit{Grid/Low-Fidelity}: The visual state is discrete and symbolic (e.g., Chess, 2048), where perception is simplified to identifying grid occupancy.
        \item \textit{Non-Grid/High-Fidelity}: The environment involves continuous, high-resolution visual inputs (e.g., Red Dead Redemption 2), requiring advanced computer vision capabilities to handle occlusion, lighting changes, and 3D spatial reasoning.
    \end{itemize}
    
\textbf{Temporal Dynamics (Time Constraint)}: This dimension defines the requirement for inference speed and reaction time.
    \begin{itemize}
        \item \textit{Non-Real-Time (Turn-Based)}: The environment pauses for the agent (e.g., Civilization VI), allowing unlimited time for reasoning and planning.
        \item \textit{Real-Time}: The environment evolves continuously (e.g., Forza Horizon 5). Agents must make decisions within strict latency bounds (e.g., milliseconds), testing their ability to process information under pressure.
    \end{itemize}
    
\textbf{Causal Linearity (Narrative/State Linearity)}: This dimension evaluates the predictability and branching factor of the causal structure.
    \begin{itemize}
        \item \textit{Linear}: The game follows a fixed progression path with singular objectives (e.g., Super Mario). The causal link between action and outcome is direct and predictable.
        \item \textit{Non-Linear (Open-Ended)}: The game features branching narratives, open worlds, or emergent gameplay (e.g., Minecraft, Genshin Impact). Agents must handle ambiguity, set their own sub-goals, and adapt to complex, multi-cause-multi-effect systems.
    \end{itemize}

The concrete difficulty tiers of each game is determined by its game factors. Table \ref{tab:game-factors} presents a detailed comparison of the 15 selected games across key game factors.

\begin{table}[htbp]
    \centering
    \caption{Analysis of game factors across GameVerse. The factors are evaluated on a \textbf{5-point scale}, where \textbf{1 represents easy} and \textbf{5 represents hard}.}
    \label{tab:game-factors}
    \resizebox{\textwidth}{!}{
    \begin{tabular}{l | ccc | ccc | ccc | ccc | ccc }
        \toprule
        \multirow{2}{*}{\textbf{Game Factors}} & \multicolumn{3}{c|}{\textbf{Markov Grid}} & \multicolumn{3}{c|}{\textbf{Non-Real-time Linear}} & \multicolumn{3}{c}{\textbf{Non-Real-time Non-Linear}} & \multicolumn{3}{c}{\textbf{Real-time Linear}} & \multicolumn{3}{c|}{\textbf{Real-time Non-Linear}} \\
        \cmidrule(lr){2-4} \cmidrule(lr){5-7} \cmidrule(lr){8-10} \cmidrule(lr){11-13} \cmidrule(lr){14-16} 
         & \textbf{TicTacToe} & \textbf{Baba} & \textbf{2048} & \textbf{Maze} & \textbf{AngryBird} & \textbf{Slay} & \textbf{Attorney} & \textbf{Civ VI} & \textbf{Scene} & \textbf{Snake} & \textbf{PvZ} & \textbf{Horizon} & \textbf{Metro} & \textbf{Genshin} & \textbf{RDR 2} \\
        \midrule
        Env. Complexity &1 &2 &2 &1 &3 &4 &4 &5 &5 &2 &4 &4 &3 &5 &5 \\
        Action Space    &1 &1 &1 &1 &3 &3 &3 &5 &5 &1 &2 &3 &4 &5 &5 \\
        Rule Difficulty &1 &2 &3 &1 &2 &4 &3 &5 &5 &2 &3 &3 &4 &5 &5 \\
        Reaction Time   &1 &1 &1 &1 &1 &1 &1 &1 &1 &2 &3 &5 &4 &4 &5 \\
        Content Scale   &1 &2 &3 &1 &1 &3 &3 &3 &5 &1 &3 &3 &3 &4 &5 \\
        \midrule
        \textbf{Average} &1.0 &1.6 &2.0 &1.0 &2.0 &3.0 &2.8 &3.8 &4.2 &1.6 &3.0 &3.6 &3.6 &4.6 &5.0 \\
        \textbf{Difficulty} &Easy &Medium &Hard &Easy &Medium &Hard &Easy &Medium &Hard &Easy &Medium &Hard &Easy &Medium &Hard \\
        \bottomrule
    \end{tabular}
    }
\end{table}

\subsection{Detailed Game Introduction}
\textbf{(a) Tic-Tac-Toe} is a fundamental zero-sum strategy game played on a grid, representing the simplest form of adversarial reasoning. The objective is to achieve a line of three markers while blocking the opponent. Despite its simplicity, it serves as a baseline for testing an agent's basic \textbf{logical reasoning} and understanding of \textbf{turn-taking dynamics} in a fully observable, deterministic environment.

\textbf{(b) Baba Is You}~\cite{babaisyou46} is a highly innovative puzzle game where the "rules" of the world are physical objects that can be manipulated. The environment consists of a grid containing objects and text blocks that define logic. The core challenge lies in \textbf{out-of-the-box thinking} and \textbf{rule manipulation}. Agents must demonstrate the ability to dynamically reprogram the environment's logic to solve spatial puzzles, testing high-level abstract reasoning and generalization.

\textbf{(c) 2048} is a mathematical sliding block puzzle where players merge numbered tiles to reach higher values. The environment is stochastic due to the random appearance of new tiles. This game tests an agent's ability to perform \textbf{lookahead planning} and manage \textbf{spatial resource constraints}. It requires balancing greedy immediate rewards with long-term board management to prevent gridlock.

\textbf{(d) Maze} is a classic navigation task within a labyrinth structure. The objective is straightforward: navigate from a starting point to a designated exit. This game isolates and tests the agent's \textbf{spatial navigation} and \textbf{pathfinding algorithms} (such as BFS/DFS equivalent reasoning) in a clean, noise-free 2D environment without external disturbances or enemies.

\textbf{(e) Angry Birds}~\cite{angrybirds45} is a physics-based puzzle game that requires players to launch projectiles to destroy structures. The environment simulates 2D physics including gravity, momentum, and collision. The primary capability tested here is \textbf{physics understanding}. Agents must infer physical properties (mass, stability) from visual inputs and perform \textbf{trajectory planning} to cause maximum structural damage with limited resources.

\textbf{(f) Slay the Spire}~\cite{slaythespire44} is a strategy game combining with roguelike progression. The agent must climb a spire by battling enemies using a deck of cards that evolves over time. This game presents a complex challenge in \textbf{stochastic planning} and \textbf{resource management}. Agents are tested on their ability to adapt to random card draws, optimize a deck for long-term synergy, and make trade-off decisions between immediate survival and future power.

\textbf{(g) Ace Attorney}~\cite{ace43} is a narrative-heavy visual novel centered on courtroom simulations. The player acts as a defense lawyer, investigating crime scenes and cross-examining witnesses. This environment heavily focuses on \textbf{Natural Language Understanding} and \textbf{logic deduction}. Agents must detect contradictions between text testimony and visual evidence, requiring deep context retention and the ability to infer truth from conflicting information.

\textbf{(h) Civilization VI}~\cite{civil42} is a grand strategy game of immense complexity. Players guide a civilization from the Stone Age to the Information Age. The environment involves exploring a map hidden by fog of war, managing economies, and engaging in diplomacy. This game is the ultimate test for \textbf{hierarchical planning} and \textbf{strategic reasoning}. Agents must balance competing priorities over thousands of steps, handling a massive state space with long-term horizons.

\textbf{(i) Scene Investigator}~\cite{Scene41} is a deductive detective game set in detailed 3D crime scenes. The objective is to observe the environment meticulously to reconstruct past events. Unlike standard object detection, this game tests \textbf{visual deduction} and \textbf{causal inference}. Agents must notice subtle visual cues and link them to form a coherent narrative explanation of the crime.

\textbf{(j) Snake}~\cite{snake40} is a real-time arcade game requiring the player to control a growing line. The environment is a contained grid where the agent must react quickly to consume food while avoiding collisions. This game primarily tests \textbf{reflexes} and \textbf{spatial planning under pressure}. As the snake grows, the free space diminishes, requiring the agent to plan efficient paths to avoid trapping itself in a rapidly closing environment.

\textbf{(k) Plants vs. Zombies}~\cite{PvZ39} is a strategic tower defense game. Players must defend a home from incoming waves of enemies by placing plants with specific functions. The environment requires managing a generated resource and spatial positioning. It tests \textbf{tactical positioning} and \textbf{real-time resource allocation}. Agents must recognize enemy types and counter them with the appropriate unit composition while optimizing the economy.

\textbf{(l) Forza Horizon 5}~\cite{horizon38} is a photorealistic racing simulator. The environment offers high-fidelity 3D visuals and realistic vehicle dynamics. The objective is to drive vehicles efficiently across diverse terrains. This game serves as a benchmark for \textbf{continuous control} and \textbf{high-speed visual perception}. Agents must interpret complex visual scenes (blur, lighting, geometry) in real-time and map them to precise steering and throttle commands.

\textbf{(m) Mini Metro}~\cite{minimetro37} is a strategy simulation about designing a subway map. The environment is an abstract, evolving graph where nodes appear randomly. The agent must connect them with limited lines to transport passengers. This game tests \textbf{graph optimization} and \textbf{dynamic network management}. Agents must constantly redesign the network topology to alleviate congestion, demonstrating adaptability to stochastic demand surges.

\textbf{(n) Genshin Impact}~\cite{Genshin36} is a vast open-world action RPG with an elemental combat system. The environment is rich with traversable terrain, puzzles, and enemies. The game tests \textbf{embodied exploration} and \textbf{multi-tasking}. Agents are evaluated on their ability to navigate complex 3D topography, understand elemental interaction mechanics, and follow multi-step quest instructions in a non-linear world.

\textbf{(o) Red Dead Redemption 2}~\cite{RDR235} is a pinnacle of open-world simulation, known for its narrative depth and environmental realism. The setting is a detailed recreation of the American frontier. This game tests \textbf{comprehensive general intelligence}, including social interactions, realistic physics navigation, and narrative comprehension. Agents must operate within a living world where characters and animals react realistically, requiring a high degree of common sense and contextual understanding.

\section{Game Environment Details and Extended Analysis.}
\label{appd}
\subsection{Tic-Tac-Toe}
\subsubsection*{D.1.1 Game Description for Tic-Tac-Toe}
\begin{wrapfigure}{r}{0.4\textwidth}
\vspace{-70pt}
    \centering
    \includegraphics[width=\linewidth]{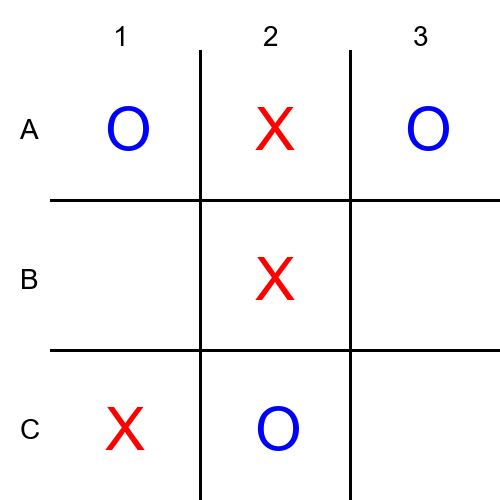}
    \caption{Screenshot of Tic-Tac-Toe}
    \label{fig:tictactoetitle}
\end{wrapfigure}

\textbf{Game Environment.}
Tic-Tac-Toe is a classic zero-sum strategy game played on a grid. Two players take turns marking spaces with X and O to achieve a line of three in a horizontal, vertical, or diagonal row. The first player to get three in a row wins. If the grid is filled and no one has three in a row, the game is a draw. Tic-Tac-Toe is a "solved game." This means that mathematically, the outcome is already known if both players play perfectly. If both players make the best possible moves every turn, the game will always end in a draw. It is impossible to win against a perfect opponent; you can only hope they make a mistake. We use the MinMax algorithm ~\cite{minmax53} as the opponent of agent. MinMax is a decision rule used in artificial intelligence and game theory for two-player games like Tic-Tac-Toe or Chess. It helps a computer decide the perfect move.

(1) \textbf{Game state.} Fully observable $3 \times 3$ grid represented symbolically or visually.

(2) \textbf{Semantic action space.} Discrete; placing a mark on an empty cell coordinates $(x, y), x\in[1,3],y\in[1,3]$.

(3) \textbf{Main GUI action space.} Mouse Input, especially \textit{CLICK}.

(4) \textbf{Evaluation task.} Play against a MinMax algorithm; performance is measured by the draw rate in 20 trials, where the agent plays as X and O for 10 trials, respectively. Formally, the normalized score is defined as:
\begin{equation*}
S = \begin{cases} 
100 & \text{Draw} \\
\frac{n}{5} \times 100 & \text{Lose (X First)} \\
\frac{n}{4} \times 100 & \text{Lose (O Second)}
\end{cases}, \quad \text{where } S \text{ is the process score, and } n \text{ is the number of pieces.}
\end{equation*}

(5) \textbf{Expert video content.} The expert video is a skill tutorial. This expert tutorial provides strategic guidance on Tic-tac-toe, detailing optimal moves for both first and second players to ensure a non-losing outcome.

\subsubsection*{D.1.2 Game Prompt for Tic-Tac-Toe}
\begin{figure}[p]
    \centering
        \includegraphics[height=0.95\textheight]{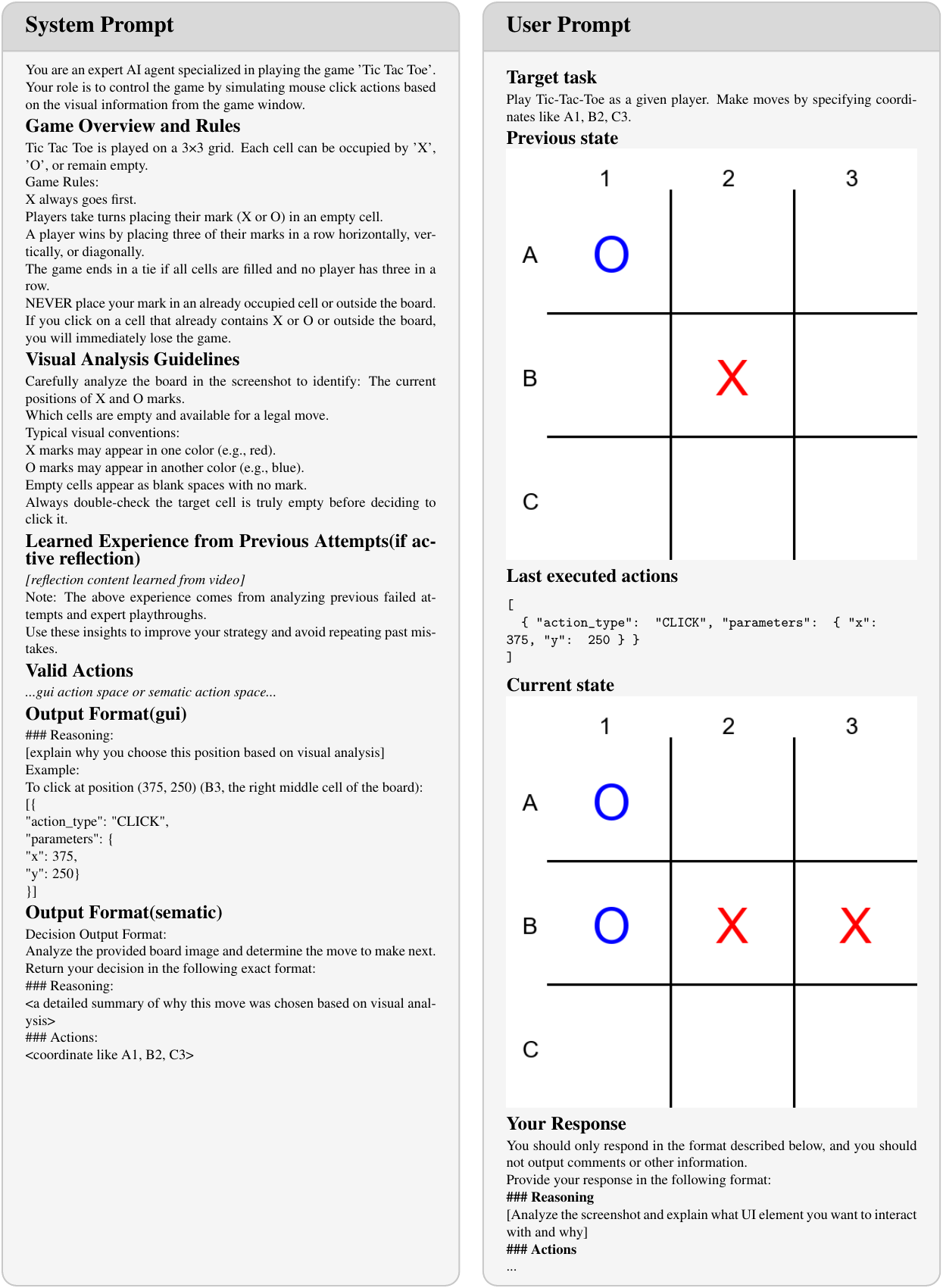}
    \caption{Tic-Tac-Toe prompt}
    \label{fig:Tic-Tac-Toe prompt}
\end{figure}

Our implementation of Tic-Tac-Toe uses zero-shot agent. We provide the full structure of our prompts in Figure~\ref{fig:Tic-Tac-Toe prompt}.
,

\subsubsection*{D.1.3 Detailed Analysis for Tic-Tac-Toe}
\begin{figure}[ht]
    \centering
        \includegraphics[width=\linewidth]{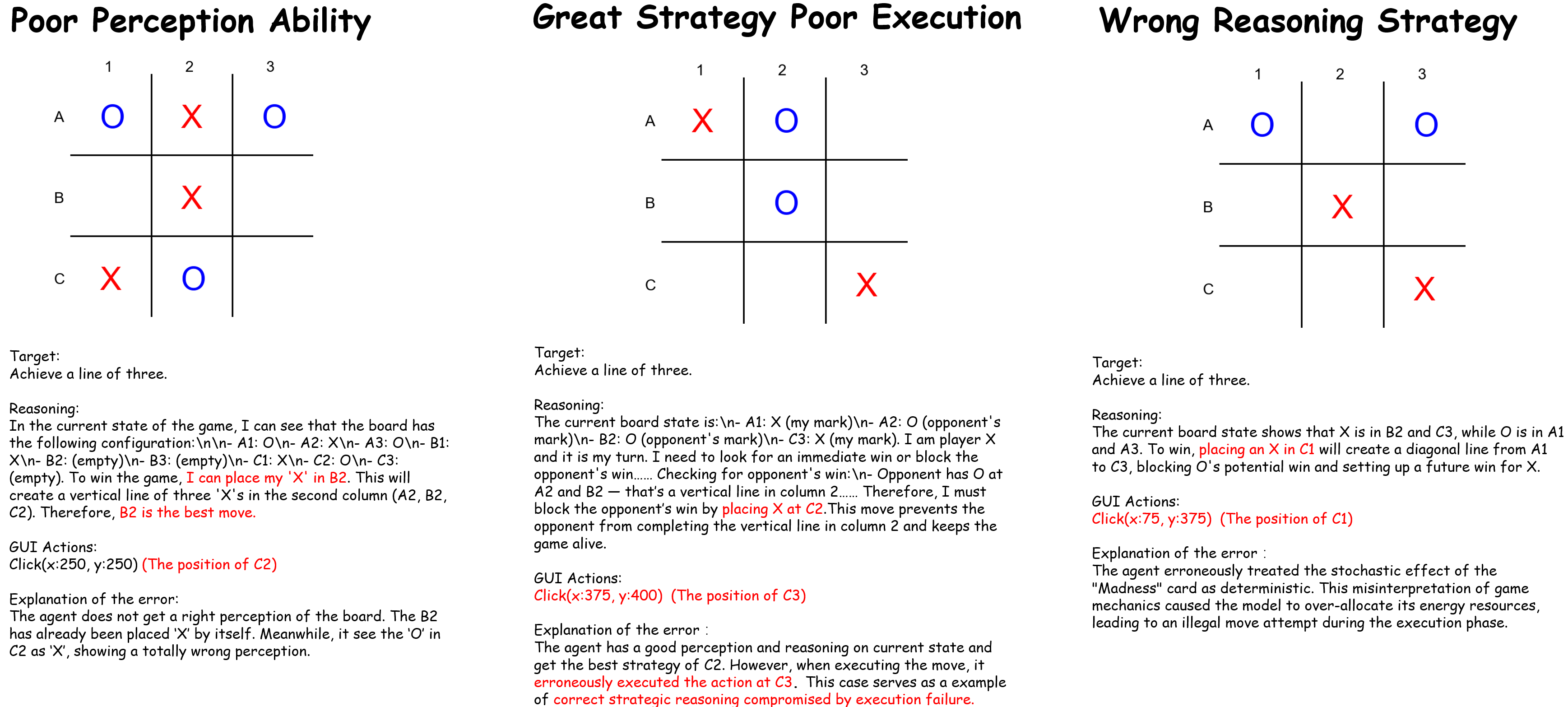}
    \caption{Tic-tac-toe typical errors}
    \label{fig:tictactoe typical errors}
    \vspace{-10pt}
\end{figure}

Despite the deceptive simplicity of Tic-Tac-Toe, our experimental results reveal fundamental deficiencies in the visual-reasoning pipeline of VLMs. While the state space is relatively small, the strict spatial rigidity of the grid highlights critical failures in grounding and rule adherence. As illustrated in Figure~\ref{fig:tictactoe typical errors}, errors generally manifest in three distinct categories: Perceptual Hallucination, Semantic-Execution Gap, and Rule Fabrication.

\begin{table}[htbp]
    \centering  
    \begin{tabular}{lcccc}
        \toprule
        \textbf{Model} & \textbf{GUI} & \textbf{GUI VR.} & \textbf{Semantic} & \textbf{Semantic VR.} \\
        \midrule
          Qwen3-VL-8B      & $53.0 \pm 15.0$ & $52.0 \pm 15.0$ & $52.0 \pm 14.0$ & $67.0 \pm 13.0$ \\

        Qwen3-VL-32B     & $71.0 \pm 17.0$ & $63.0 \pm 7.0$ & $-$             & $-$             \\

        GPT-4o-mini      & $34.0 \pm 17.0$ & $43.0 \pm 12.0$ & $-$             & $-$             \\

        GPT-4o           & $59.0 \pm 4.0$ & $64.0 \pm 10.0$ & $67.0 \pm 10.0$ & $64.0 \pm 8.0$ \\

        Seed-1.8         & $\mathbf{92.0 \pm 3.0}$ & $\mathbf{100.0 \pm 0.0}$ & $-$             & $-$             \\

        Gemini-2.5-Flash & $88.0 \pm 16.0$ & $95.0 \pm 12.0$ & $94.0 \pm 9.0$ & $\mathbf{100.0 \pm 0.0}$ \\

        Gemini-2.5-Pro   & $90.0 \pm 9.0$ & $\mathbf{100.0 \pm 0.0}$ & $\mathbf{100.0 \pm 0.0}$ & $\mathbf{100.0 \pm 0.0}$ \\
        \bottomrule
    \end{tabular}
    \vspace{10pt}
     \caption{Tic-Tac-Toe Raw Scores}
    \label{tab:tic_tac_toe_raw}
    \vspace{-15pt}
\end{table}

\textbf{Perceptual Hallucination.} \\
A primary failure mode, labeled as "Poor Perception Ability", involves the agent’s inability to accurately ground visual tokens into a coherent internal state. In the depicted instance, the model correctly identifies the grid structure but hallucinates the occupancy status of critical cells. Specifically, the reasoning trace explicitly claims that cell B2 is "empty" and viable for a winning move, directly contradicting the visual evidence where B2 is already occupied by the agent's own mark ('X'). This suggests that even in low-complexity visual environments, VLMs struggle to maintain object permanence and state consistency, leading to "confidently wrong" decisions based on flawed premises.

\textbf{Semantic-Execution Gap}. \\
We observe a profound disconnection between strategic reasoning and motor control, mirroring the "GUI-to-Semantic Gap" observed in complex environments. As demonstrated in the "Great Strategy Poor Execution" case, the agent successfully identifies the opponent's threat (a vertical line in column 2) and correctly deduces the optimal blocking strategy ("placing X at C2"). However, the downstream execution module fails to map this semantic intent to the correct pixel coordinates, executing a click at C3 instead. This "correct thought, wrong action" phenomenon indicates that the reasoning capabilities of large models are often bottlenecked by imprecise visual-motor alignment, where the abstract concept of a grid coordinate fails to translate into accurate spatial localization.

\textbf{Rule Fabrication and Logical Inconsistency.} \\
The third failure mode, "Wrong Reasoning Strategy," highlights a breakdown in logical coherence where the agent hallucinates game mechanics to justify suboptimal moves. In the example, the agent proposes placing a mark at C1 to "create a diagonal line from A1 to C3." This reasoning is fundamentally flawed on two levels: first, A1 is occupied by the opponent ('O'), making a connection impossible; second, the agent conflates the opponent's pieces with its own to fabricate a win condition. These errors in Tic-Tac-Toe suggest that VLMs may struggle with basic rule synthesis, occasionally reverting to generative hallucinations rather than adhering to the rigid logic of deterministic games.

\subsection{Baba Is You} 
\subsubsection*{D.2.1 Game Description for Baba Is You}

\begin{wrapfigure}{r}{0.55\textwidth}
\vspace{-20pt}
    \centering
    \includegraphics[width=\linewidth]{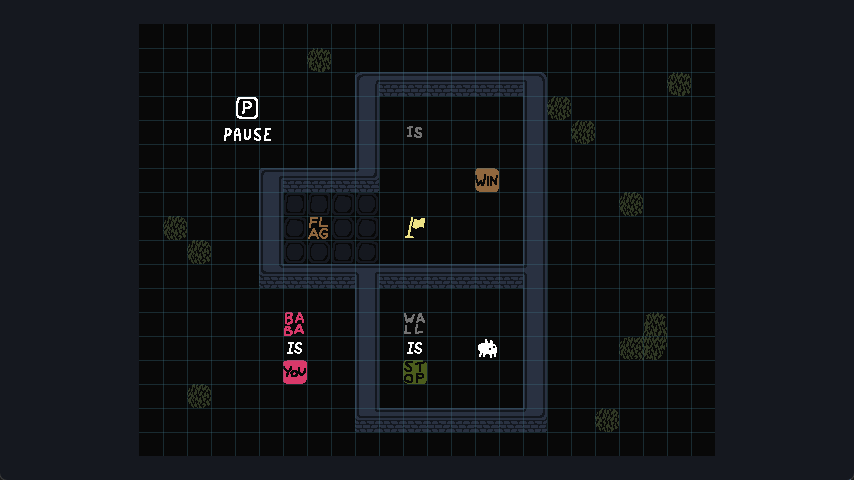}
    \caption{Screenshot of Baba Is You}
    \label{fig:baba is you game title}
\end{wrapfigure}

\textbf{Game Environments.} 
Baba Is You ~\cite{babaisyou46}  is a highly innovative puzzle game where the ``rules'' of the world are represented as physical objects that can be manipulated by the player. The environment consists of a grid containing objects and text blocks that define the current logic (e.g., "Wall is Stop"). The core challenge lies in out-of-the-box thinking and rule manipulation. Agents must demonstrate the ability to dynamically reprogram the environment's logic to solve spatial puzzles, testing high-level abstract reasoning and generalization. Unlike static puzzles, the solution often requires the agent to fundamentally alter the governing laws of the game world, pushing the boundaries of semantic understanding and causal reasoning in a low-fidelity grid setting.

\noindent\textbf{(1) Game state.} Discrete and symbolic grid (Grid/2D) containing objects and rule blocks.

\noindent\textbf{(2) Main GUI action space.} Keyboard Input, especially \textit{KeyPress}.

\noindent\textbf{(3) Evaluation task.} Experiments were conducted on Level 0 and Level 1 of the Baba Is You game to evaluate agent performance. We impose a maximum limit of 30 steps per episode. The predefined milestones for Level 0 are specified as: Initiate Movement and Object Interaction, Clear Obstacles, Trigger Win Condition, and Level Completion. For Level 1, the evaluation milestones include: Initial Movement, Rule Break: Wall is Stop, Crossing the Obstacle, Create Winning Condition, and Level Completion. The performance of the agent was quantified by the count of milestones successfully accomplished in Level 0 and Level 1, where the final performance value was calculated as the mean of at least three independent trial runs. The evaluation metric was normalized by the total number of milestones across the two levels (i.e., 9), and the formal definition of the normalized score is given as follows:
\begin{equation*}
S_{\text{norm}} = \frac{\frac{1}{n}\sum_{i=1}^n M_i}{M_{\text{total}}} \times 100
\end{equation*}
where $M_i$ denotes the number of milestones accomplished in the $i$-th trial, $n$ represents the total number of independent trials, and $M_{\text{total}}=9$ is the total number of predefined milestones for Level 0 and Level 1.

\noindent\textbf{(4) Expert video content.} The expert video is a skill-oriented instructional tutorial that specifically analyzes and rectifies the cognitive misconceptions in puzzle-solving for the relevant level of Baba Is You, including regarding walls as permanent barriers, moving rule components in a reverse manner, and stacking text blocks on their corresponding physical objects. This video conducts an in-depth dissection of the core level mechanics and the logical principles of rule manipulation, covering key elements such as rule modification, component movement and operation planning. It thereby demonstrates the necessity of adopting scientific puzzle-solving strategies, and guides the agent to make decisions based on the intrinsic nature of rules and operational pre-judgment, so as to avoid irreversible errors.

\subsubsection*{D.2.2 Game Prompt For Baba Is You}
\begin{figure}[p]
    \centering
        \includegraphics[height=0.95\textheight]{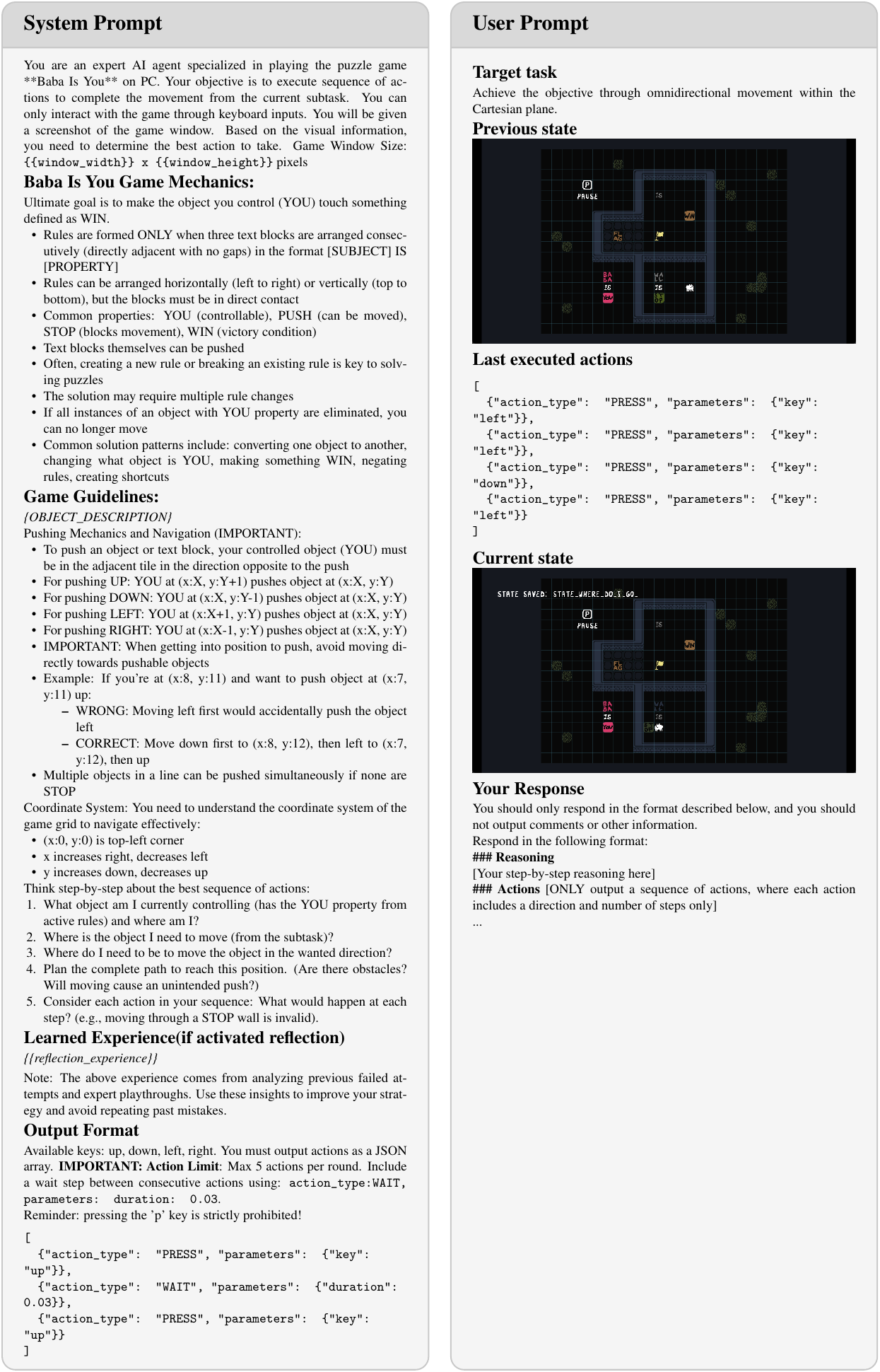}
    \caption{Baba Is You prompt}
    \label{fig:BabaIsYou prompt}
\end{figure}

Our implementation of Baba is you uses zero-shot agent. We provide the full structure of our prompts in Figure~\ref{fig:BabaIsYou prompt}.

\subsubsection*{D.2.3  Detailed Analysis For Baba Is You}

\begin{wrapfigure}{r}{0.5\textwidth} 
  \centering
  \vspace{-5pt} 
  \begin{tabular}{lcc}
    \toprule
    \textbf{Model} & \textbf{GUI} & \textbf{GUI VR.} \\
    \midrule
      Gemini-2.5-Flash      & $80.0 \pm 0.0$      & $80.0 \pm 0.0$\\

    Gemini-2.5-Pro        & $80.0 \pm 0.0$ & $80.0 \pm 0.0$  \\

    Qwen3-VL-8B           & $60.0 \pm 0.0$      & $70.0 \pm 8.2$\\

    Qwen3-VL-32B          & $73.3 \pm 9.4$ & $80.0 \pm 0.0$\\

    GPT-4o                & $61.0 \pm 1.4$      & $60.0 \pm 0.0$\\

    GPT-4o-mini           & $60.0 \pm 0.0$      & $66.7 \pm 11.5$\\

    Seed-1.8       & $80.0 \pm 0.0$& $80.0 \pm 0.0$             \\
    \bottomrule
  \end{tabular}
  \caption{Raw scores in Baba Is You}
  \label{tab:BabaIsYou scores}
  \vspace{-10pt} 
\end{wrapfigure}

\begin{figure}[ht]
    \centering
        \includegraphics[width=\linewidth]{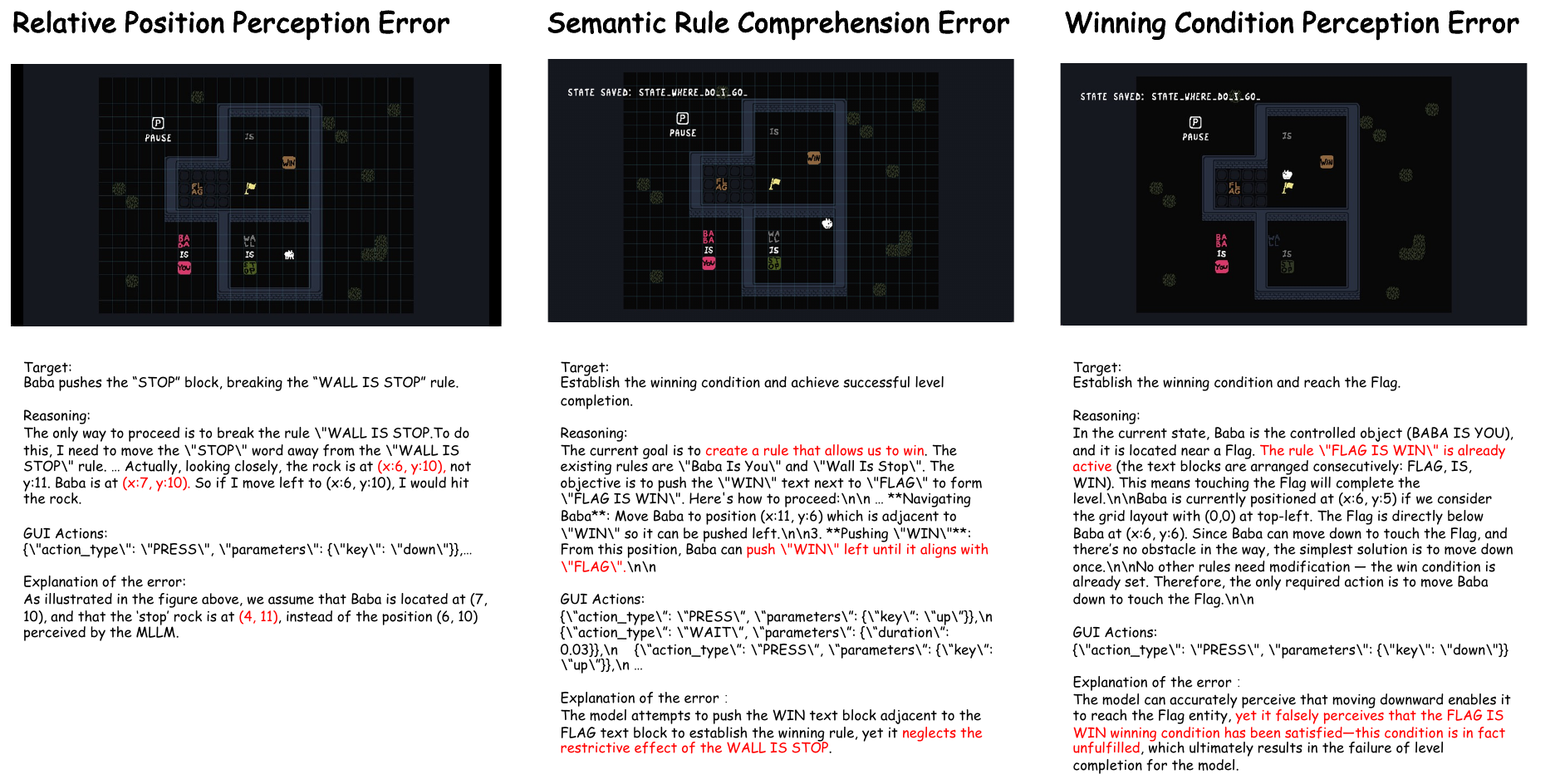}
    \caption{Typical errors in Baba Is You}
    \label{fig:BabaIsYou errors}
\end{figure}

Experimental results for Baba Is You demonstrate that the performance of agents exhibits a notable ceiling, which is highly correlated with the models’ capabilities in visual localization and semantic comprehension. Visual localization constitutes a critical bottleneck for VLMs in playing this game: only models with robust localization capabilities (e.g., Qwen3-VL-32B) can occasionally achieve accurate relative localization, while all other models suffer from varying degrees of hallucination in relative position localization, thereby failing to generate valid action commands and complete the corresponding tasks. In Level 1, breaking the semantic rule of WALL IS STOP serves as the primary critical benchmark for evaluating the reasoning capabilities of VLM. Models with advanced reasoning capabilities (e.g., Gemini-2.5-Pro, Gemini-2.5-Flash, Seed-1.8) can accurately comprehend this rule and disable the stop rule for walls, whereas less capable models (e.g., GPT-4o, GPT-4o-mini, Qwen3-VL-8B) are unable to grasp the underlying logic of the rule, which renders them unable to proceed with the game. The milestone of establishing a winning condition in Level 1 poses an arduous challenge for all current models; across all conducted tests, no model successfully established a valid winning condition, and thus failed to achieve the final victory. Two typical error types documented in  \cref{fig:BabaIsYou errors}—relative position perception error and semantic rule comprehension error—serve as representative examples of the aforementioned deficiencies.

\textbf{Spatial Localization Bias}\\ 
We observe a pronounced "hallucination" phenomenon regarding localization precision in current mainstream models. Although high-reasoning models (e.g., Gemini-2.5-Pro, Seed-1.8) exhibit sophisticated capabilities in parsing semantic rules and "Sokoban" mechanics—enabling them to generate strategic path planning and execution sequences based on perceived spatial data—they are consistently hindered by the fine-grained localization precision required by the game's grid map. Consequently, these models typically stall at the fourth milestone of Level 1 ("Constructing Victory Rules"). This suggests that spatial localization accuracy has become a critical bottleneck preventing mainstream models from achieving full level completion.

\textbf{The Gap in Semantic Rule Comprehension}\\
In the context of Baba Is You, the ability to comprehend semantic rules serves as a primary differentiator for model performance. In Level 1, models with superior reasoning faculties (e.g., Gemini-2.5-Pro, Seed-1.8) successfully decode the "WALL IS STOP" logic and proactively devise plans to "break" the rule to advance the game state. Conversely, models with weaker reasoning capabilities (e.g., GPT-4o, GPT-4o-mini) demonstrate inconsistent performance: they occasionally grasp the rule but lack the strategic motivation to manipulate it; in other instances, they fail to recognize the physical constraints of the rule entirely, attempting to bypass walls directly to reach the flag, which results in the agent (Baba) remaining obstructed by environmental objects.

\newpage 

\textbf{Video Reflection Analysis}

\begin{wrapfigure}{r}{0.5\textwidth} 
    \centering
    \small
    \includegraphics[width=\linewidth]{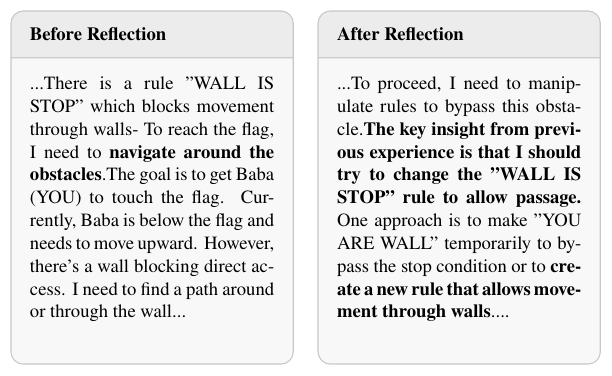}
    \caption{Qwen3-VL-8B reflection improvement example}
    \label{fig:Qwen3-VL-8B_reflection Baba}
\end{wrapfigure}

Experimental results indicate that while reflection mechanisms can enhance model reasoning, their efficacy is fundamentally constrained by the interplay between a model's baseline intelligence and its spatial localization precision. For state-of-the-art models such as Gemini-2.5-Pro and Seed-1.8, the comparative analysis of expert demonstrations and self-failure videos yields negligible gains, suggesting that their primary bottleneck has shifted from semantic comprehension to the precise spatial localization of grid-map entities. In contrast, for mid-tier models like Qwen3-VL-32B, video reflection effectively bridges the gap in rule comprehension, yet localization inaccuracies continue to impede successful level completion. Furthermore, in models with weaker reasoning capabilities like Qwen3-VL-8B, GPT-4o, and GPT-4o-mini, the impact of reflection is bifurcated by their inherent spatial faculties: Qwen3-VL-8B leverages its superior localization to achieve a marked performance leap, whereas the GPT-4o series—hampered by profound deficiencies in spatial perception—fails to translate rule comprehension into effective execution, resulting in only marginal score increments or stochastic fluctuations.

In summary, the primary utility of the reflection mechanism lies in its ability to augment the logical reasoning faculties of lower-performing models—within the constraints of their inherent spatial localization capabilities—thereby enhancing their overall performance in Baba Is You.

\subsection{2048}
\subsubsection*{D.3.1 Game Description for 2048}

\textbf{Game Environments.}  2048 is a mathematical sliding block puzzle where players merge numbered tiles to reach higher values. The environment is stochastic due to the random appearance of new tiles after every move. This game tests an agent's ability to perform lookahead planning and manage spatial resource constraints. It requires balancing greedy immediate rewards with long-term board management to prevent gridlock. As a benchmark, 2048 evaluates whether an agent can maintain a strategic structure (such as a snake-chain formation) over a long horizon. The game's probabilistic nature forces the agent to adapt its strategy dynamically, making it a strong test of mathematical reasoning and spatial foresight within a confined grid.

\begin{wrapfigure}{r}{0.4\textwidth}
    \centering
    \includegraphics[width=\linewidth]{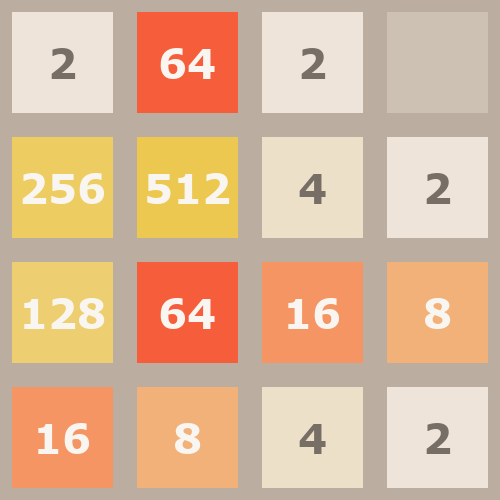}
    \caption{Screenshot of 2048}
    \label{fig:2048title}
\end{wrapfigure}

\noindent\textbf{(1) Game state.} Discrete and symbolic grid where perception is simplified to identifying grid occupancy and the specific numerical values on tiles.

\noindent\textbf{(2) Main GUI action space.} Keyboard Input,  especially \textit{KeyPress} direction keys

\noindent\textbf{(3) Evaluation task.} Play the game of \textit{2048} for multiple independent episodes. We impose a maximum limit of 1000 steps per episode. The agent’s performance is quantified by the normalized score achieved in each episode, defined as the ratio between the accumulated merge score and a reference score corresponding to successfully forming the 2048 tile. The final evaluation score is computed as the average normalized score over all test runs:
\begin{equation*}
S_{\text{norm}} = \frac{1}{N_{\text{test}}}\sum_{i=1}^{N_{\text{test}}} \frac{S_i}{S_{\text{ref}}} \times 100,
\end{equation*}

where $S_i$ denotes the merge score obtained in the $i$-th test run, $N_{\text{test}}$ denotes the number of test episodes, and $S_{\text{ref}} = 20000$ represents the approximate score required to achieve the 2048 tile. The resulting score is clipped to the range $[0,100]$.

\noindent\textbf{(4) Expert video content.} The expert video shows a session of \textit{2048} gameplay, highlighting stable board management via corner anchoring and monotonic tile ordering. It demonstrates risk-aware move selection under stochastic tile spawning, guiding agents to prioritize long-term board stability over greedy merges.

\subsubsection*{D.3.2 Game Prompt for 2048}
\begin{figure}[p]
    \centering
        \includegraphics[height=0.95\textheight]{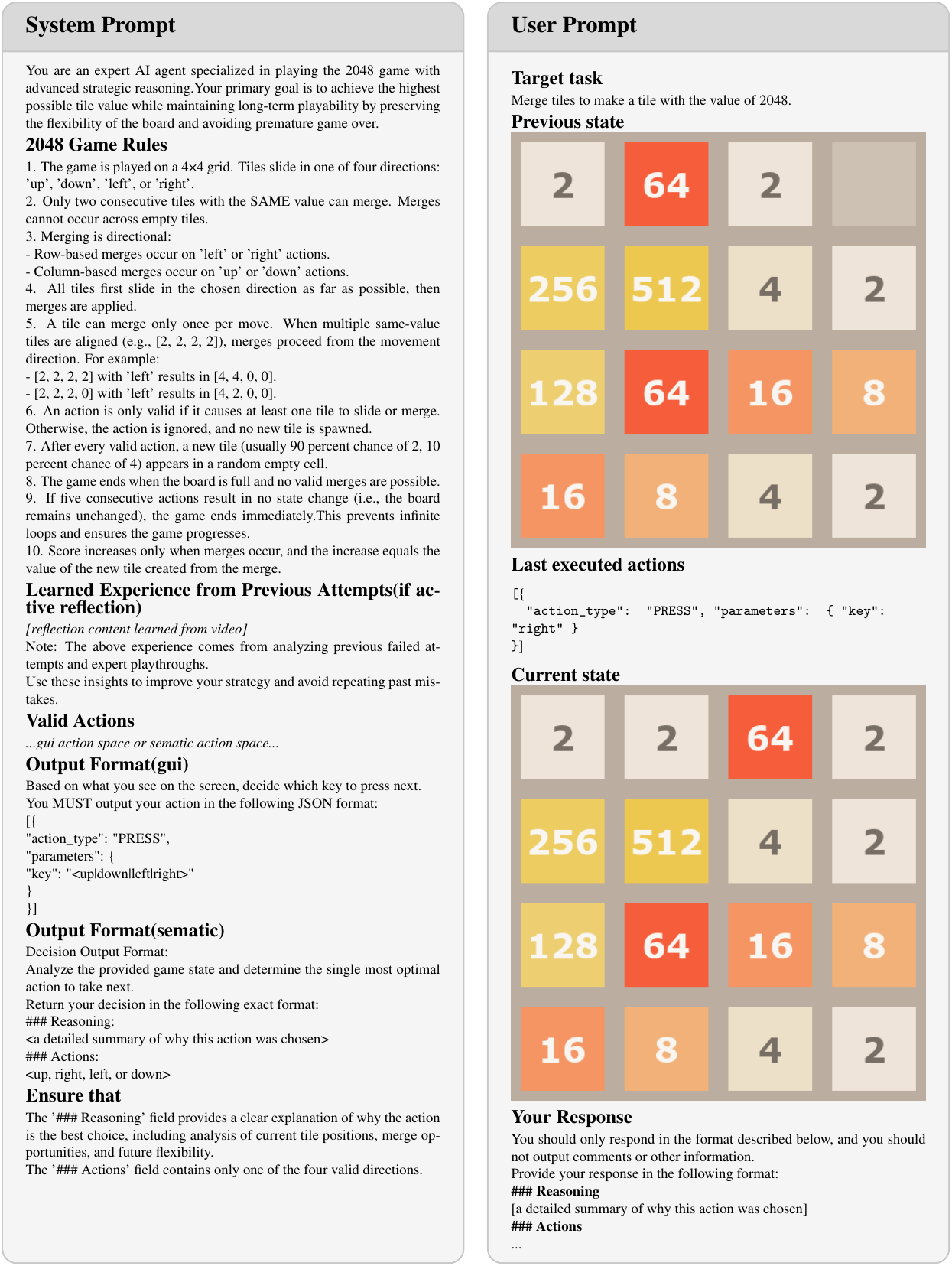}
    \caption{2048 prompt}
    \label{fig:2048 prompt}
\end{figure}

Our implementation of 2048 uses zero-shot agent. We provide the full structure of our prompts in Figure~\ref{fig:2048 prompt}.

\subsubsection*{D.3.3 Detailed Analysis for 2048}

\begin{figure}[ht]
    \centering
        \includegraphics[width=\linewidth]{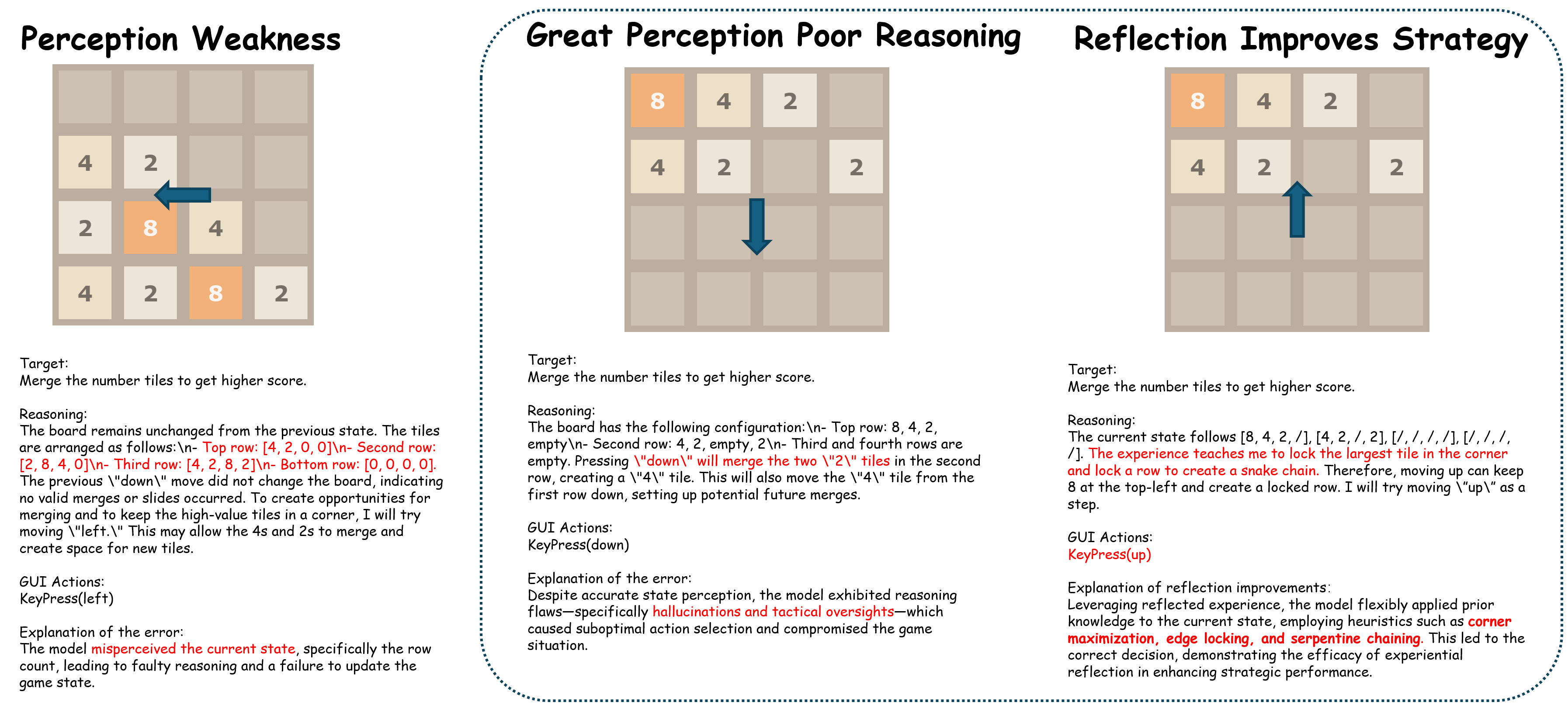}
    \caption{2048 typical errors and reflection improvements}
    \label{fig:2048 errors}
\end{figure}

The experimental results in the 2048 environment highlight a distinct dichotomy between visual state parsing and strategic planning. While the game features a discrete and deterministic sliding mechanics on a 4x4 grid, VLMs exhibit specific failure modes rooted in spatial misalignment and mechanical hallucination. As illustrated in Figure~\ref{fig:2048 errors}, errors primarily manifest as perceptual weakness and reasoning hallucination.

\begin{wrapfigure}{r}{0.5\textwidth} 
    \centering
    \begin{tabular}{lcc} 
        \toprule
        \textbf{Model} & \textbf{GUI} & \textbf{GUI VR.} \\
        \midrule
        \textbf{Qwen3-VL-8B}      & $875.6 \pm 417.9$  & $1014.9 \pm 238.8$ \\
        \textbf{Qwen3-VL-32B}     & $1273.6 \pm 636.8$ & $1213.9 \pm 736.3$ \\
        \textbf{GPT-4o-mini}      & $258.7 \pm 79.6$   & $278.6 \pm 238.8$  \\
        \textbf{GPT-4o}           & $756.2 \pm 179.1$  & $676.6 \pm 218.9$  \\
        \textbf{Seed-1.8}         & $1810.9 \pm 1034.8$& $3621.8 \pm 1671.6$\\
        \textbf{Gemini-2.5-Flash} & $1552.2 \pm 417.9$ & $2268.6 \pm 756.2$ \\
        \textbf{Gemini-2.5-pro}   & $2626.8 \pm 616.9$ & $5253.6 \pm 1353.2$\\
        \bottomrule
    \end{tabular}
    \vspace{5pt}
    \caption{2048 Raw Scores}
    \label{tab:2048_raw}
    \vspace{-15pt}
\end{wrapfigure}

\textbf{Perception Weakness}. \\
A fundamental failure mode observed is the inability to accurately encode the geometric configuration of the board. In the "Perception Weakness" instance, the agent attempts to plan a move but fails to ground its reasoning in the actual visual state. The model explicitly misidentifies the row count and the specific arrangement of tiles, hallucinating a board state that does not align with the provided screenshot. This "State Parsing Failure" leads to a cascading error where subsequent reasoning is logically sound but applied to a non-existent game state, resulting in actions that fail to optimize the actual board configuration.

\textbf{Reasoning Hallucination and Tactical Oversight}. \\
Even when the visual perception is relatively accurate, we observe a breakdown in forward simulation, labeled as "Great Perception Poor Reasoning". Here, the model correctly identifies the tiles but exhibits a "mechanical hallucination" regarding the game physics. The reasoning trace suggests that pressing "down" will merge two distinct "2" tiles in the second row; however, given the grid constraints and intervening empty spaces, this merge is physically impossible in the predicted manner. This disconnect indicates that while the model may recognize symbols (numbers), it struggles to consistently simulate the sliding mechanics that govern tile interaction, leading to tactical oversights and inefficient board manipulation.

\textbf{Video Reflection Analysis.} \\
In contrast to the purely reasoning-based approaches, the introduction of a reflection mechanism yields a quantifiable improvement in strategic depth for 2048. As demonstrated in Figure~\ref{fig:2048 errors} (right), the reflection module successfully bridges the gap between immediate tactical moves and long-term planning. The "Reflection Improves Strategy" case illustrates the model's ability to synthesize high-level heuristics from prior failures. By explicitly recalling the need to "lock the largest tile in the corner" and form a "snake chain," the agent overrides greedy, short-term merging behaviors in favor of structured play. This demonstrates that experiential reflection allows the model to abstract complex positional concepts—specifically corner maximization and edge locking—which are essential for sustaining game play at higher tile values. The transition from random merging to maintaining a monotonic chain structure suggests that reflection serves as a critical regularizer, constraining the action space to moves that preserve long-term board viability.

\newpage

\subsection{Maze}
\subsubsection*{D.4.1 Game Description for Maze}

\begin{wrapfigure}{r}{0.35\textwidth}
\vspace{-40pt}
    \centering
    \includegraphics[width=\linewidth]{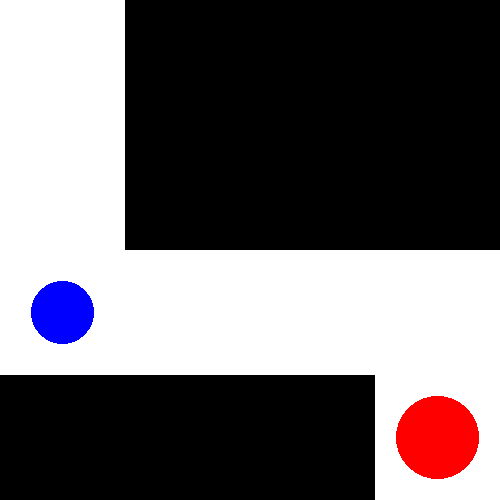}
    \caption{Screenshot of Maze}
    \label{fig:maze game title}
\end{wrapfigure}

\textbf{Game Environments.}  Maze is a classic navigation task within a labyrinth structure. The objective is straightforward: navigate from a starting point to a designated exit. This game isolates and tests the agent's spatial navigation and pathfinding algorithms (such as BFS/DFS equivalent reasoning) in a clean, noise-free 2D environment without external disturbances or enemies. It serves as a specific diagnostic tool for "Non-Real-Time Linear" capabilities, verifying if the agent can parse visual geometry and plan a viable route through obstacles. The absence of time pressure allows for a pure evaluation of the agent's visual perception of boundaries and its ability to execute a planned sequence of movements without deviation.

\noindent\textbf{(1) Game state.} A clean, noise-free 2D environment without external disturbances or enemies.

\noindent\textbf{(2) Main GUI action space.} Keyboard Input, especially \textit{KeyPress} direction keys. 

\noindent\textbf{(3) Evaluation task.} Reach the finish. We impose a maximum limit of 20 steps per episode. The agent's performance is quantified by the normalized score from distance to the finish. The score is formally defined as:

\begin{equation*}
    Score = (1 - \frac{D_{end}}{D_{begin}}) \times 100
\end{equation*}

where \(D_{begin}\) denotes the distance to the finish at the game beginning, and \(D_{end}\) denotes the distance to the finish at the game end. This normalization constrains the score to the range \([0, 199]\), where a value of 1 indicates optimal performance and 0 signifies no progress. 

\noindent\textbf{(4) Expert video content} The expert video is a skill tutorial that introduces the ideas and steps for solving a maze using the DFS method.

\subsubsection*{D.4.2 Game Prompt for Maze}

\begin{wrapfigure}{r}{0.5\textwidth}
    \vspace{-40pt}
    \includegraphics[width=\linewidth]{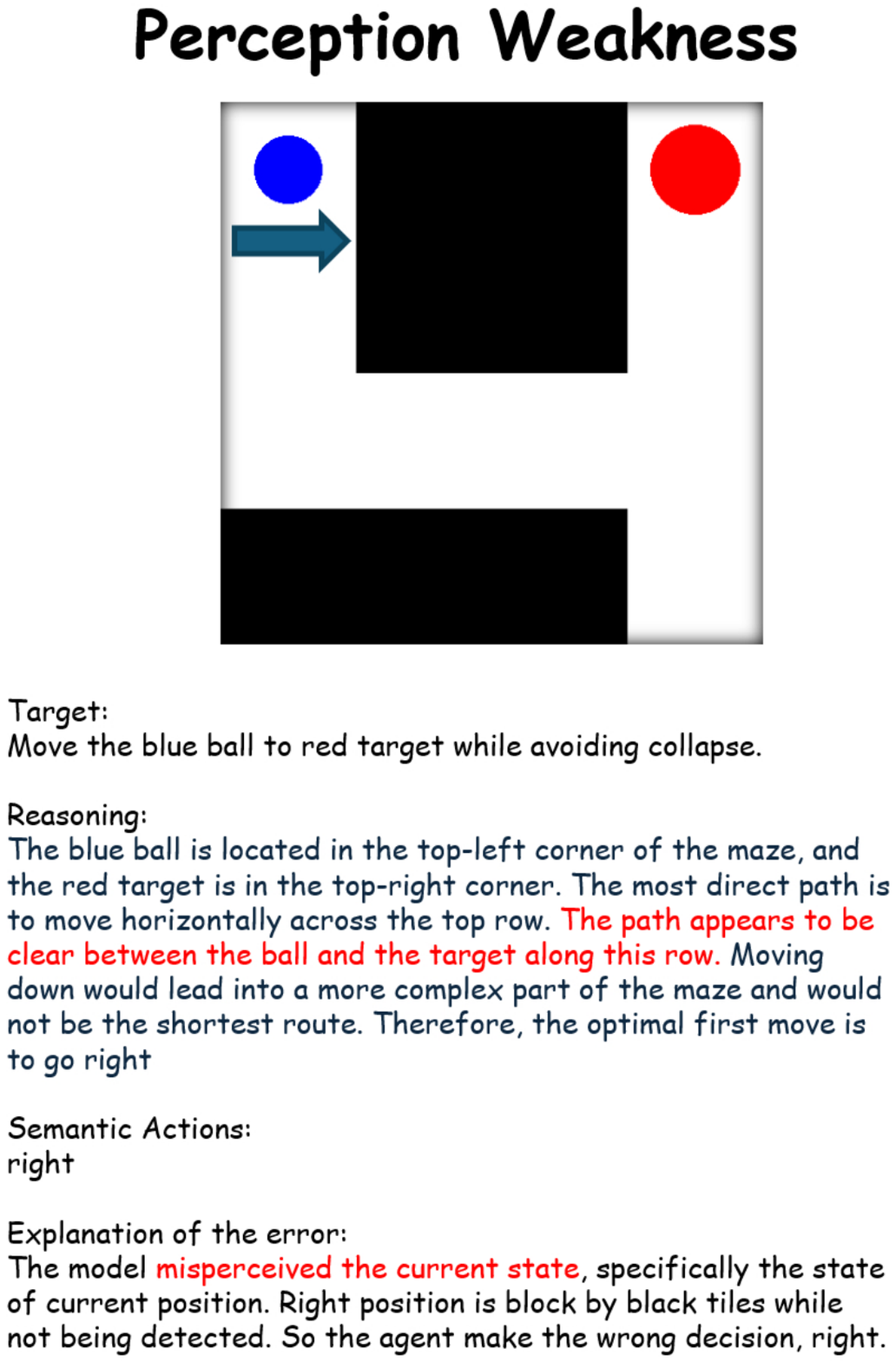}
    \caption{Maze typical error}
    \label{fig:maze error}
\end{wrapfigure}

\begin{figure}[p]
    \centering
    \includegraphics[height=0.95\textheight]{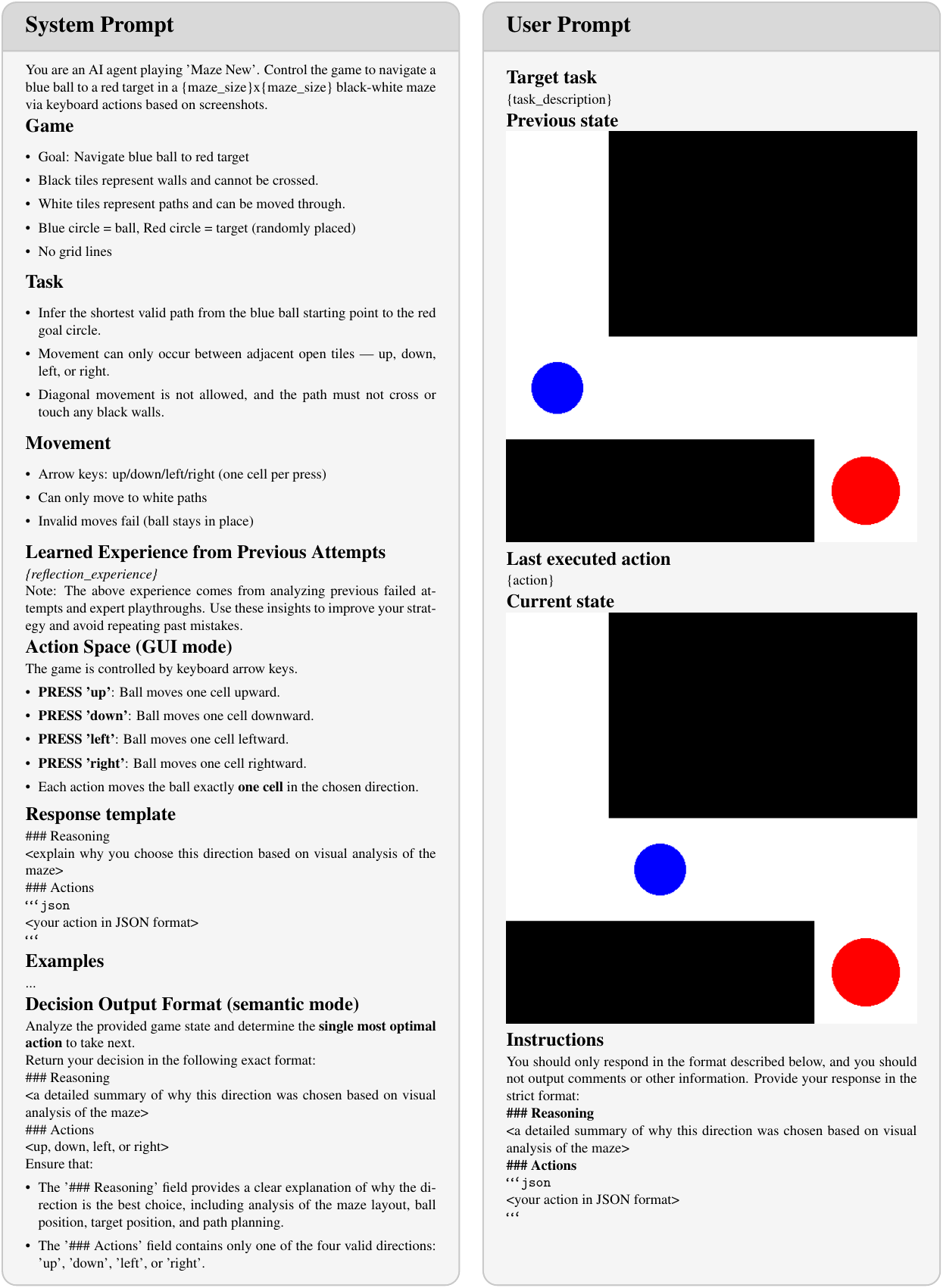}
    \caption{Maze prompt}
    \label{fig:maze prompt}
\end{figure}

Our implementation of the Maze task uses zero-shot agent. We provide the full structure of our prompts in Figure~\ref{fig:maze prompt}.

\subsubsection*{D.4.3 Detailed Analysis for Maze}

The experimental results in the Maze environment reveal a fundamental disconnect between high-level goal seeking and low-level spatial grounding. As illustrated in Figure~\ref{fig:maze error}, the dominant failure mode is Perceptual Weakness (Obstacle Blindness).
    
\begin{table}[h]
    \captionsetup{singlelinecheck=false, justification=raggedright} 
    \caption{Maze Raw Scores}
    \begin{tabular}{lcc}
        \toprule
        \textbf{Model} & \textbf{GUI} & \textbf{GUI VR.} \\
        \midrule
         \textbf{Qwen3-VL-8B}      & $80.0 \pm 45.0$ & $89.0 \pm 31.0$ \\

        \textbf{Qwen3-VL-32B}     & $100.0 \pm 0.0$ & $100.0 \pm 0.0$ \\

        \textbf{GPT-4o-mini}      & $77.0 \pm 43.0$ & $91.0 \pm 26.0$ \\

        \textbf{GPT-4o}           & $80.0 \pm 44.0$ & $85.0 \pm 39.0$ \\

        \textbf{Seed-1.8}         & $88.0 \pm 21.0$ & $100.0 \pm 0.0$ \\

        \textbf{Gemini-2.5-Flash} & $100.0 \pm 0.0$ & $100.0 \pm 0.0$ \\

        \textbf{Gemini-2.5-Pro}   & $100.0 \pm 0.0$ & $100.0 \pm 0.0$ \\
        \bottomrule
    \end{tabular}
    \label{tab:maze_raw}
\end{table}

\textbf{Perceptual Weakness (Obstacle Blindness).} \\
In the depicted instance, the model exhibits a severe hallucination of traversability. Despite a black wall explicitly blocking the horizontal path, the reasoning trace asserts that the route is "clear" and attempts a direct move toward the target. This error indicates that the VLMs prioritize a simple geometric heuristic—minimizing Euclidean distance to the goal—over visual constraint verification. The agent effectively ignores the semantic significance of "wall" pixels, defaulting to an optimistic path planning strategy that directly contradicts the physical constraints of the environment.

\newpage
\clearpage

\subsection{Angry Birds}

\subsubsection*{D.5.1 Game Description for Angry Birds}

\begin{wrapfigure}{r}{0.55\textwidth}
    \centering
    \includegraphics[width=\linewidth]{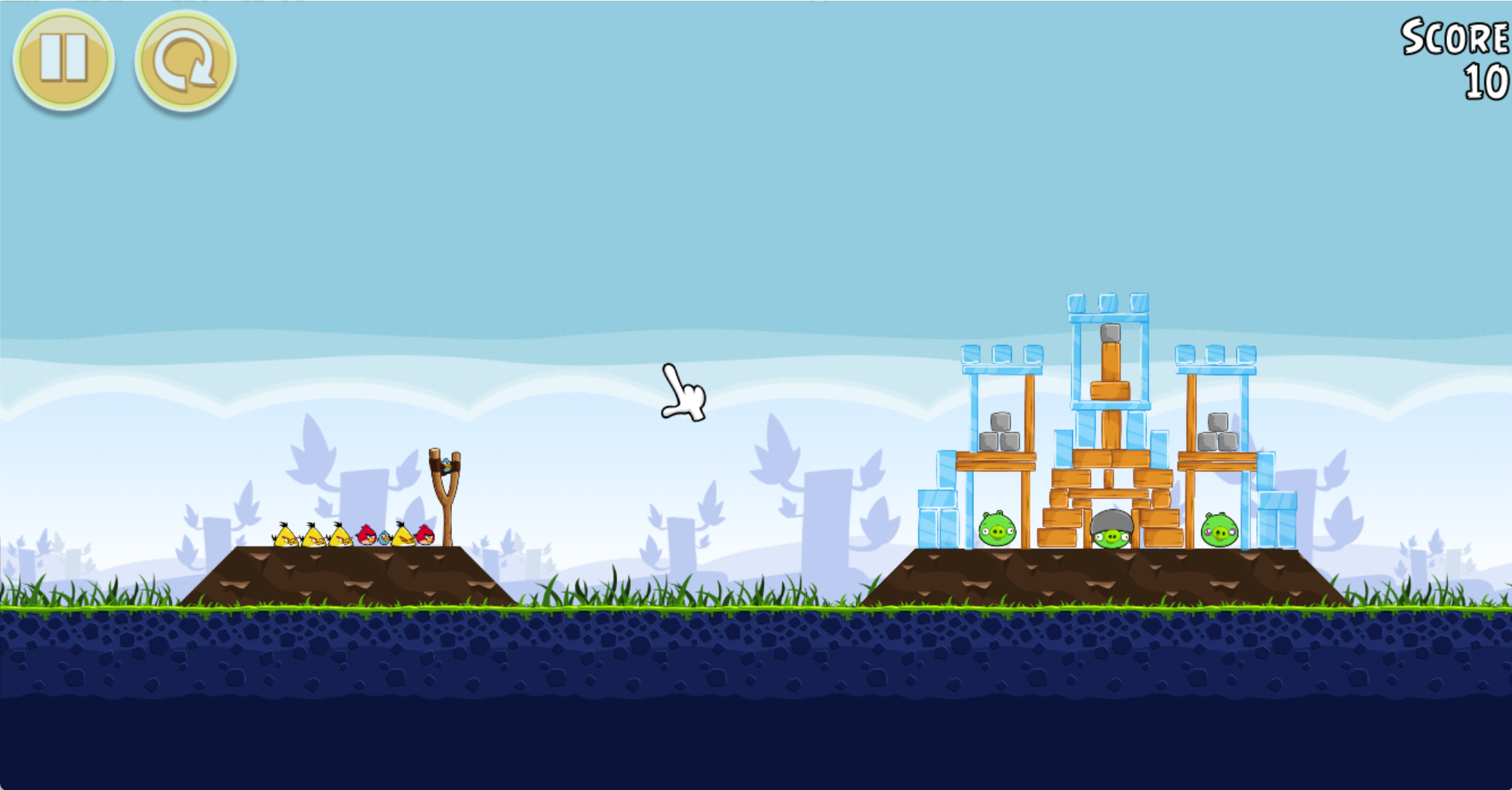}
    \caption{Screenshot of Angry Birds}
    \label{fig:angry birds game title}
\end{wrapfigure}

\textbf{Game Environments.} Angry Birds~~\cite{angrybirds45} is a physics-based puzzle game that requires players to launch projectiles to destroy structures. The environment simulates 2D physics including gravity, momentum, and collision. The primary capability tested here is physics understanding. Agents must infer physical properties (mass, stability) from visual inputs and perform trajectory planning to cause maximum structural damage with limited resources. This game challenges VLMs to translate visual observations into an intuitive physics engine, requiring them to predict how objects will fall, roll, or shatter. It bridges the gap between static visual recognition and dynamic causal prediction, testing whether agents can understand the consequences of force and angle in a simulated physical world.

\noindent\textbf{(1) Game state.} 2D physics simulation environment.

\noindent\textbf{(2) Semantic action space.} Shoot(angle, power) and wait.

\noindent\textbf{(3) Main GUI action space.} Mouse \textit{Drag} (Drag to set angle and power, release to launch).

\noindent\textbf{(4) Evaluation task.} Complete the first three levels. We impose a maximum limit of 30 steps per episode. The agent's performance is quantified by the normalized score. The score is formally defined as:
\begin{equation*}
    \text{\textit{S}}_{\textit{norm}} = \frac{\sum_{i=1}^{3} \text{\textit{S}}_\textit{i}}{S_{max}} \times 100, \quad \text{where $S_{max} = 133000$ represents the maximum score}
\end{equation*}
where \(\sum_{i=1}^{3} \text{\textit{S}}_\textit{i}\) denotes the sum of raw points accumulated in the first three levels, and $S_{max}$ represents the maximum attainable points across these levels. This normalization constrains the score to the range \([0, 100]\), where a value of 1 indicates optimal performance (full points in all three levels) and 0 signifies no progress. 

\noindent\textbf{(5) Expert video content.} The video demonstrates the optimal strategy for clearing the levels. By leveraging the terrain and structural layout indirectly, it completes the levels using the fewest number of birds.

\subsubsection*{D.5.2 Game Prompt for Angry Birds}
\begin{figure}[p]
    \centering
        \includegraphics[height=0.95\textheight]{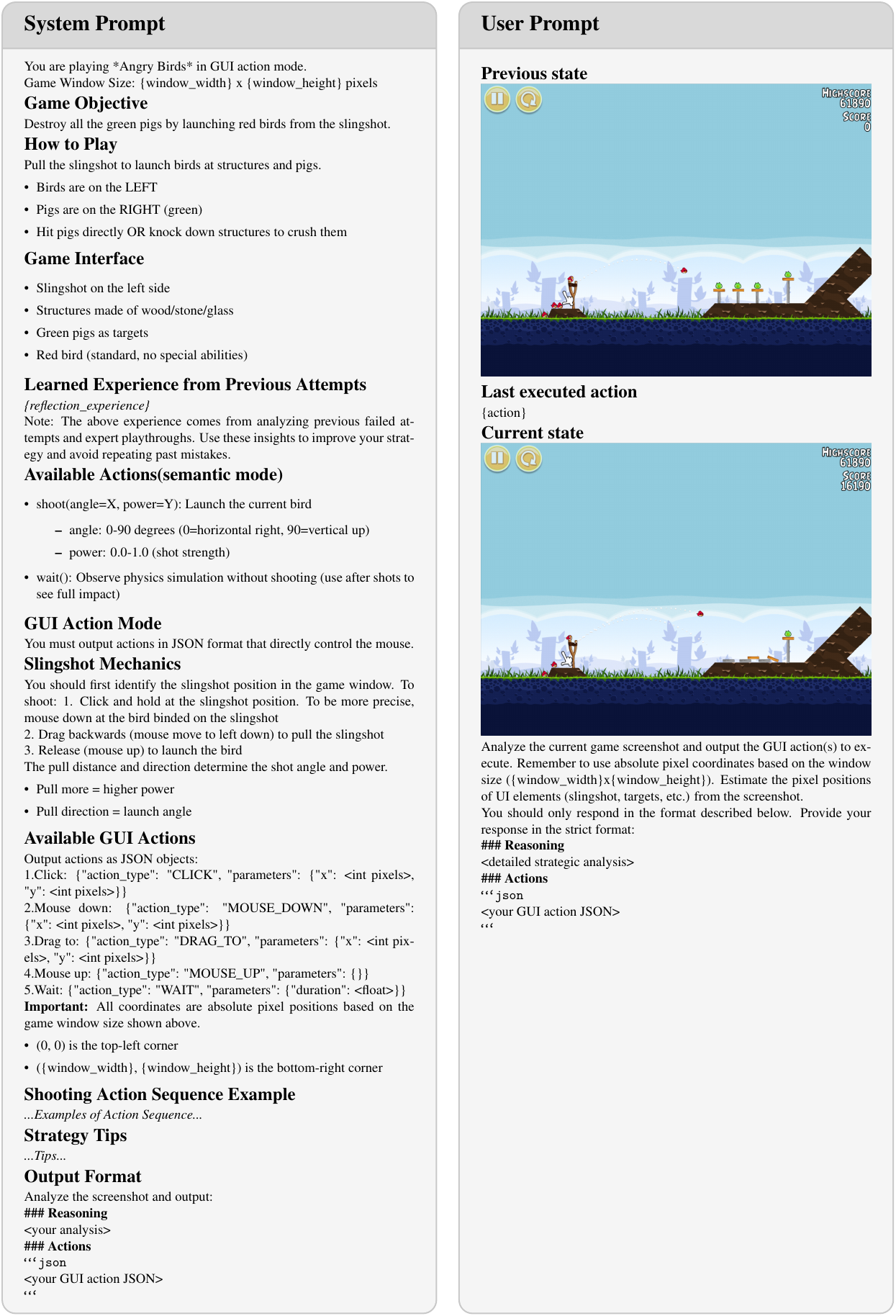}
    \caption{angry birds prompt}
    \label{fig:angrybirds prompt}
\end{figure}

Our implementation of Angry Birds uses zero-shot agent.
We provide the full structure of our prompts in Figure~\ref{fig:angrybirds prompt}.

\subsubsection*{D.5.3 Detailed Analysis for Angry Birds}

The experimental evaluation in Table~\ref{tab:raw score of Angry birds} exposes critical deficiencies in the model's ability to ground abstract physical intuitions into precise continuous control parameters. Unlike discrete action spaces, this domain requires the agent to internalize a physics engine's logic (gravity, momentum, collision), leading to distinct failure modes labeled as \textbf{Semantic-Action Contradiction} and \textbf{Trajectory Planning Failure}.

\begin{table}[htbp]
\centering
\begin{tabular}{lcccc}
\toprule
\textbf{Model} & \textbf{GUI} & \textbf{GUI VR.} & \textbf{Semantic} & \textbf{Semantic VR.} \\ 
\midrule
Qwen3-VL-8B      & 26068 $\pm$ 11039 & 41496 $\pm$ 14364 & 49476.0 $\pm$ 15561 & 84056 $\pm$ 20083 \\
Qwen3-VL-32B     & 55328 $\pm$ 9842  & 79002 $\pm$ 14497 & -                     & -                     \\
GPT-4o-mini      & 22211 $\pm$ 15029 & 17689 $\pm$ 13566 & -                     & -                     \\
GPT-4o           & 19019 $\pm$ 11970 & 20482 $\pm$ 16359 & 65569 $\pm$ 17290 & 81795 $\pm$ 23408 \\
Seed-1.8         & 39102 $\pm$ 9443  & 76076 $\pm$ 16891 & -                     & -                     \\
Gemini-2.5-Flash & 12502 $\pm$ 10906 & 13566 $\pm$ 8512  & 60914 $\pm$ 12236 & 90972 $\pm$ 20482 \\
Gemini-2.5-Pro   & 43491 $\pm$ 17024 & 56126 $\pm$ 15029 & 85386.0 $\pm$ 17689 & 107730 $\pm$ 21014 \\
\bottomrule
\end{tabular}
\vspace{5pt}
\caption{Raw Score of Angry Birds}
\label{tab:raw score of Angry birds}
\vspace{-10pt}
\end{table}

\begin{figure}[ht]
    \centering
        \includegraphics[width=\linewidth]{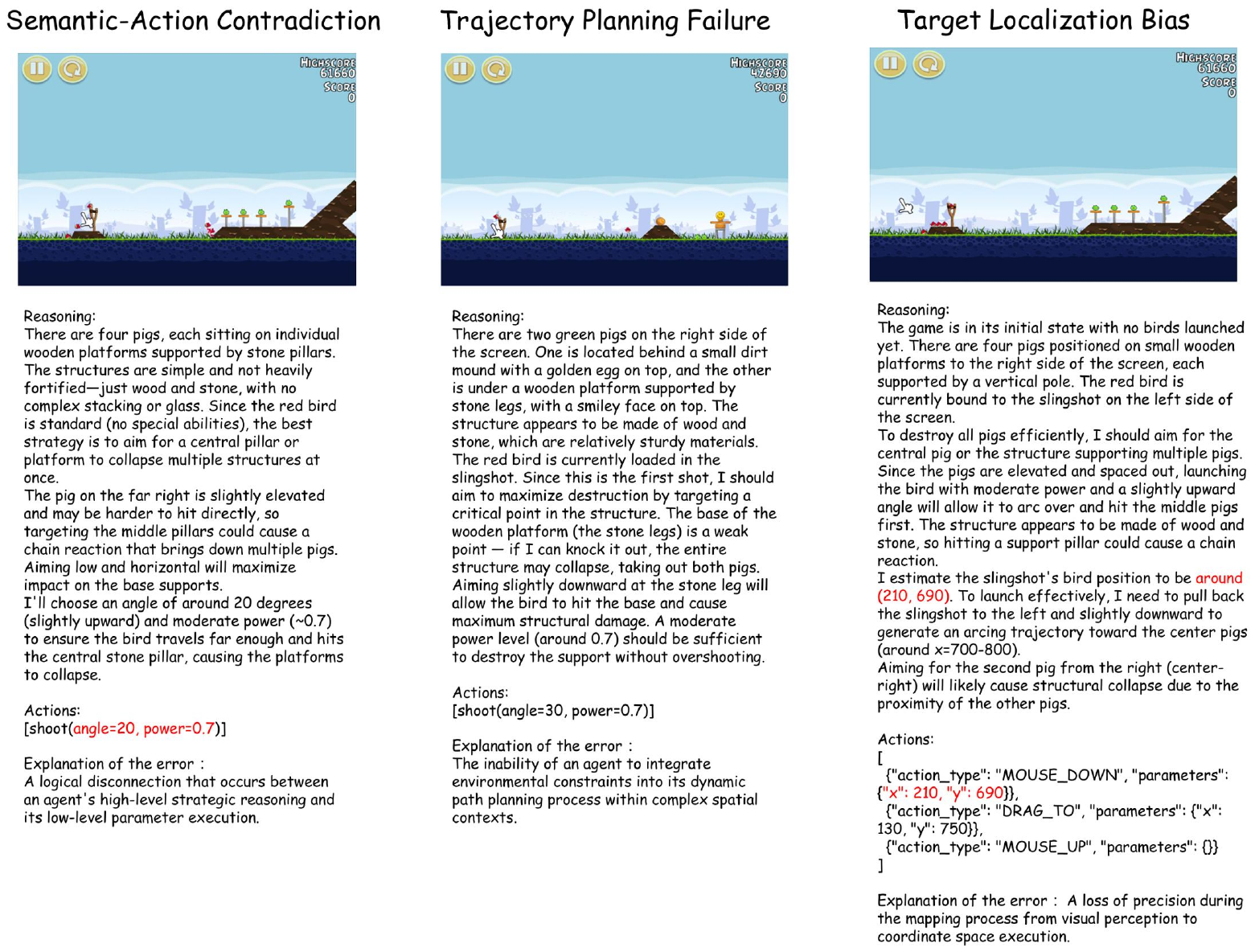}
    \caption{Angry Birds Typical Errors}
    \label{fig:angry birds errors}
\end{figure}

\textbf{Semantic-Action Contradiction.} \\
A prevalent error mode involves a fundamental dissociation between high-level strategic formulation and low-level parameter execution. As illustrated in Figure \ref{fig:angry birds errors} (left), the agent constructs a coherent strategy: "Aiming low and horizontal will maximize impact on the base supports". However, the subsequent action generation (angle=20, power=0.7) fails to align with this semantic intent, producing a trajectory that contradicts the "low and horizontal" descriptor. This phenomenon suggests that while the VLM can reason qualitatively about physics (identifying weak points and structural vulnerabilities), it suffers from a "grounding gap" where it cannot accurately translate these qualitative descriptors into the specific numerical values required by the game's physics engine.

\textbf{Trajectory Planning Failure.} \\
The second failure mode highlights the agent's inability to integrate environmental constraints into its dynamic path planning. In the "Trajectory Planning Failure" instance (Figure \ref{fig:angry birds errors}, center), the agent attempts to target a specific "stone leg" to induce a structural collapse. The reasoning asserts that "Aiming slightly downward... will allow the bird to hit the base". Yet, the generated action (angle=30) typically results in an upward arc in standard coordinate systems, or fails to account for the intervening "small dirt mound" mentioned in the reasoning. This indicates that the model struggles to simulate the parabolic nature of the projectile in complex spatial contexts, often linearizing the path and failing to account for gravity or intermediate obstacles that render the theoretical trajectory invalid.

\textbf{Video Reflection Analysis.} \\
To further explore the potential for model improvement, we introduced a video reflection mechanism. In this framework, a model with higher reasoning capabilities (such as GPT-4o-mini) performs a comprehensive analysis of another model's (such as Qwen's) failure videos alongside expert demonstration videos to generate ``Joint Reflection'' experiences used to guide subsequent operations. Based on the comparison of experimental results(Table~\ref{tab:angry birds expert comprison scores}) and generated reflection insights, the utility of Joint Reflection is significantly superior to that of standalone failure reflection or expert reflection. 

\begin{wraptable}{r}{0.65\textwidth} 
  \centering
  \small
  \setlength{\tabcolsep}{3pt}
  \begin{tabular}{lcccc}
    \toprule
    \textbf{Model} & \textbf{Baseline} & \textbf{Expert VR.} & \textbf{Failure VR.} & \textbf{Joint VR.} \\
    \midrule
    GPT-4o-mini       & $55.0 \pm 6.0$ & $73.0 \pm 8.0$ & $57.0 \pm 13.0$ & $\mathbf{77.0 \pm 7.0}$ \\

    Qwen3-VL-32B & $50.0 \pm 11.0$ & $48.0 \pm 12.0$ & $54.0 \pm 13.0$ & $\mathbf{74.0 \pm 10.0}$ \\
    \bottomrule
  \end{tabular}
  \vspace{5pt}
  \caption{Performance comparison of Expert vs. Failure reflection}
  \label{tab:angry birds expert comprison scores}
  \vspace{-15pt}
\end{wraptable}

The core advantages of this approach stem from addressing the inherent limitations of single-source reflection: while expert reflection in isolation only provides positive guidance without alerting the agent to its typical errors, failure reflection alone offers negative feedback that struggles to generate high-level strategic heuristics without a successful paradigm to follow. Joint Reflection achieves a comprehensive coverage of knowledge by combining strategic insights with failure lessons, thereby enhancing the practicality and error tolerance of the strategies. However, experimental observations show that the resulting improvement in final scores remains constrained by the aforementioned semantic-action contradictions. This suggests that the bottleneck for certain models in physical tasks lies not only in ``Strategy Discovery'' but also in a lack of precision regarding underlying ``Motor-control Alignment''.

\subsection{Slay the Spire}
\subsubsection*{D.6.1 Game Description for Slay the Spire}

\begin{wrapfigure}{r}{0.5\textwidth}
\vspace{-40pt}
    \centering
    \includegraphics[width=\linewidth]{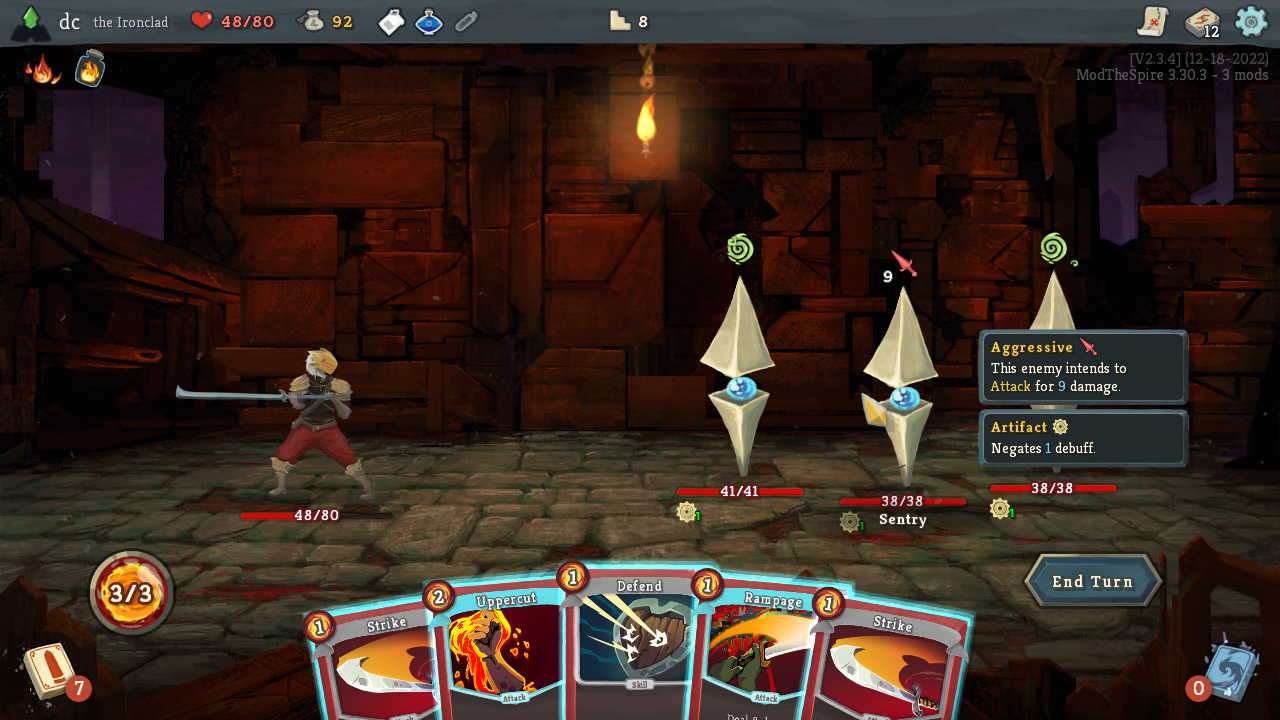}
    \caption{Screenshot of Slay the Spire}
    \vspace{-10pt}
    \label{fig:slay the spire game title}
\end{wrapfigure}

\textbf{Game Environments.} 
Slay the Spire~\cite{slaythespire44} is a strategy game combining deck-building mechanics with roguelike progression. The agent must climb a spire by battling enemies using a deck of cards that evolves over time. This game presents a complex challenge in stochastic planning and resource management. 

Agents are responsible for in-battle decision making and post-combat deck management. These agents are tested on their ability to adapt to random card draws, optimize a deck for long-term synergy, and make trade-off decisions between immediate survival and future power. As a "Non-Real-Time Non-Linear" benchmark, it requires the agent to understand complex text-based card effects and synergize them with visual game states, evaluating high-level strategic planning where current choices significantly impact future viability.

Specifically, regarding the semantic implementation, we utilize the environment wrapper and interface design introduced in ~\cite{oark5}.

\noindent\textbf{(1) Game state} 2D turn-based environment with visible cards and enemy intents.

\noindent\textbf{(2) Semantic action space} Play a card with an optional target; End the current game turn; Select a card to add to the deck. 

\noindent\textbf{(3) Main GUI action space} Mouse \textit{Move}, \textit{Click} and \textit{Drag}(hold down the left mouse button, move to the target monster and release).

\noindent\textbf{(4) Evaluation task} Climb the spire. We impose a maximum limit of 200 steps per episode.The agent's performance is quantified by the maximum floor reached, averaged over a minimum of three trials. The evaluation metrics were divided by 51 (the maximum possible number of floors) . The normalized score is defined as follows: 

\begin{equation*}
S_{norm} = \frac{\frac{1}{n}\sum_{i=1}^nS_{i}}{S_{max}} \times100,  \text{where } S_{max} \text{ is the maximum possible score, and } n \text{ is the number of trials.}
\end{equation*}

\noindent\textbf{(5) Expert video content} The expert video is a skill tutorial video. Specifically, it addresses and corrects erroneous intuitions regarding health management (over-reliance on resting for HP recovery) and card acquisition (the "collect-all" mentality leading to bloated decks). By conducting a deep dive into the underlying game mechanics—including deck cycling, draw probabilities, and keywords such as "Exhaust" and "Ethereal"—the tutorial demonstrates the necessity of streamlining decks through selective drafting and card removal. The guide leads agents to make decision based on probabilistic reasoning and long-term utility.

\subsubsection*{D.6.2 Game Prompt For Slay the Spire}
\begin{figure}[p]
    \centering
        \includegraphics[height=0.95\textheight]{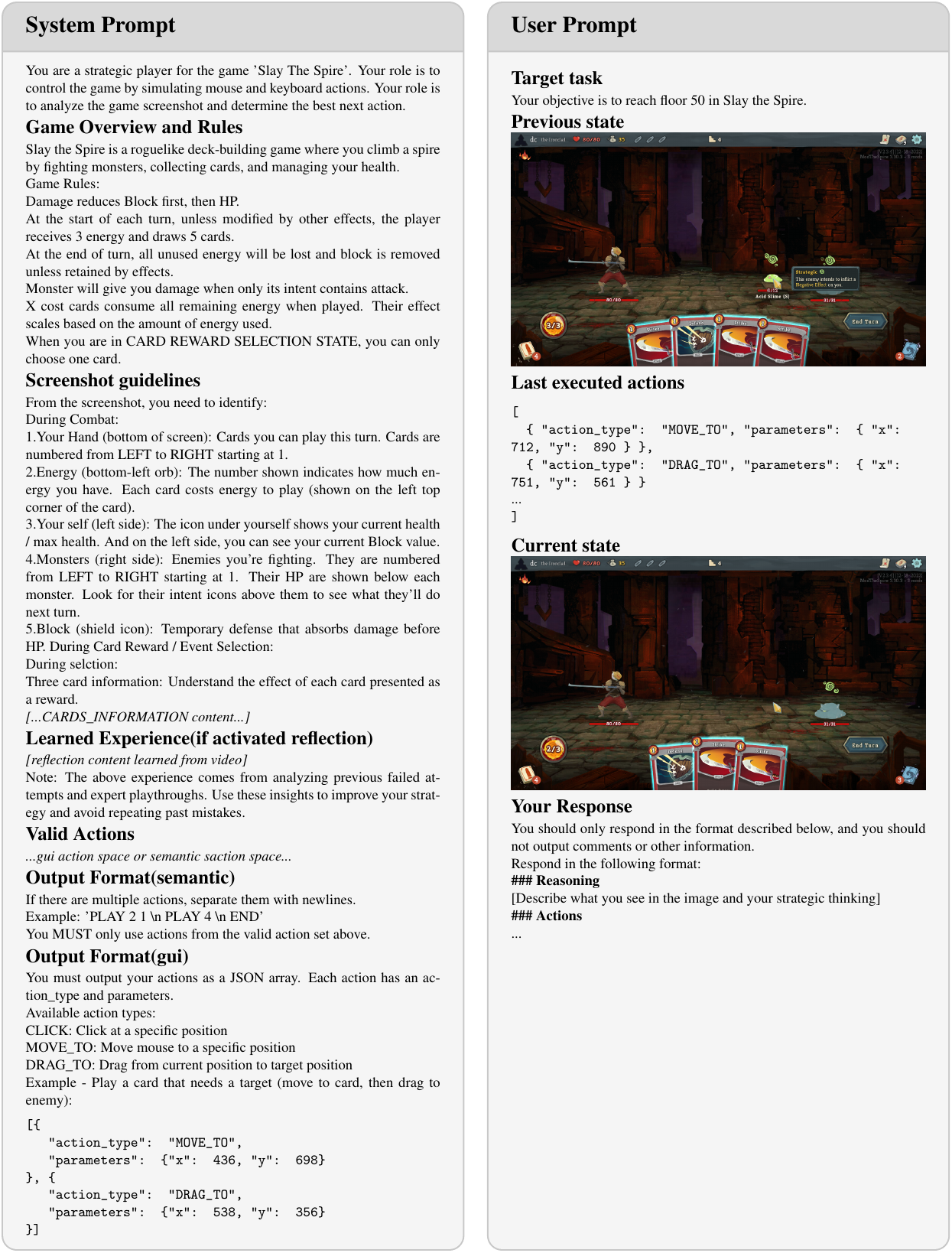}
    \caption{slay the spire prompt}
    \label{fig:slay the spire prompt}
\end{figure}

We provide the full structure of our prompts in Figure~\ref{fig:slay the spire prompt}.

\subsubsection*{D.6.3 Detailed Analysis For Slay the Spire}

\begin{wrapfigure}{r}{0.65\textwidth} 
  \centering
  \vspace{-5pt} 
  \small 
  \begin{tabular}{lcccc}
    \toprule
   \textbf{Model} & \textbf{GUI} & \textbf{Semantic} & \textbf{GUI VR.} & \textbf{Semantic VR.} \\
    \midrule
    Gemini-2.5-Flash      & $1 \pm 0$      & $6 \pm 0$      & $1 \pm 0$     & $8.0 \pm 0.0$ \\
    Gemini-2.5-Pro        & $2.0 \pm 1.73$ & $19.75 \pm 11.5$ & $4 \pm 0.0$   & $14.0 \pm 4.0$ \\
    Qwen3-VL-8B           & $6 \pm 0$      & $6 \pm 0$      & $3.33 \pm 4.0$ & $2.25 \pm 2.5$ \\
    Qwen3-VL-32B          & $4.2 \pm 1.92$ & --             & $5.75 \pm 0.5$ & -- \\
    GPT-4o                & $1 \pm 0$      & $13.5 \pm 5.0$  & $1 \pm 0$     & $14.0 \pm 4.0$ \\
    GPT-4o-mini           & $1 \pm 0$      & --             & $1 \pm 0$     & -- \\
    Seed-1.8       & $20.5 \pm 10.84$& --             & $14.0 \pm 4.0$ & -- \\
    \bottomrule
  \end{tabular}
  \caption{raw scores in slay the spire}
  \label{tab:slay the spire raw scores}
  \vspace{-10pt}
\end{wrapfigure}

\begin{figure}[ht]
    \centering
        \includegraphics[width=\linewidth]{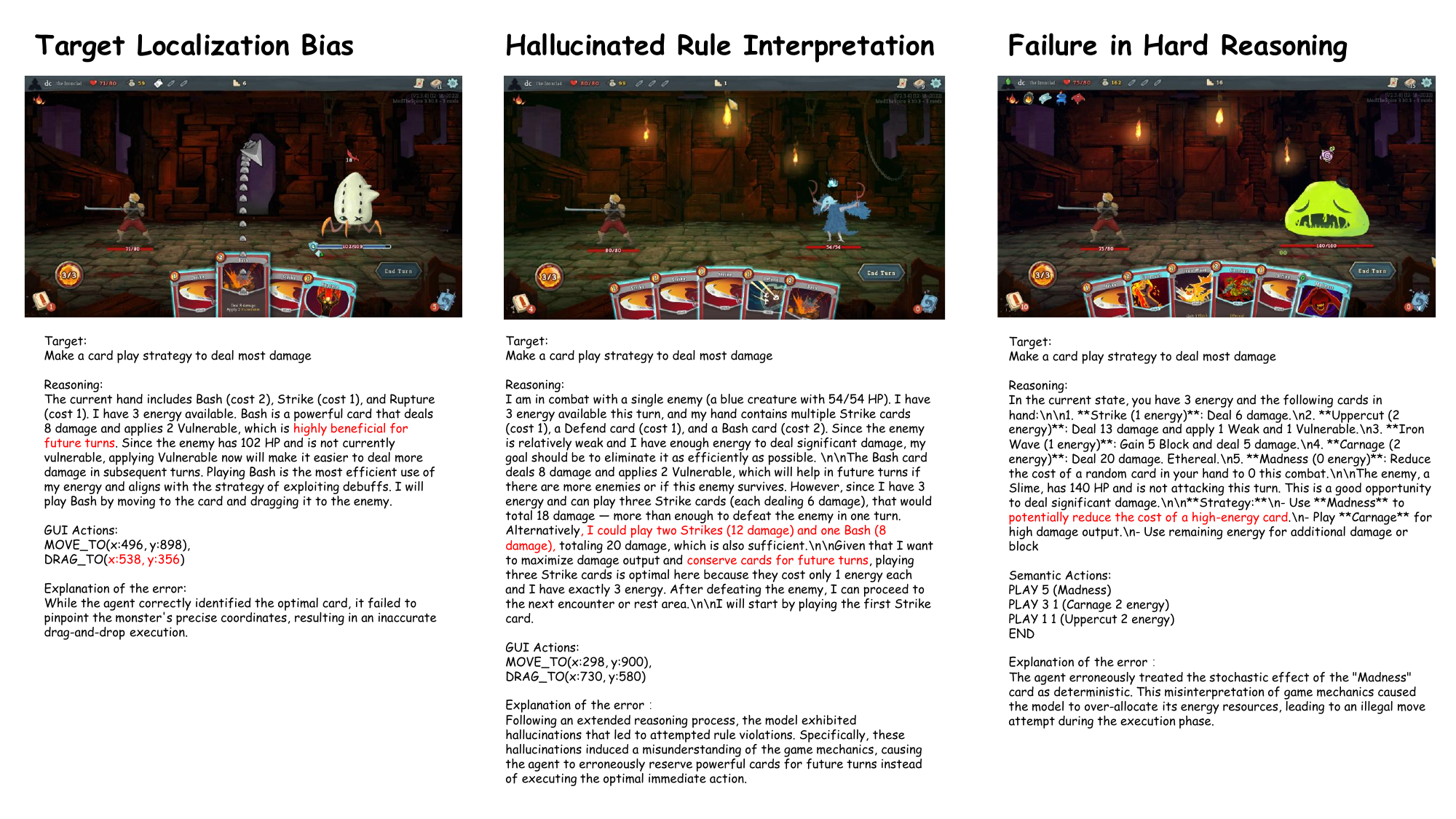}
    \caption{slay the spire typical errors}
    \label{fig:slay the spire typical errors}
    \vspace{-20pt}
\end{figure}

The experimental results in Slay the Spire reveal a distinct performance ceiling correlated with difficulty progression. As illustrated in \cref{tab:slay the spire raw scores}, the sixth floor—typically the first Elite encounter—acts as a critical filter for VLMs. While high-reasoning models (Gemini-2.5-Pro, Seed-1.8) occasionally navigate deeper, most suffer catastrophic failure at this juncture. This concentration of deaths at Boss and Elite encounters underscores a critical vulnerability: the extremely low tolerance for error in high-stakes battles. Unlike basic combat, where mistakes are recoverable, reasoning flaws here lead to sub-optimal strategies that make defeat inevitable against such formidable opponents. The 'Failure in Hard Reasoning' and "Hallucinated Rule Interpretation" in \cref{fig:slay the spire typical errors} showcase a representative instance where, despite correct GUI execution, the model's inability to calculate lethal damage or manage defensive resources leads to a terminal state in an Elite encounter.

\textbf{Execution Gap.} \\
We observe a profound "GUI-to-Semantic Gap," particularly in reasoning models like Gemini-2.5-Pro. Despite its high strategic intelligence, it frequently fails to translate intent into precise pixel-level coordinates. In contrast, models specifically pre-trained on GUI tasks, such as Qwen3-VL-8B, Qwen3-VL-32B and Seed-1.8, demonstrate a significantly narrower gap, showing superior localization capabilities. However, this specialized training introduces a secondary failure mode: a regression in instruction-following stability. Even when explicitly prompted to utilize specific pixel ranges, these models often revert to normalized [0, 1000] coordinates learned during pre-training. 

A nuanced failure observed across nearly all models is the "Target Localization Bias." While models can reliably locate card assets at the bottom of the screen, they struggle with spatial perception of the battlefield. Agents frequently misidentify environmental decor—such as pillars or background textures—as enemy targets. As demonstrated in \cref{fig:slay the spire typical errors} (left), models frequently fail to bind attack actions to the correct enemy hitbox, instead identifying background architecture as valid targets. This suggests that while VLMs possess rudimentary object detection for standard UI elements, they lack the spatial intelligence required to distinguish interactive game entities from static background assets in complex, non-standardized scenes.

\begin{wrapfigure}{r}{0.5\textwidth} 
    \vspace{-10pt}
    \centering
    \small
    \includegraphics[width=\linewidth]{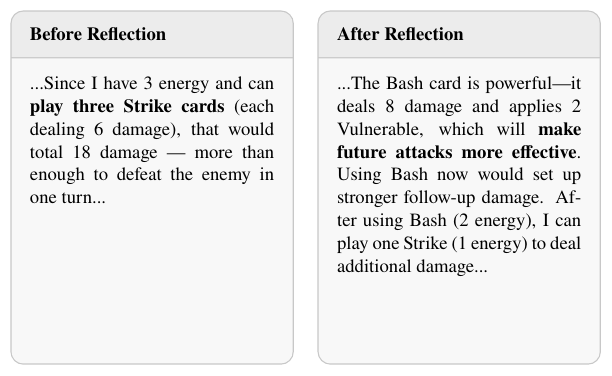}
    \caption{Example of Qwen3-VL-32B Reflection Improvement}
    \label{fig:Qwen3-VL-32B_reflection}
    \vspace{-20pt}
\end{wrapfigure}

\textbf{Video Reflection Analysis.} \\
A significant finding is that the utility of reflection is inversely proportional to the model's baseline reasoning capability. For powerful proprietary models like Gemini-2.5-Pro and GPT-4o, the inclusion of video-based tips resulted in negligible performance gains. These models likely possess enough latent world knowledge to infer optimal strategies. Conversely, smaller models like Qwen-8B struggle to synthesize video data, often falling victim to "reasoning hallucinations" where they attempt to apply contextually irrelevant strategies. Mid-tier models like Gemini-2.5-Flash and Qwen3-VL-32B, showed a stable positive trajectory (\cref{fig:Qwen3-VL-32B_reflection}), utilizing reflection to consistently defeat additional elite enemies.

However, the introduction of a reflection mechanism, intended to allow models to learn from historical failure videos, yielded counter-intuitive results. In the GUI action space, reflection often led to a decrease in overall performance. This is primarily attributed to "Execution Interference," where the cognitive load of processing past mistakes appears to degrade the model’s precision in pixel localization.

\subsection{Ace Attorney}

\subsubsection*{D.7.1 Game Description for Ace Attorney}

\begin{wrapfigure}{r}{0.55\textwidth}
\vspace{-20pt}
    \centering
    \includegraphics[width=\linewidth]{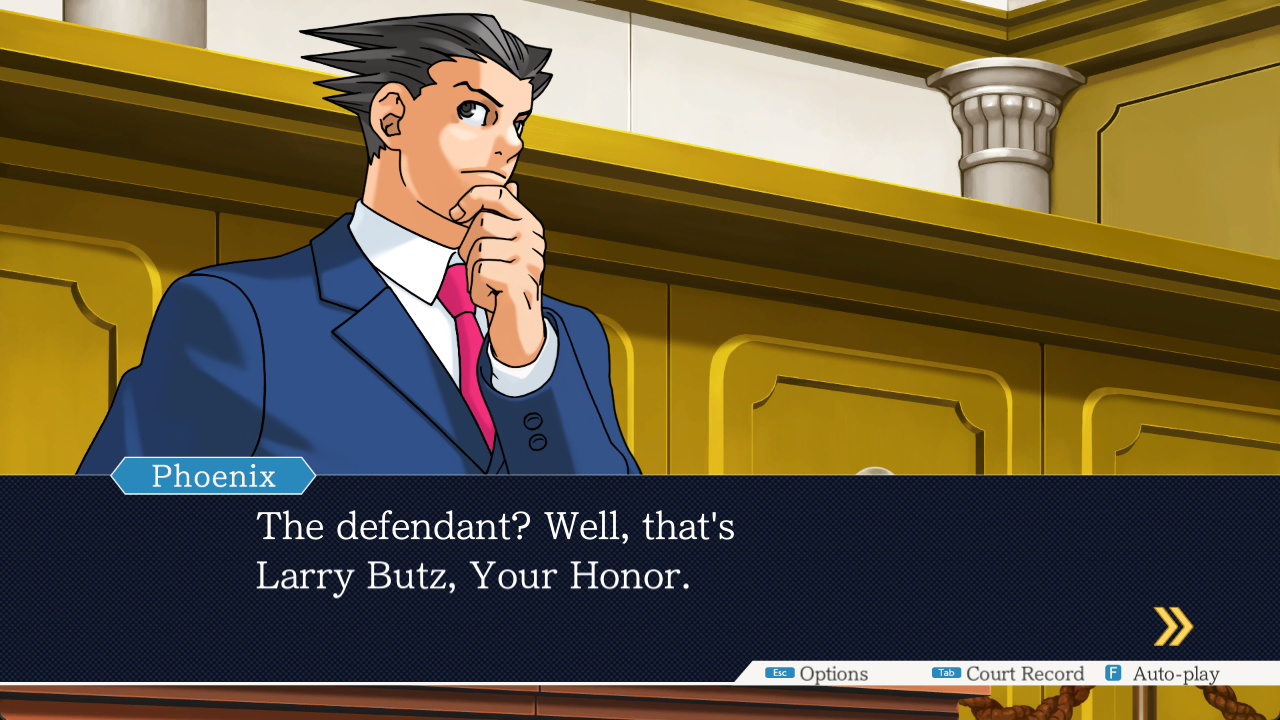}
    \caption{Screenshot of Ace Attorney}
    \vspace{-10pt}
    \label{fig:Ace game title}
\end{wrapfigure}

\textbf{Game Environments.}  Ace Attorney~\cite{ace43} is a narrative-driven visual novel centered on courtroom simulations. The player acts as a defense lawyer, investigating crime scenes and cross-examining witnesses. This environment heavily focuses on high-level Natural Language Understanding and logic deduction. Agents must detect contradictions between text testimony and visual evidence, requiring deep context retention and the ability to infer truth from conflicting information. Unlike reaction-based games, Ace Attorney tests the agent's "slow thinking" capabilities—reading comprehension, critical thinking, and the synthesis of multimodal clues. It serves as a benchmark for evaluating how well VLMs can reason through complex narratives and align visual details with textual inconsistencies.

\noindent\textbf{(1) Game state.} High-fidelity 2D visual novel with text testimony and visual evidence.

\noindent\textbf{(2) Main GUI action space.} Mouse \textit{Click} (Select dialogue options, present evidence).

\noindent\textbf{(3) Evaluation task.} Complete the first chapter and identify the real culprit. We impose a maximum limit of 500 steps per episode. The agent's performance is quantified by the average number of completed milestones. The milestones are generated from the expert video and are listed in Table~\ref{tab:ace_milestone}. To obtain a standardized score, the number of completed milestones is normalized by the total number of milestones, and the calculation formula is as follows:

\begin{equation*}
S_{norm} = \frac{N_{\text{completed}}}{N_{\text{total}}} \times 100
\end{equation*}

where \( N_{\text{completed}} \) denotes the number of milestones completed by the agent, and \( N_{\text{total}} \) denotes the total number of milestones (9 in total, as listed in Table~\ref{tab:ace_milestone}). The normalized score ranges from 0 to 100.

\begin{table}[h]
  \centering
  \small
  \begin{tabular}{ll c ll}
    \toprule
    \textbf{No.} & \textbf{Milestone Title} & & \textbf{No.} & \textbf{Milestone Title} \\
    \midrule
    1 & Identify Defendant           & & 6 & Reveal Weapon's Secret \\
    2 & Identify Victim              & & 7 & Sound the Clock \\
    3 & Identify Cause of Death      & & 8 & Solve the Time Gap \\
    4 & Expose Time Contradiction    & & 9 & Not Guilty Verdict \\
    5 & Expose TV Logic Hole         & &   & \\ 
    \bottomrule
  \end{tabular}
  \caption{Milestones of Ace Attorney}
  \label{tab:ace_milestone}
\end{table}

\noindent\textbf{(4) Expert video content.} The video illustrates the process of completing Chapter One, including choosing the correct dialogue options and selecting the proper evidence to raise objections.

\subsubsection*{D.7.2 Game Prompt for Ace Attorney}
\begin{figure}[p]
    \centering
        \includegraphics[height=0.9\textheight]{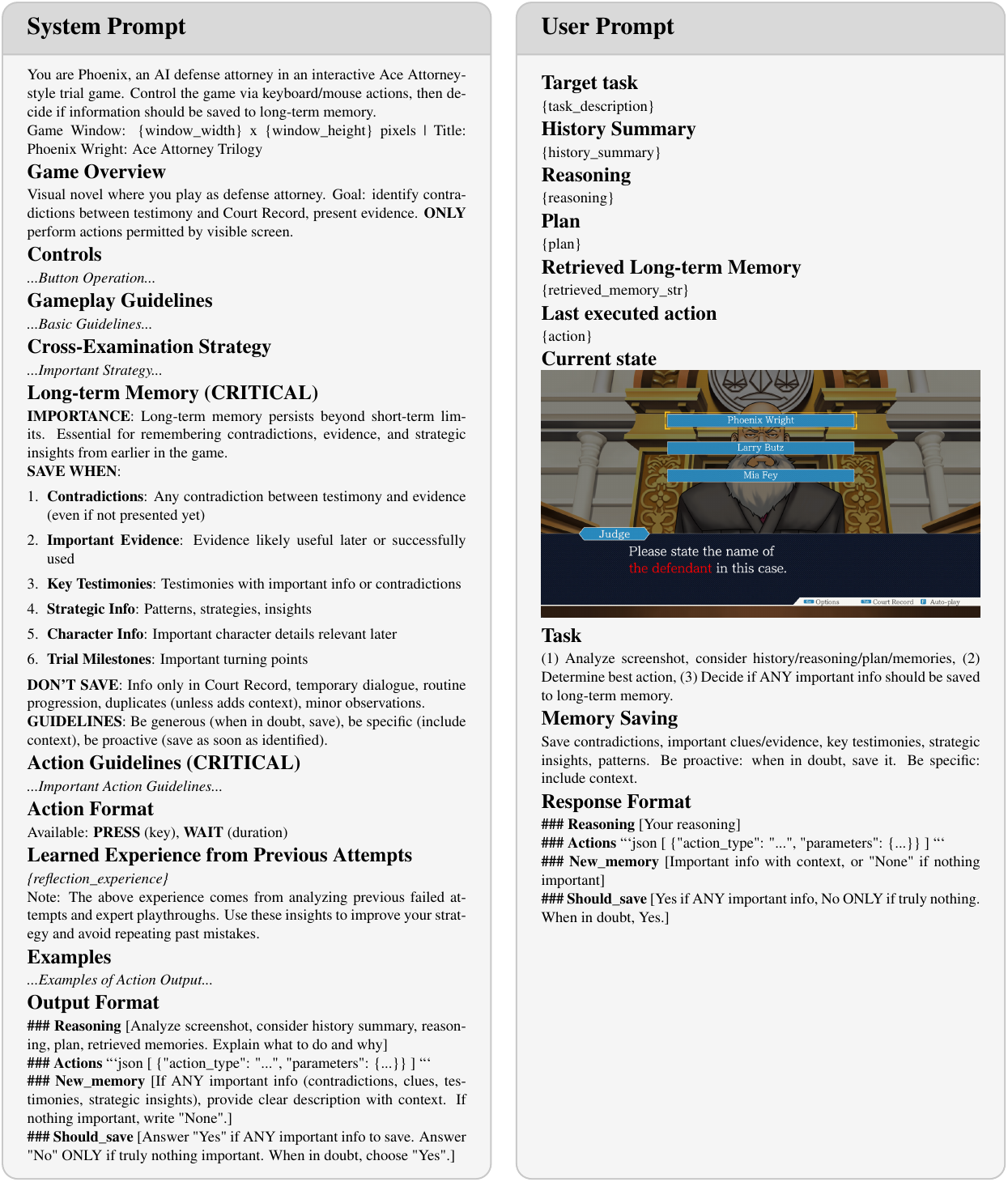}
    \caption{Ace Attorney Prompt}
    \label{fig:ace_prompt}
\end{figure}

Our implementation of Ace Attorney uses memory agent.We provide the full structure of our prompts in Figure~\ref{fig:ace_prompt}.

\subsubsection*{D.7.3 Detailed Analysis for Ace Attorney}
\begin{figure}[t]
    \centering
        \includegraphics[width=\linewidth]{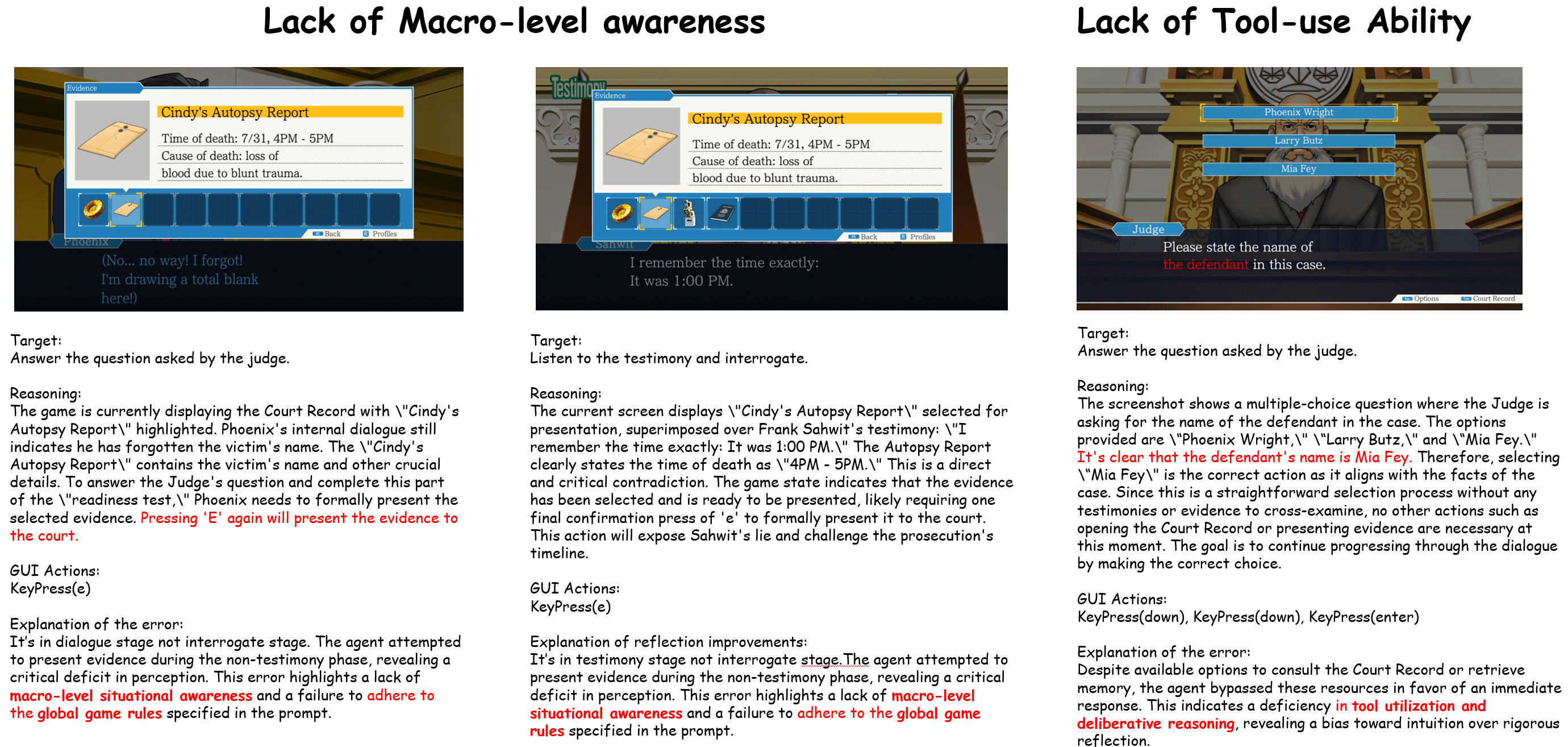}
    \caption{Ace Attorney typical errors}
    \label{fig:pwaat typical errors}
\end{figure}

\begin{wrapfigure}{r}{0.5\textwidth} 
  \centering
  \vspace{-5pt} 
  \small 
  \begin{tabular}{lcc}
    \toprule
\textbf{Model} & \textbf{GUI} & \textbf{GUI VR. } \\
    \midrule
     Qwen3-VL-8B      & $7.0 \pm 6.0$ & $26.0 \pm 19.0$ \\

    Qwen3-VL-32B     & $17.0 \pm 6.0$ & $37.0 \pm 6.0$ \\

    GPT-4o-mini      & $0.0 \pm 0.0$ & $0.0 \pm 0.0$ \\

    GPT-4o           & $18.0 \pm 6.0$ & $33.0 \pm 11.0$ \\

    Seed-1.8         & $33.0 \pm 0.0$ & $59.0 \pm 6.0$ \\

    Gemini-2.5-Flash & $33.0 \pm 11.0$ & $29.0 \pm 25.0$ \\

    Gemini-2.5-Pro   & $37.0 \pm 6.0$ & $48.0 \pm 6.0$ \\
    \bottomrule
  \end{tabular}
  \caption{Raw scores in Ace Attorney}
  \label{tab:attorney raw scores}
  \vspace{-10pt}
\end{wrapfigure}

The experimental results in the Ace Attorney environment underscore a significant challenge in maintaining long-horizon context and utilizing external knowledge bases. Unlike the spatial reasoning required in grid-based games, Ace Attorney demands high-fidelity text comprehension and rigorous state tracking. As illustrated in Figure~\ref{fig:pwaat typical errors}, the primary failure modes manifest as \textbf{Macro-level Contextual Misalignment} and \textbf{Deficiency in Tool Utilization}.

\textbf{Macro-level Contextual Misalignment.} \\
A critical failure mode, labeled as "Lack of Macro-level awareness", involves the agent’s inability to distinguish between distinct game phases. In the observed instance, the game is in a passive "Dialogue Phase," where the protagonist is internally monologuing about memory loss. However, the agent erroneously identifies this as an active "Interrogation Phase" and attempts to execute a "Present Evidence" action ('E'). This error highlights a failure in global state tracking, where the model conflates the semantic content of the text (mentioning the "Autopsy Report") with the procedural rules of the current gameplay state. The agent fails to inhibit action execution during non-interactive narrative sequences, revealing a lack of hierarchical awareness regarding the game's rule set.

\textbf{Deficiency in Tool Utilization and Deliberative Reasoning}. \\
The second failure mode, "Lack of Tool-use Ability", demonstrates a bias toward immediate, intuitive responding over rigorous information retrieval. When presented with a factual query from the Judge ("Please state the name of the defendant"), the optimal strategy dictates consulting the "Court Record"—an available in-game tool—to verify information. Instead, the agent bypasses this retrieval step and hallucinates an incorrect answer ("Mia Fey") based on a flawed internal probability distribution. This behavior indicates a deficit in epistemic uncertainty detection; the model fails to recognize the limits of its immediate context window and neglects to employ available tools to ground its reasoning in ground-truth data.

\subsection{Civilization VI}
\subsubsection*{D.8.1 Game Description for Civilization VI}

\begin{wrapfigure}{r}{0.5\textwidth}
\vspace{-30pt}
    \centering
    \includegraphics[width=\linewidth]{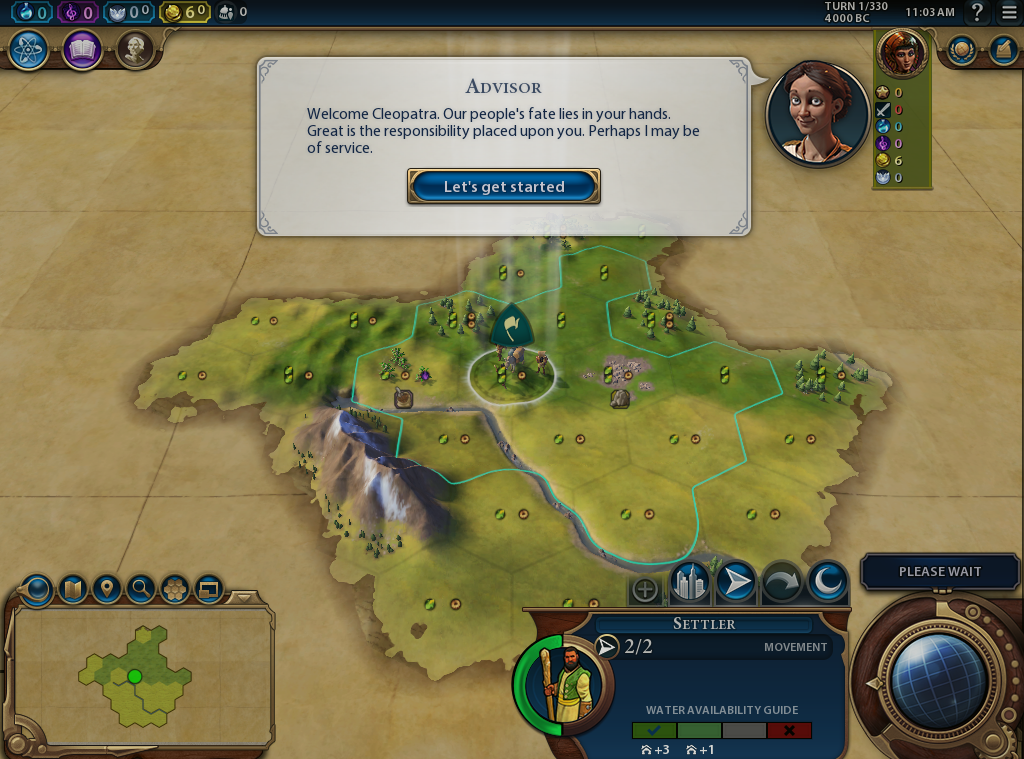}
    \caption{Screenshot of Civilization VI}
    \label{fig:civilization game title}
\end{wrapfigure}
\textbf{Game Environments.} Civilization VI~\cite{civil42} is a grand strategy game of immense complexity where players guide a civilization from the Stone Age to the Information Age. The environment involves exploring a map hidden by the fog , managing economies, engaging in diplomacy, and commanding the operations of various units. This game is the ultimate test for strategic planning, resource management and utilization. Agents must balance competing priorities over thousands of steps, handling a massive state space with long-term horizons. Classified as "Non-Real-Time Non-Linear," it probes the agent's ability to manage multi-faceted systems—military, culture, science—simultaneously. It challenges the model to maintain a coherent grand strategy while executing specific turn-by-turn tactical decisions in a world that is constantly changing and in need of exploration.

\noindent\textbf{(1) Game state.} 2D, turn based environment, resources and map to be explored, various indicators of development and units from different civilizations

\noindent\textbf{(2) Main GUI action space.} Mouse \textit{Click} and \textit{Drag} (Unit movement, city management, menu navigation).

\noindent\textbf{(3) Evaluation task.} Finish the tutorial.  We impose a maximum limit of 200 steps per episode. The agent's performance is quantified by the average number of completed milestones. The milestones are generated from the expert video and are listed in Table~\ref{tab:civilization_milestone}. To obtain a standardized score, the number of completed milestones is normalized by the total number of milestones, and the calculation formula is as follows:

\begin{equation*}
S_{norm} = \frac{N_{\text{completed}}}{N_{\text{total}}} \times100
\end{equation*}

where \( N_{\text{completed}} \) denotes the number of milestones completed by the agent, and \( N_{\text{total}} \) denotes the total number of milestones (12 in total, as listed in Table~\ref{tab:civilization_milestone}).

\begin{table}[h]
  \centering
  \caption{Milestones of Civilization VI}
  \label{tab:civilization_milestone}
  \small 
  \begin{tabular}{ll c ll} 
    \toprule
    \textbf{No.} & \textbf{Milestone Title} & & \textbf{No.} & \textbf{Milestone Title} \\
    \midrule
    1 & Found First City             & & 7  & Place Campus District \\
    2 & Select First Technology      & & 8  & Optimize Military Policy \\
    3 & Tribal Village Discovery     & & 9  & Declare Surprise War \\
    4 & Meet Rival Civilization      & & 10 & Capture Enemy Settler \\
    5 & Discover Natural Wonder      & & 11 & Found City with Captured Unit \\
    6 & Found Second City            & & 12 & Capture Capital \& Victory \\
    \bottomrule
  \end{tabular}
\end{table}

\noindent\textbf{(4) Expert video content.} The expert video is a demonstration video. Specifically, it demonstrates how to follow the tutorial's prompts to build cities, select technologies, change policies, and explore the world. Through making diplomatic decisions and developing military strength, the video also teaches players how to achieve victory by defeating competitors through warfare.

\subsubsection*{D.8.2 Game Prompt For Civilization VI}
\begin{figure}[p]
    \centering
        \includegraphics[width=\linewidth]{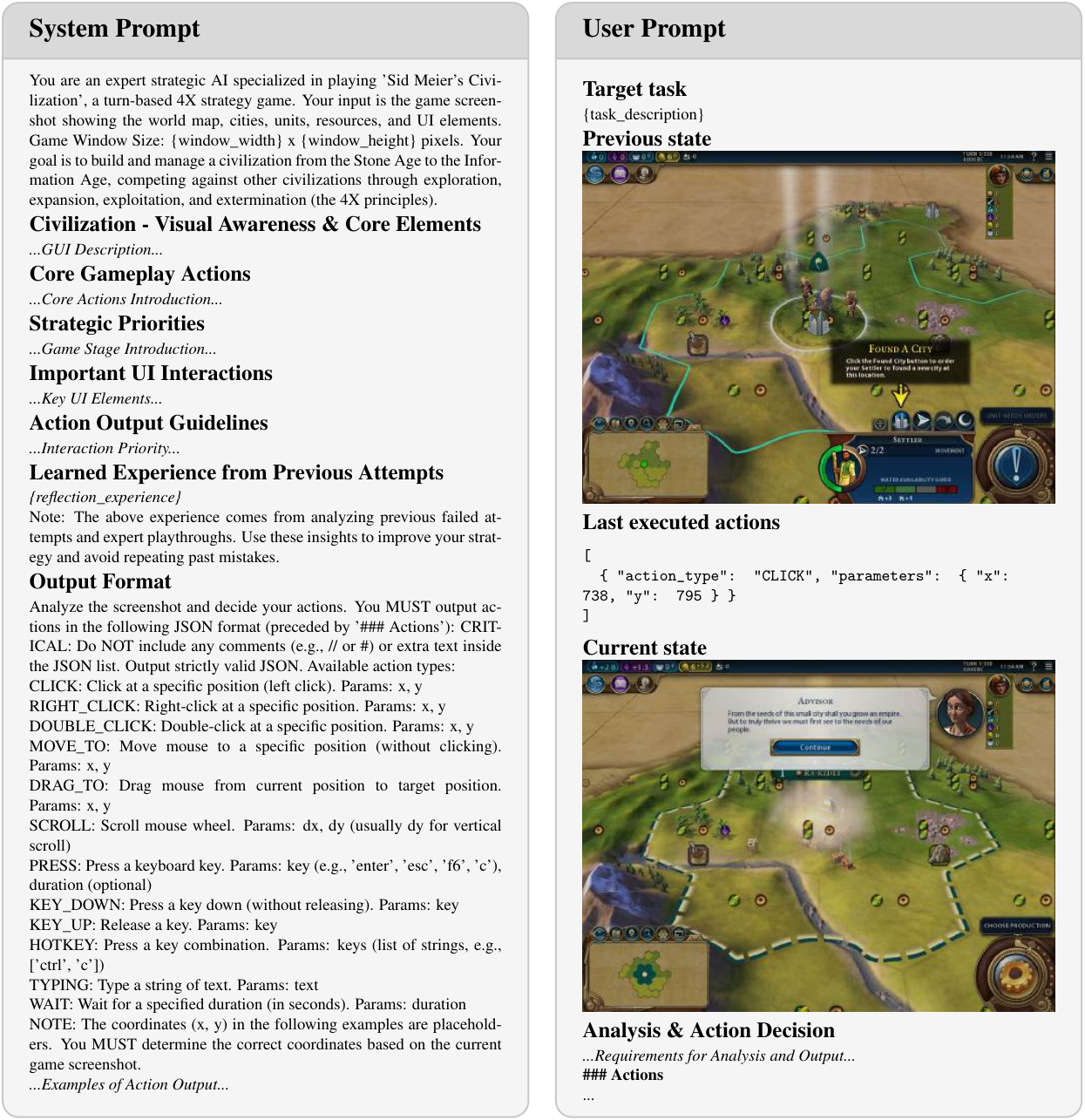}
    \caption{Civilization VI zero-shot agent prompt}
    \label{fig:civilization_zeroshot_prompt}
\end{figure}

\begin{figure}[p]
    \centering
        \includegraphics[width=\linewidth]{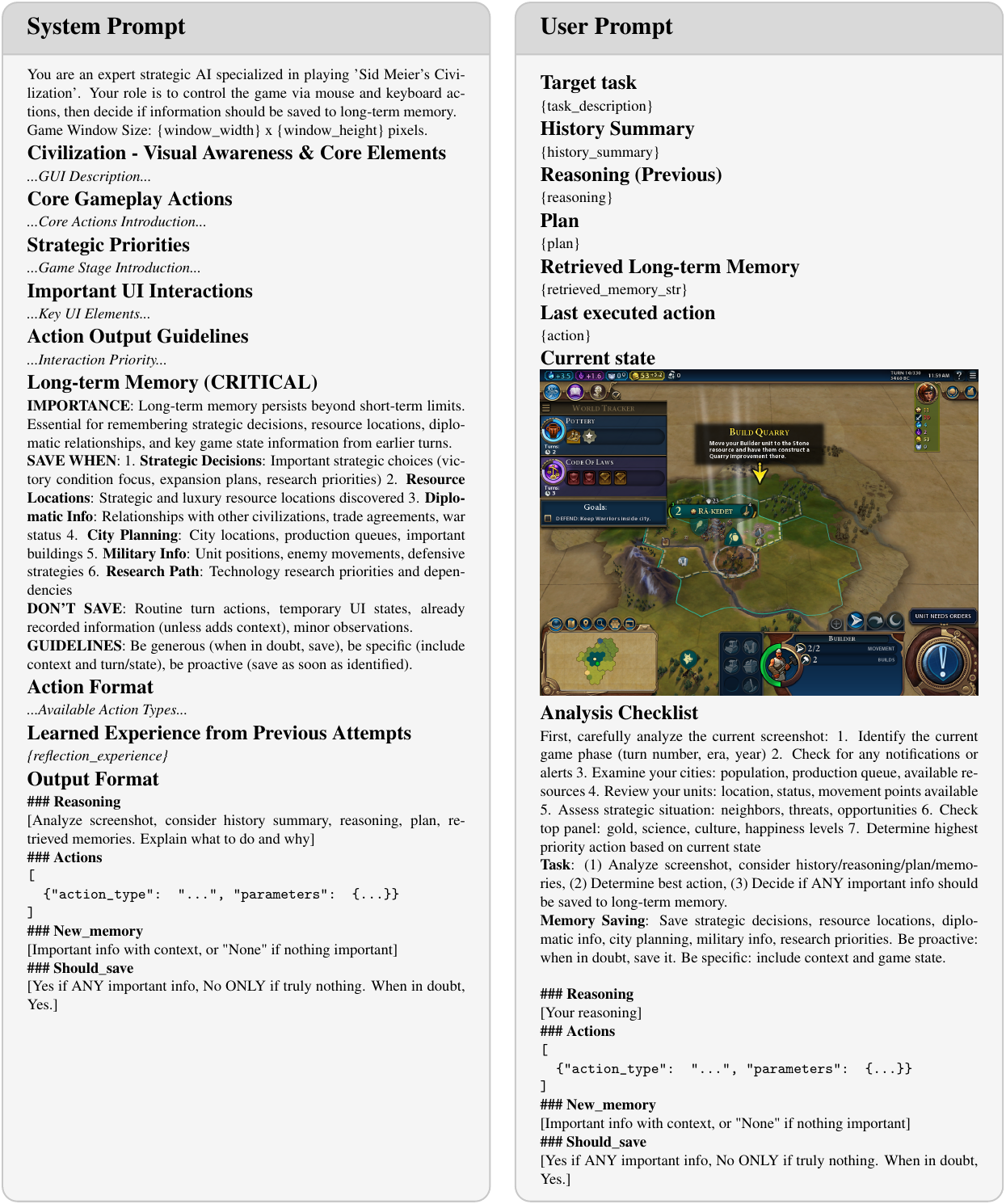}
    \caption{Civilization VI memory agent prompt}
    \label{fig:civilization_memory_prompt}
\end{figure}

Our implementation of Civilization VI uses both zero-shot agent and memory agent. We provide the full structure of our zero-shot agent prompts in Figure~\ref{fig:civilization_zeroshot_prompt} and our memory agent prompt in Figure~\ref{fig:civilization_memory_prompt}.

\subsubsection*{D.8.3 Detailed Analysis For Civilization VI}

\begin{figure}[ht]
    \centering
        \includegraphics[width=\linewidth]{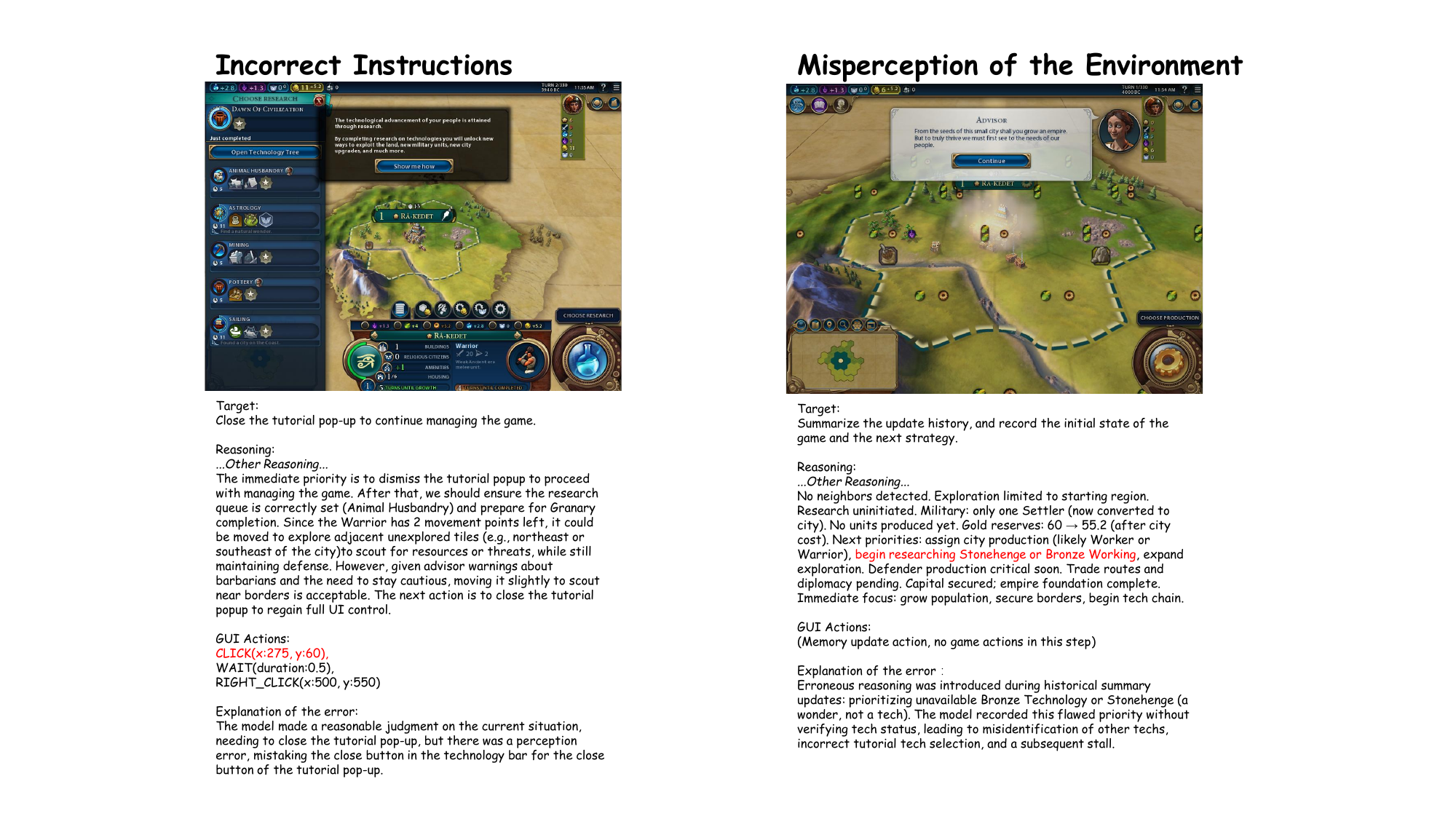}
    \caption{Civilization VI errors}
    \label{fig:civi_errors}
    \vspace{-20pt}
\end{figure}

Our experimental evaluation in \textit{Civilization VI} reveals that the stability of \textbf{GUI Grounding capability} is the foundational determinant of agent performance in this complex interactive environment, as evidenced by both quantitative results in Table~\ref{tab:civi_scores} and qualitative error patterns in Figure~\ref{fig:civi_errors}. 

\begin{table}[htbp]
    \centering
    \small
    \begin{tabular}{lcccc}
        \toprule 
        \textbf{Model} & \textbf{Zero-shot} & \textbf{Memory} & \textbf{Zero-shot VR.} & \textbf{Memory VR.} \\
        \midrule 
        Gemini-2.5-Flash & $0.0 \pm 0.0$ & $0.0 \pm 0.0$ & $0.0 \pm 0.0$ & $0.0 \pm 0.0$ \\

        Gemini-2.5-Pro & $0.0 \pm 0.0$ & $0.0 \pm 0.0$ & $0.0 \pm 0.0$ & $0.0 \pm 0.0$ \\

        Qwen3-VL-8B & $0.0 \pm 0.0$ & $0.0 \pm 0.0$ & $0.0 \pm 0.0$ & $0.0 \pm 0.0$ \\

        Qwen3-VL-32B & $19.4 \pm 4.0$ & $8.3 \pm 0.0$ & $16.7 \pm 0.0$ & $8.3 \pm 0.0$ \\

        GPT-4o & $0.0 \pm 0.0$ & $0.0 \pm 0.0$ & $0.0 \pm 0.0$ & $0.0 \pm 0.0$ \\

        GPT-4o-mini & $0.0 \pm 0.0$ & $0.0 \pm 0.0$ & $0.0 \pm 0.0$ & $0.0 \pm 0.0$ \\

        Seed-1.8 & $16.7 \pm 6.8$ & $22.2 \pm 10.4$ & $19.4 \pm 7.9$ & $19.4 \pm 7.9$ \\
        \bottomrule 
    \end{tabular}
    \vspace{5pt}
    \caption{Scores in Civilization VI (Mean $\pm$ Std)}
    \label{tab:civi_scores}
    \vspace{-15pt}
\end{table}

First, models lacking reliable GUI Grounding (e.g., GPT-4o, Gemini-2.5-Flash, Gemini-2.5-Pro, GPT-4o-mini) achieved a score of $0.00 \pm 0.00$ across all configurations. This complete failure stems from their inability to perform basic UI interactions, such as identifying and clicking tutorial pop-up buttons or navigating overlapping interface elements---a deficit that prevents them from completing even the earliest milestones. Similarly, Qwen3-VL-8B, which exhibits unstable GUI Grounding, also failed to yield valid scores, confirming that consistent spatial understanding of game interfaces is a prerequisite for meaningful progress.

For models with relatively robust GUI Grounding (Qwen3-VL-32B, Seed-1.8), additional challenges emerged in \textbf{long-term planning consistency} and \textbf{reasoning robustness}. As shown in Table~\ref{tab:civi_scores}, Qwen3-VL-32B demonstrated degraded performance when equipped with a memory mechanism (Memory: $0.083 \pm 0.000$; Memory + VR.: $0.083 \pm 0.000$) compared to its zero-shot baseline (Zero-shot: $0.194 \pm 0.040$). This counterintuitive result reflects Qwen3-VL-32B’s limitations in organizing and reasoning over complex temporal information: the memory module introduced erroneous prioritizations (e.g., targeting unavailable "Bronze Technology" or misclassifying Stonehenge as a technology, as illustrated in the right panel of Figure~\ref{fig:civi_errors}), which disrupted tutorial compliance and overshadowed the benefits of iterative refinement. In contrast, Seed-1.8 equipped with stronger deliberative reasoning capabilities—exhibited performance gains with memory augmentation (Memory + VR.: $0.194 \pm 0.079$ vs. Zero-shot: $0.167 \pm 0.068$), indicating that reliable long-term reasoning can leverage contextual memory to enhance strategic consistency.

Complementing these quantitative trends, qualitative analysis of typical errors further exposes the gaps in VLM capabilities. The left panel of Figure~\ref{fig:civi_errors} demonstrates a failure of spatial GUI understanding: the model misidentified a close button in the technology bar as the tutorial pop-up’s closure control, a mistake that halted early-game progress due to a lack of precise interface grounding. When combined with the hallucination-driven error in the right panel, these cases collectively demonstrate that current VLMs still lack two critical competencies for complex game AI: (1) \textbf{spatial intelligence} to resolve overlapping interface ambiguities, and (2) \textbf{reasoning robustness} to prevent hallucination propagation and maintain consistent long-term planning.

\subsection{Scene Investigators}
\subsubsection*{D.9.1 Game Description for Scene Investigators}

\begin{wrapfigure}{r}{0.55\textwidth}
    \centering
    \includegraphics[width=\linewidth]{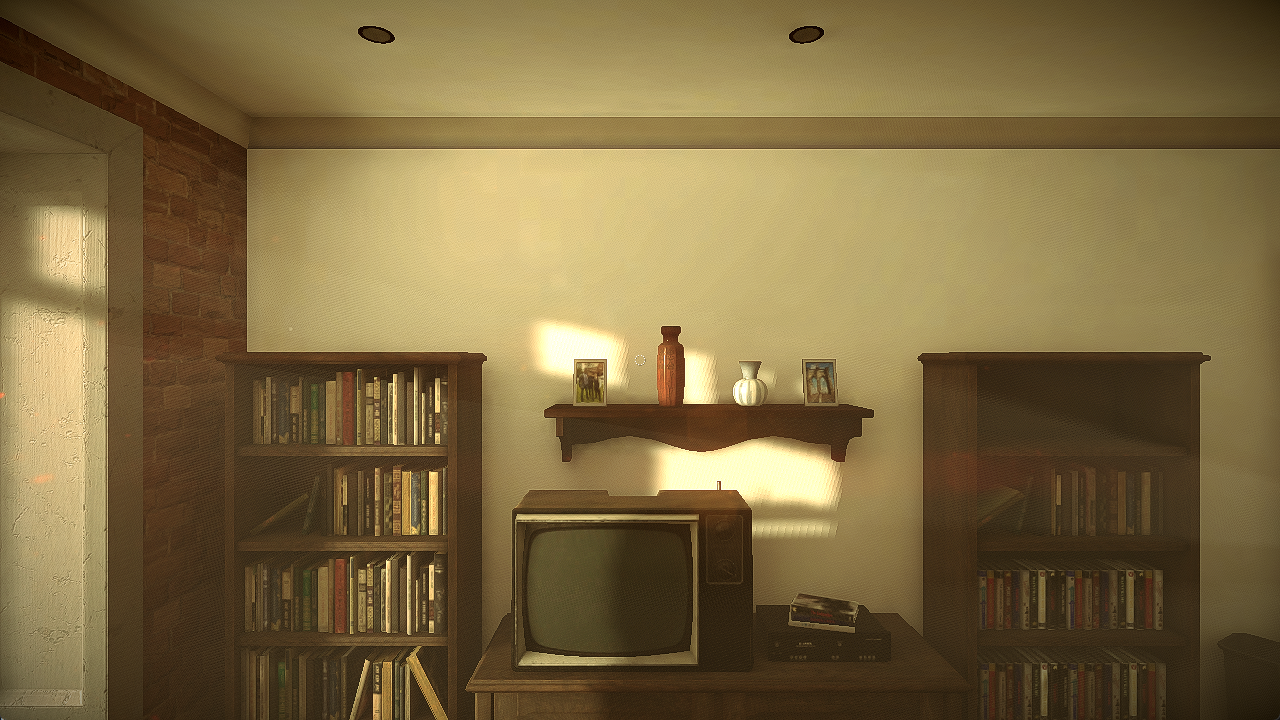}
    \caption{Screenshot of Scene Investigators}
    \label{fig:scenetitle}
\end{wrapfigure}

\textbf{Game Environments.} Scene Investigators~\cite{Scene41} is a deductive detective game set in detailed 3D crime scenes. The objective is to observe the environment meticulously to reconstruct past events. Unlike standard object detection, this game tests visual deduction and causal inference. Agents must notice subtle visual cues and link them to form a coherent narrative explanation of the crime. It evaluates the VLM's ability to act as an active investigator, synthesizing fragmented visual evidence into a logical whole. The game demands high-fidelity visual perception to identify minute details (like a timestamp on a receipt or a blood splatter pattern) and high-level reasoning to infer the sequence of events that produced them.

\noindent\textbf{(1) Game state.} High-fidelity 3D environment requiring visual deduction.

\noindent\textbf{(2) Main GUI action space.} Mouse Click and Movement (Navigate crime scenes, inspect objects).

\noindent\textbf{(3) Evaluation task.} Experiments were conducted on the Scene Investigators (Demo) environment to evaluate agent performance.  We impose a maximum limit of 1000 steps per episode. The task evaluation is structured around a series of predefined milestones, specified as: Start New Game, Discover Birthday Card, Find Letter in Drawer, Analyze Calendar Dates, Locate Guest List, Examine Purse Contents, Discover Key Logic Note, Find Medical Evidence, Inspect Wallet and ID, Scene Reconstruction Analysis, Submit Case Answers, and Level Completion. The performance of the agent was quantified by the count of milestones successfully accomplished within the Demo, where the final performance value was calculated as the mean of at least three independent trial runs. The evaluation metric was normalized by the total number of milestones within the level (i.e., 12), and the formal definition of the normalized score is given as follows:
\begin{equation*}
S_{\text{norm}} = \frac{\frac{1}{n}\sum_{i=1}^n M_i}{M_{\text{total}}} \times 100
\end{equation*}
where $M_i$ denotes the number of milestones accomplished in the $i$-th trial, $n$ represents the total number of independent trials, and $M_{\text{total}} = 12$ is the total number of predefined milestones for the Scene Investigators Demo.

\noindent\textbf{(4) Expert video content.} This expert tutorial video is a skill-oriented instructional presentation that systematically analyzes and rectifies cognitive misconceptions within the investigative puzzle-solving process. It specifically addresses logical fallacies such as aimless environmental observation, the isolation of evidentiary variables, and the failure to establish causal links between spatial clues and character identities.The video conducts an in-depth dissection of the core heuristic mechanics and the logical principles of evidence synthesis, covering critical elements such as objective-driven data retrieval, micro-textual analysis of physical artifacts, and the real-time cross-referencing of environmental cues. By demonstrating the necessity of adopting structured deductive strategies, the video guides the agent to formulate decisions based on the intrinsic logic of motive and identity verification. This approach ensures the construction of a self-consistent evidentiary loop and prevents irreversible errors in the final inferential judgment.

\subsubsection*{D.9.2 Game Prompt For Scene Investigators}
\begin{figure}[p]
    \centering
        \includegraphics[height=0.95\textheight]{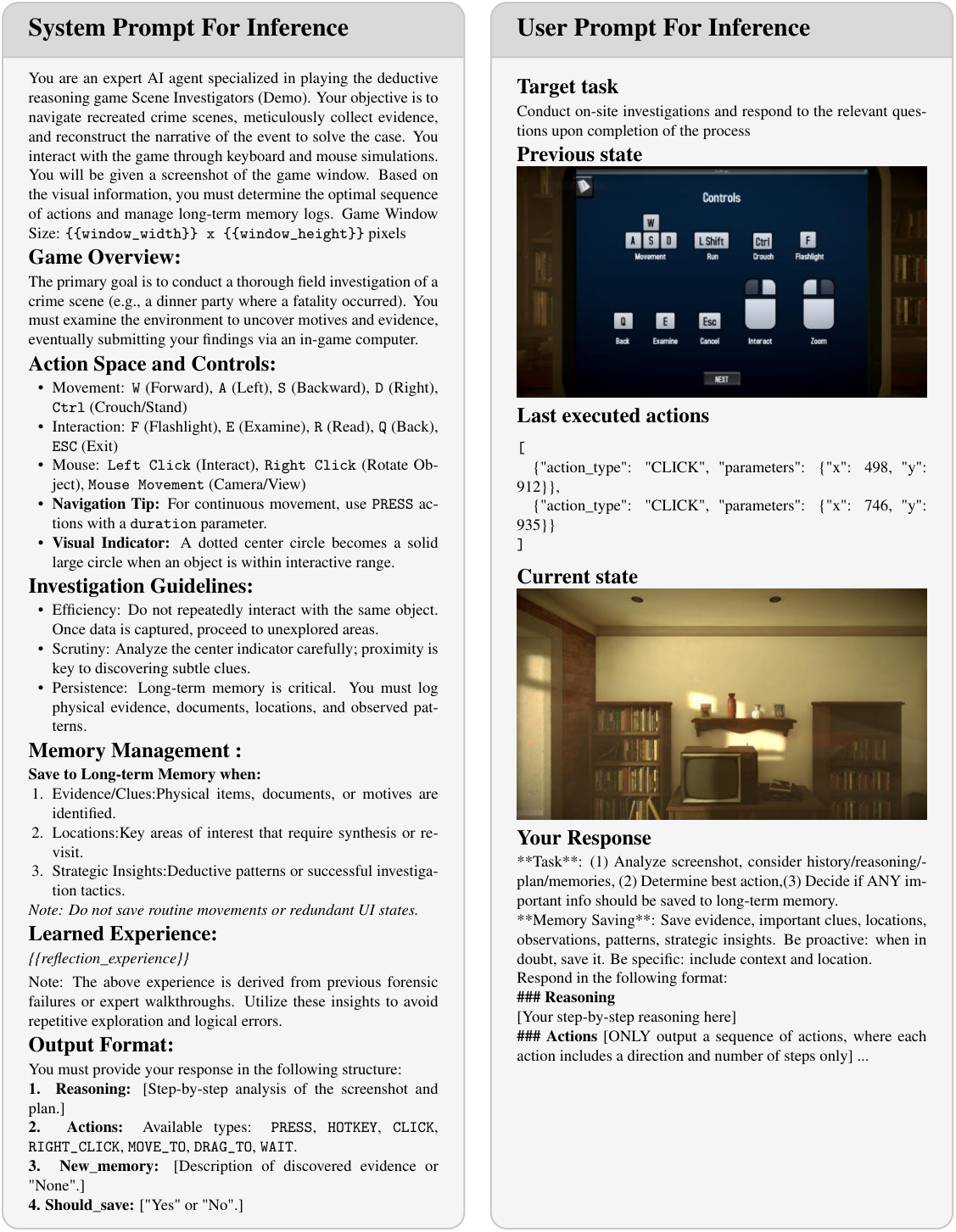}
    \caption{Scene Investigators inference prompt}
    \label{fig:Scene Investigators inference prompt}
\end{figure}

\begin{figure}[p]
    \centering
        \includegraphics[height=0.95\textheight]{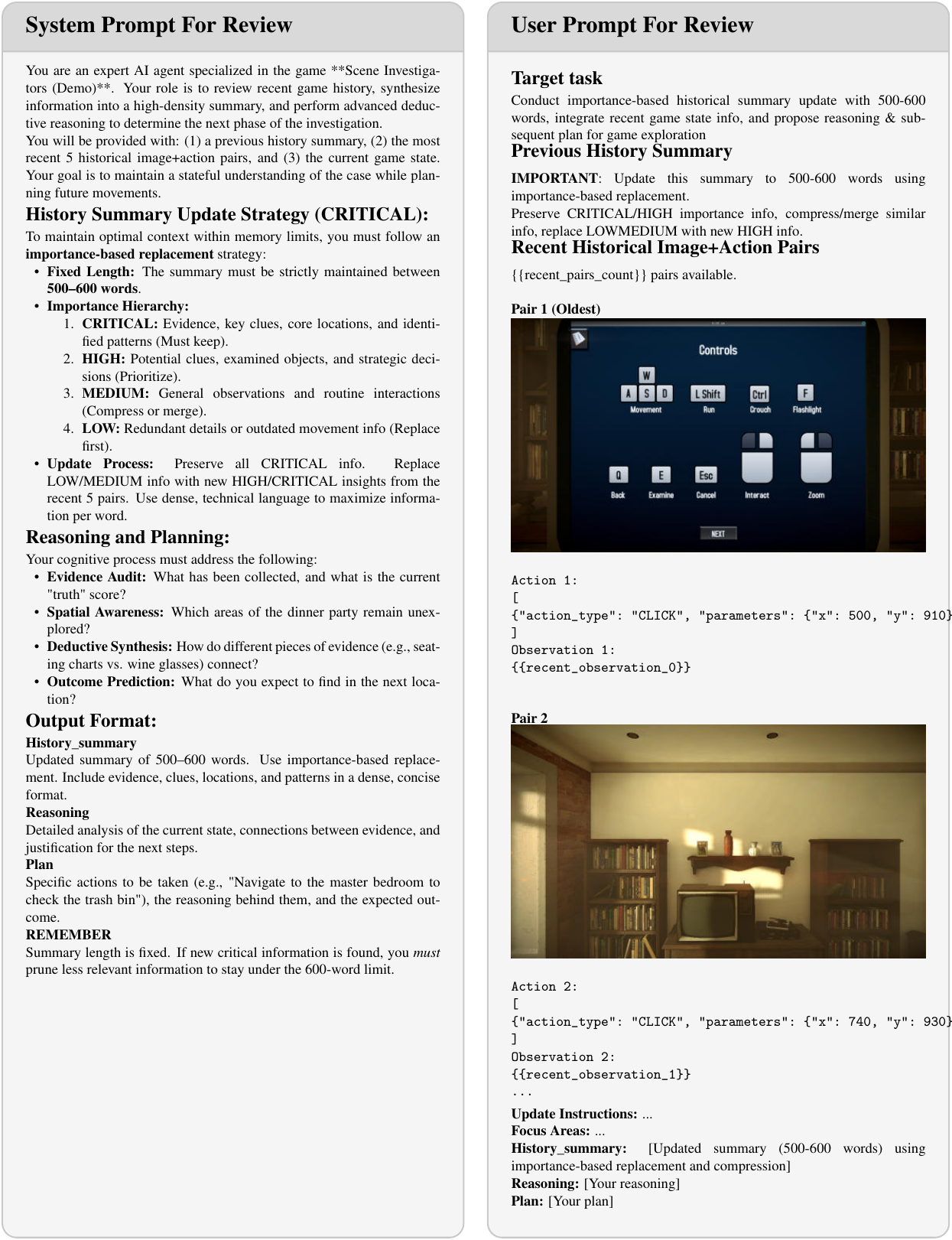}
    \caption{Scene Investigators review prompt}
    \label{fig:Scene Investigators review prompt}
\end{figure}

We provide the full structure of our prompts in Figure~\ref{fig:Scene Investigators inference prompt} and Figure~\ref{fig:Scene Investigators review prompt}.

\subsubsection*{D.9.3  Detailed Analysis For Scene Investigators}

\begin{wrapfigure}{r}{0.6\textwidth} 
  \centering
  \vspace{-5pt} 
  \small 
  \begin{tabular}{lccc}
    \toprule
   \textbf{Model} & \textbf{Zeroshot} & \textbf{Memory} & \textbf{Memory VR.} \\
    \midrule
Gemini-2.5-Flash      & --      & $2.78 \pm 3.9$     & $8.3 \pm 0.0$\\

    Gemini-2.5-Pro        & $0.0 \pm 0.0$ & $0.0 \pm 0.0$ & $0.0 \pm 0.0$  \\

    Qwen3-VL-8B           & $5.6 \pm 3.93$      & $5.6 \pm 3.93$     & $5.6 \pm 3.93$\\

    Qwen3-VL-32B          & $8.3 \pm 0.0$ & $8.3 \pm 0.0$ & $8.3 \pm 0.0$\\

    GPT-4o                & --      & $0.0 \pm 0.0$  & $0.0 \pm 0.0$\\

    GPT-4o-mini           & --      & $2.78 \pm 3.9$ & $2.78 \pm 3.9$\\

    Seed-1.8       & --      & $2.78 \pm 3.9 $ & $8.3 \pm 0.0$             \\
    \bottomrule
  \end{tabular}
  \caption{raw scores in Scene Investigators}
  \label{tab:Scene Investigators scores}
  \vspace{-10pt} 
\end{wrapfigure}

\begin{figure}[h]
    \centering
        \includegraphics[width=\linewidth]{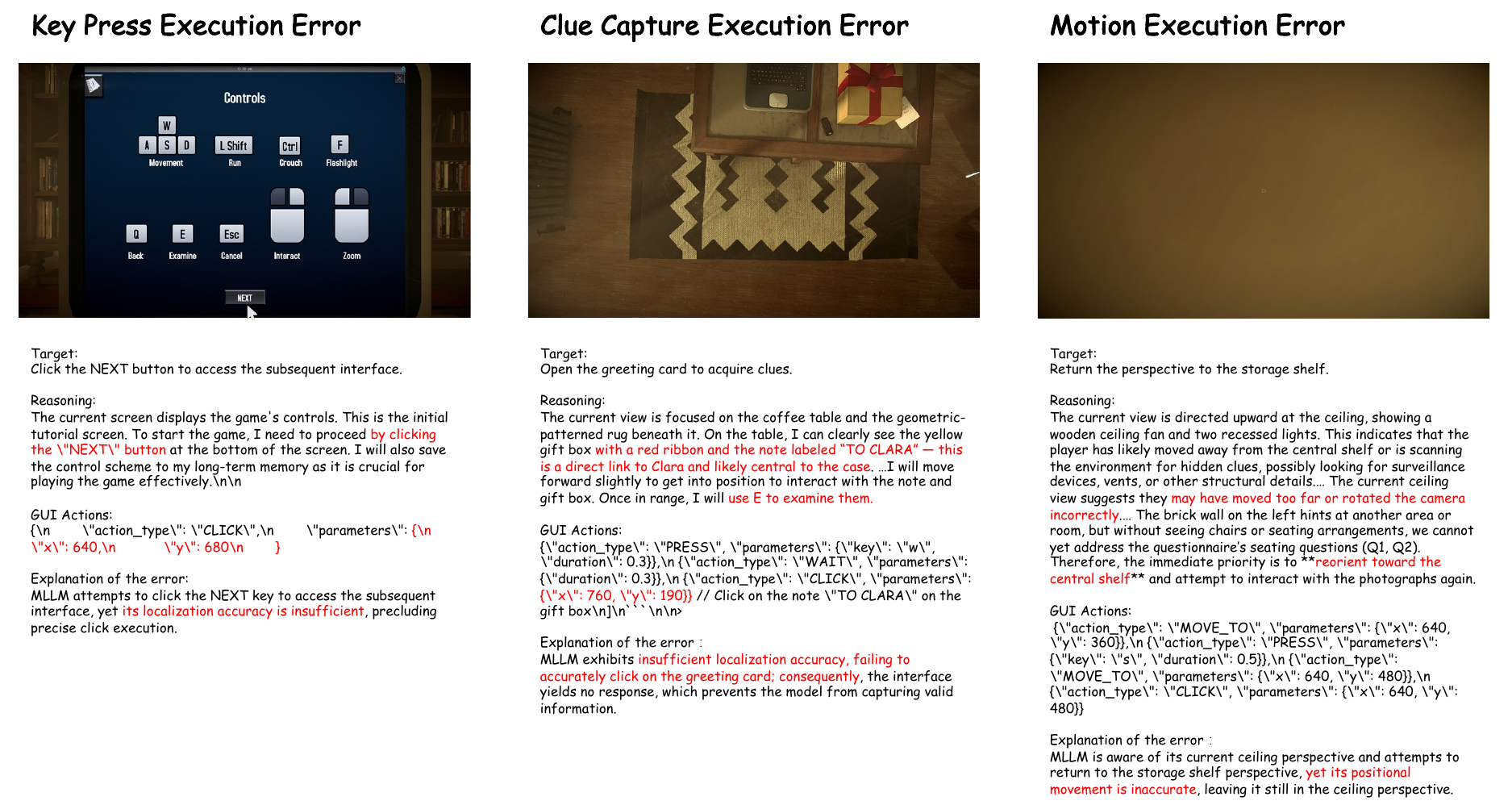}
    \caption{Scene Investigators errors}
    \label{fig:Scene Investigators errors}
\end{figure}

Experimental results on Scene Investigators (Table~\ref{tab:Scene Investigators scores}) reveal significant performance bottlenecks in agents, which are highly correlated with their capabilities in visual grounding and action execution. Visual grounding, in particular, emerges as the critical bottleneck for VLMs within this environment. Only models possessing robust pixel-level localization capabilities within the normalized game interface (e.g., Qwen3-VL-32B) demonstrate the ability to precisely manipulate the mouse to click the game initiation button. Conversely, other models (e.g., Seed-1.8, Gemini-2.5-Flash) are capable of perceiving the game interface and identifying the general location of the "START" button, moving the cursor to its vicinity. However, due to spatial deviations from the precise coordinates, these models frequently fail to establish accurate interaction. Upon successfully entering the game environment—whether through high localization precision (as seen in Qwen3-VL-32B) or stochastic success—the models generally exhibit correct control over navigation and camera perspective. Nevertheless, given the complex nature of the game environment and the stringent requirements for mouse click precision, all tested models failed to correctly interact with evidentiary objects within the scene. Figure~\ref{fig:Scene Investigators errors} illustrates three representative types of execution errors, comprehensively elucidating how insufficient pixel localization precision precipitates task failure.

\textbf{Pixel-level Localization Bias}\\
We observe a pronounced "hallucination" phenomenon regarding pixel-level localization precision in current mainstream models. Although leading VLM such as Gemini-2.5-Pro and Seed-1.8 demonstrate the capability to perceive the game interface, the stringent precision required for button actuation proves to be a limiting factor. While the coordinates predicted by these models typically cluster in the vicinity of the target, they frequently fall outside the effective interaction region. Consequently, most models achieve the first milestone (initiating a new game) only probabilistically.
During experimentation, we identified significant deviations in the pixel coordinates output by Qwen3-VL-32B and Qwen3-VL-8B prior to interface normalization. Drawing upon the Qwen3 technical report, which specifies a training image resolution of [1000, 1000], we normalized the game interface to these dimensions. This adjustment yielded a substantial improvement in button actuation accuracy. However, constrained by parameter scale, the localization precision of Qwen3-VL-8B remains inferior to that of Qwen3-VL-32B. Furthermore, upon entering the 3D game environment, the requirement for pixel localization precision regarding object interaction becomes even more rigorous; as a result, none of the tested models are currently capable of interacting with clues to advance the gameplay.

\begin{figure}[ht]
    \centering
    \small
    \includegraphics[width=\linewidth]{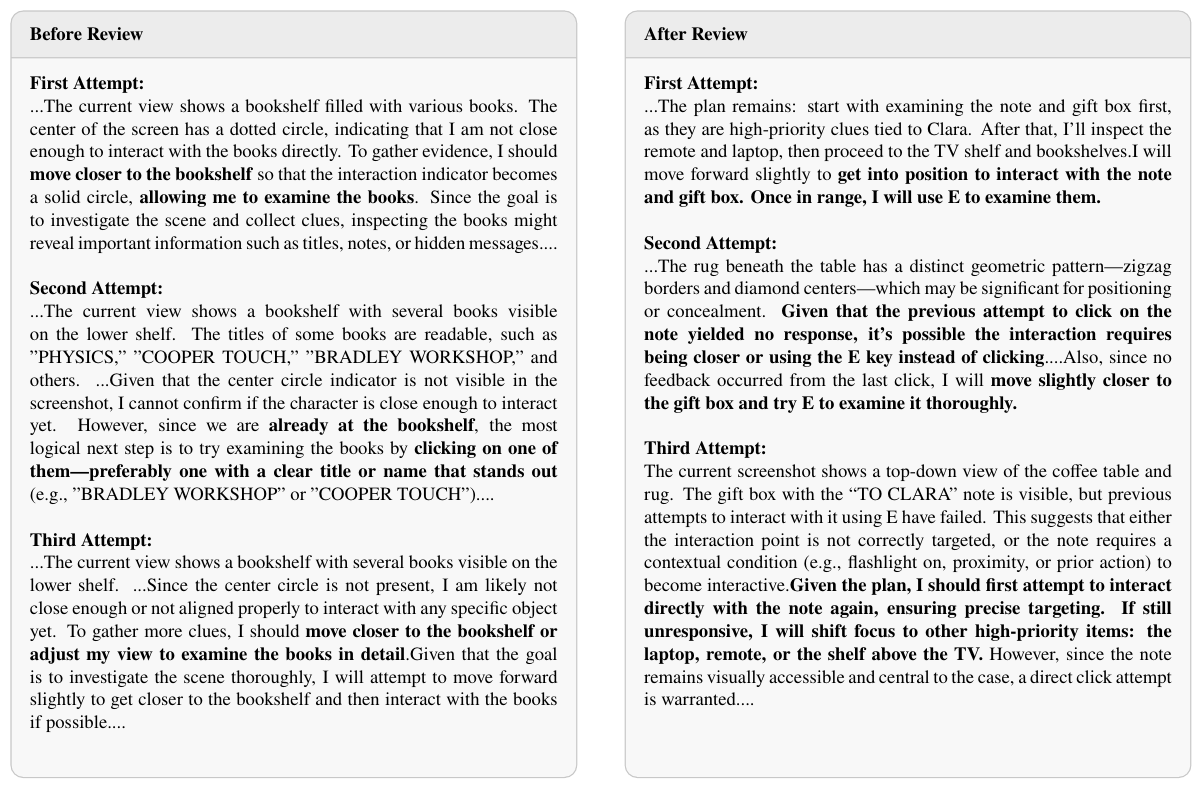}
    \caption{Qwen3-VL-32B memory improvement example}
    \label{fig:Qwen3-VL-32B_memory}
\end{figure}

\textbf{Memory Analysis} \\
Experimental results indicate that while the incorporation of a memory mechanism did not yield a statistically significant improvement in final scores, a qualitative analysis of the reasoning process—as depicted in Figure~\ref{fig:Qwen3-VL-32B_memory}—reveals critical behavioral shifts. Prior to the integration of memory, the model exhibited a tendency to engage in repetitive cognitive loops when encountering non-interactive objects. Lacking historical context, it would reiterate identical reasoning and output sequences in each observation cycle, resulting in stagnation at specific interaction points.\\
Conversely, with the memory module enabled, the model demonstrates self-correction capabilities. Following an initial failure, it calibrates its current reasoning based on prior outputs. Notably, after persistent unresponsiveness (e.g., three attempts), the model initiates a reflection process, inferring that the object is "non-interactive" and strategically deciding to "transition scenes for further investigation." In summary, the memory mechanism effectively prevents the agent from entering deadlock states caused by inherently non-interactive objects or interaction failures due to insufficient click precision, thereby enhancing the model's robustness in advancing the gameplay during case investigations.

\begin{wrapfigure}{r}{0.5\textwidth} 
    \centering
    \small
    \includegraphics[width=\linewidth]{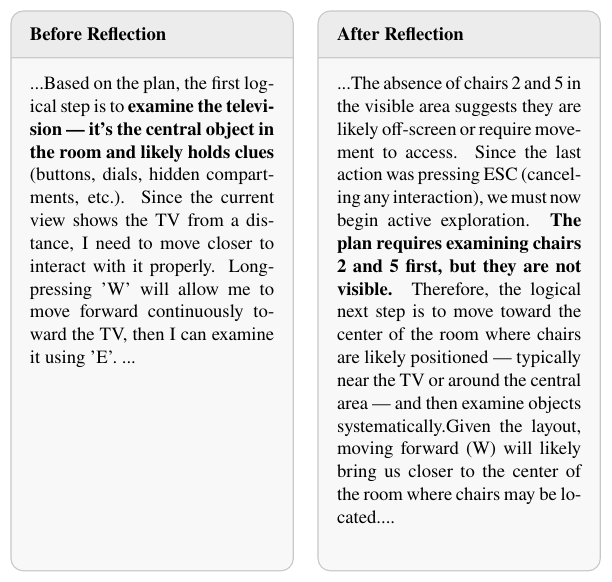}
    \caption{Qwen3-VL-8B reflection improvement example}
    \label{fig:Qwen3-VL-8B_reflection Scene}
\end{wrapfigure}

\textbf{Video Reflection Analysis}\\
Experimental results indicate that while the incorporation of the reflection mechanism did not yield a statistically significant improvement in final scores—a limitation primarily attributed to the models' bottlenecks in pixel-level localization—an analysis of the reasoning process (as shown in Figure~\ref{fig:Qwen3-VL-8B_reflection Scene}) reveals a distinct optimization in strategy. In the absence of reflection, models tended to rely on associative analysis and speculative inference linking currently observed objects to questionnaire items.\\

\newpage

In contrast, by leveraging expert videos that document the complete clue collection workflow, the model distills prior knowledge regarding objects that contain critical evidence. Consequently, with the reflection mechanism enabled, the model prioritizes investigating these target objects upon entering the scene, thereby significantly reducing the temporal overhead associated with filtering out irrelevant distractors.

\subsection{Snake}

\subsubsection*{D.10.1 Game Description for snake}

\textbf{Game Environments.} 
Snake~\cite{snake40} is a continuous, real-time arcade environment necessitating immediate reaction to visual input. The agent controls a growing line within a bounded grid, where the core mechanic is characterized by diminishing fault tolerance: as the agent successfully consumes food, the snake's length increases, progressively restricting navigable space and amplifying the probability of fatal collisions. This environment primarily evaluates the VLM's capacity for dynamic risk-reward management—balancing the aggressive pursuit of rewards against the immediate necessity of survival. Snake serves as a benchmark for inference latency and perception speed. As a real time game environment, high inference latency inherent to this game leads to immediate collisions. Given the environment's relatively low visual complexity, this benchmark emphasizes the need for sparse reasoning—evaluating whether the VLM can efficiently parse simple inputs and render decisions without expending excessive inference time on trivial states.

\begin{wrapfigure}{r}{0.35\textwidth}
    \centering
    \vspace{-20pt}
    \includegraphics[width=\linewidth]{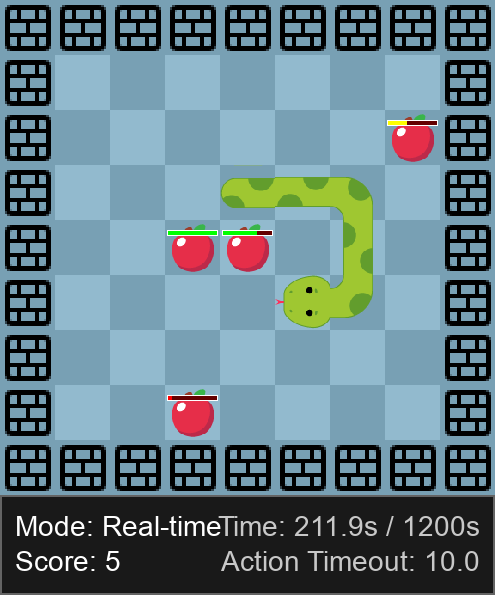}
    \caption{Screenshot of Snake}
    \label{fig:snake game title}
\end{wrapfigure}

\noindent\textbf{(1) Game state.} Real-time, contained grid environment.

\noindent\textbf{(2) Main GUI action space.} Keyboard Input (Directional arrow keys).

\noindent\textbf{(3) Evaluation task} Control the snake to eat food and avoid obstacles.  We impose a maximum limit of 100 steps per episode. The agent's performance is quantified by the total number of food items consumed by the snake prior to collision with an obstacle or its own body. The normalized score is calculated by taking the average number of food items consumed across $n$ independent trials and dividing it by a scaling factor of 10, which is defined as follows: 

\begin{equation*}
    \bar{S}_{norm} = \frac{1}{10n} \sum_{i=1}^{n} S_i
\end{equation*}

\noindent\textbf{(4) Play-through video content} This play through video presents a complete game replay demonstrating the optimal space arrangement strategy,which shows the snake successfully navigating to fully occupy the grid and reach the map's maximum score limit of 81

\subsubsection*{D.10.2 Game Prompt For Snake} 

\begin{figure}[p]
    \centering
        \includegraphics[height=0.95\textheight]{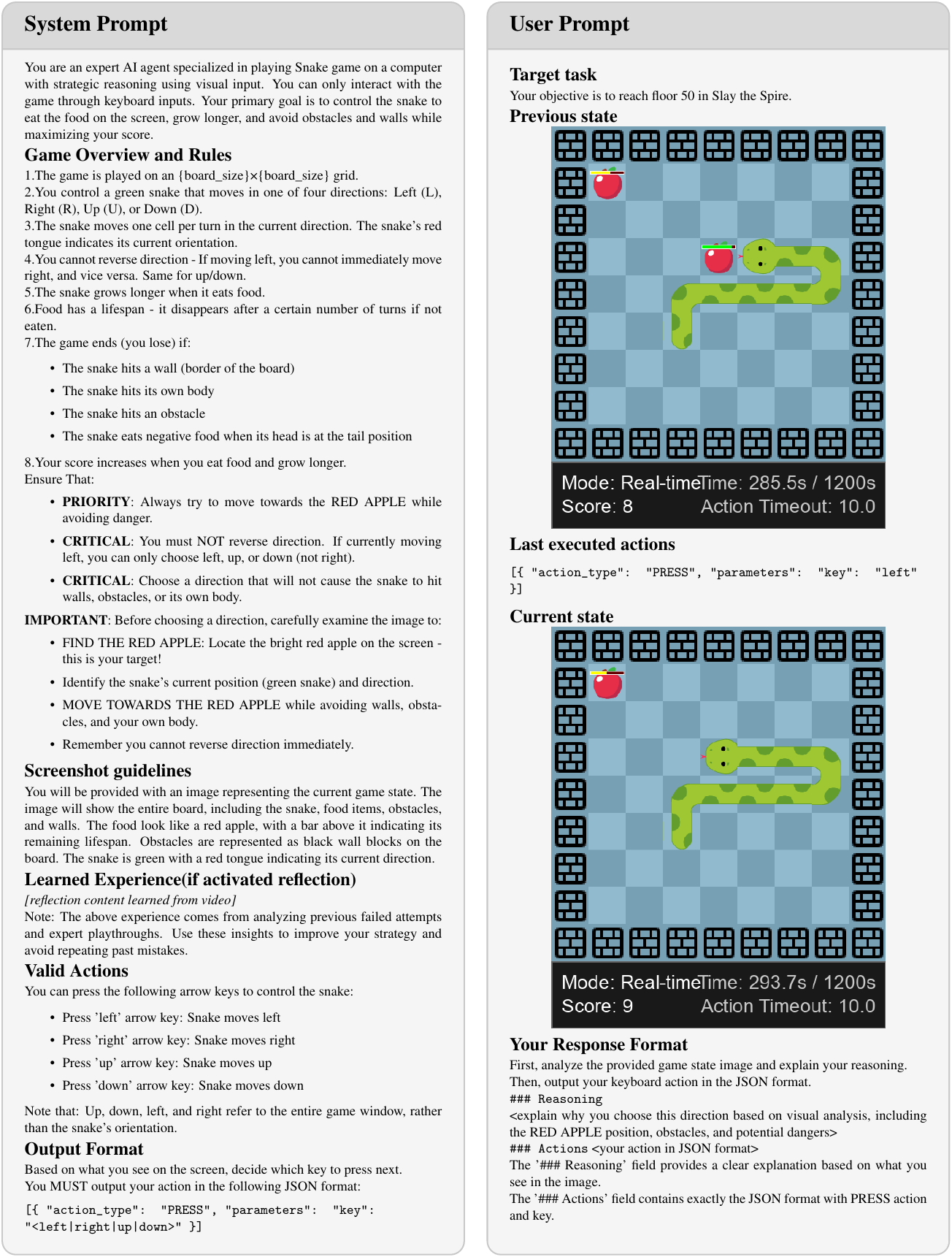}
    \caption{snake prompt}
    \label{fig:snake prompt}
\end{figure}

We provide the full structure of our prompts in Figure~\ref{fig:snake prompt}.

\subsubsection*{D.10.3 Detailed Analysis for snake}

\begin{wraptable}{r}{0.5\textwidth} 
    \centering
    \vspace{-10pt}
    \small 
    \setlength{\tabcolsep}{6pt} 
    \begin{tabular}{l c c}
        \toprule
        \textbf{Model} & \textbf{GUI} & \textbf{GUI VR.} \\
        \midrule
        Gemini-2.5-Flash & $2.00 \pm 1.80$ & $1.00 \pm 0.81$ \\
        Gemini-2.5-Pro & $2.38 \pm 3.00$ & $6.75 \pm 3.20$ \\
        Qwen3-VL-8B & $0.25 \pm 0.71$ & $0.00 \pm 0.00$ \\
        Qwen3-VL-32B & $4.38 \pm 2.61$ & $3.25 \pm 1.71$ \\
        GPT-4o & $0.00 \pm 0.00$ & $0.00 \pm 0.00$ \\
        GPT-4o-Mini & $0.00 \pm 0.00$ & $0.00 \pm 0.00$ \\
        Seed-1.8 & $0.75 \pm 1.00$ & $1.00 \pm 1.15$ \\
        \bottomrule
    \end{tabular}
    \vspace{5pt}
    \caption{Main Experiment Results(w/ vs. w/o reflection).}
    \label{tab:snake raw result}
    \vspace{-10pt}
\end{wraptable}

The performance of GPT-4o and GPT-4o-mini in the Snake environment was unexpectedly poor. These models exhibited a fundamental inability to comprehend the relative spatial relationship between the agent (the snake) and the target (food), as illustrated in the typical error examples. Furthermore, when encountering obstacles, these models frequently violated game constraints by attempting to reverse direction, indicating significant deficiencies in both spatial reasoning and instruction adherence.

\begin{figure}[ht]
    \centering
        \includegraphics[width=\linewidth]{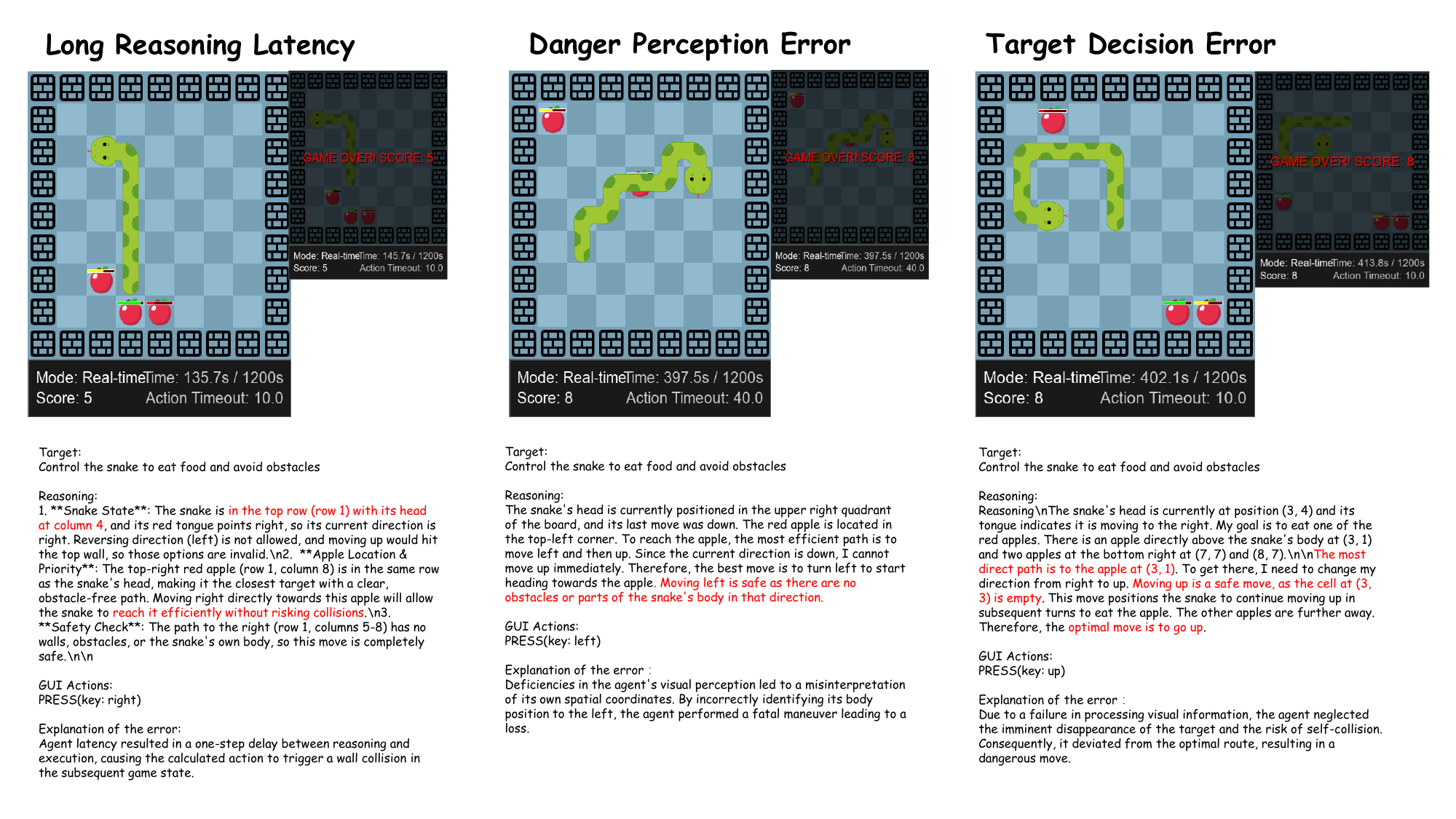}
    \caption{snake typical errors}
    \label{fig:snake errors}
\end{figure}

In contrast, Qwen3-VL-8B and Gemini-2.5-Flash demonstrated limited spatial perception, with the frequency of erroneous instructions increasing notably as the snake's body length extended. Qwen3-VL-32B exhibited significantly superior performance, maintaining score stability throughout the gameplay. However, Gemini-2.5-Pro and Seed-1.8 were severely constrained by latency-induced errors, resulting in suboptimal performance overall (see \cref{fig:snake errors} Left).

\begin{wraptable}{r}{0.45\textwidth} 
    \centering
    \setlength{\tabcolsep}{10pt} 
    \begin{tabular}{l c c c}
        \toprule
        \textbf{Model} & \textbf{10s} & \textbf{20s} & \textbf{Stop} \\
        \midrule
        Qwen3-VL-32B & $4.38$ & $4.75$ & $5.33$ \\
        GPT-4o & $0.00$ & $0.50$ & $0.00$ \\
        Gemini-2.5-Pro & $2.38$ & $6.50$ & $7.33$ \\
        Seed-1.8 & $0.75$ & $0.86$ & $9.00$ \\
        \bottomrule
    \end{tabular}
    \vspace{8pt}
    \caption{Latency Analysis}
    \label{tab:snake_latency_analysis}
\end{wraptable}

\textbf{Latency Error Analysis} \\ 
Our experiments reveal a critical misalignment between model inference depth and the actual cognitive demands of the task. While the Snake environment can be effectively navigated using straightforward spatial heuristics, advanced models lack the adaptive capacity to autonomously regulate their reasoning density (i.e., employing "sparse reasoning" when appropriate).  This rigidity leads to a "reasoning-speed mismatch," where the time consumed by complex Chain-of-Thought processes causes the agent to fail strict real-time temporal constraints.

This limitation is most paradigmatically illustrated by Seed-1.8, which demonstrated pronounced latency-induced failures. The model's single-step inference time frequently exceeded 20 seconds, creating excessive computational overhead that rendered it unsuitable for the real-time, small-grid dynamics of Snake. However, notably, the model performed robustly in the static "stop" delay perception experiments (see Table \ref{tab:snake_latency_analysis}). This performance discrepancy confirms that while the model possesses strong underlying reasoning capabilities, its inability to dynamically inhibit overly complex chain-of-thought processes creates a fatal bottleneck in time-sensitive scenarios.

\textbf{Video Reflection Analysis} \\ 
Models such as GPT-4o, GPT-4o-mini, and Qwen3-VL-8B failed to achieve performance gains through reflection. Due to inherent limitations in multimodal understanding and reasoning, these models were unable to extract actionable insights from either their own failure history or expert demonstration videos.

\begin{wrapfigure}{r}{0.4\textwidth} 
    \centering
    \vspace{-10pt}
    \small
    \includegraphics[width=\linewidth]{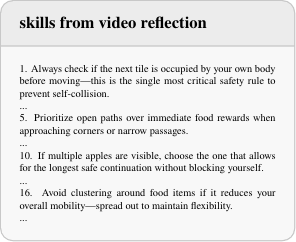}
    \caption{Qwen3-VL-32B reflection content}
    \label{fig:Qwen3-VL-32B_reflection content in snake}
\end{wrapfigure}

Conversely, models such as Qwen3-VL-32B and Gemini-2.5-Flash exhibited stagnant or slightly declined scores following video-based reflection learning. An analysis of their reflective outputs indicates that these models successfully internalized valid strategic principles, such as "Prioritize open paths over immediate food rewards" and "Avoid clustering around food items to preserve mobility."(see \cref{fig:Qwen3-VL-32B_reflection content in snake}) However, a disconnect remained between strategy formulation and execution: limited by visual perception errors (see \cref{fig:snake errors}, middle), the models frequently ignored obstacles or the snake’s own body despite correctly identifying the optimal strategy.

Notably, Gemini-2.5-Pro effectively leveraged video reflection to achieve "sparse reasoning." By adopting direct heuristics rather than engaging in redundant ab initio reasoning, the model significantly improved its reaction speed. Post-reflection, it achieved performance levels comparable to the baseline established in the static "stop" experiments.

\subsection{Plants vs. Zombies}

\subsubsection*{D.11.1 Game Description for Plants vs. Zombies}

\begin{wrapfigure}{r}{0.4\textwidth}
\vspace{-40pt}
    \centering
    \includegraphics[width=\linewidth]{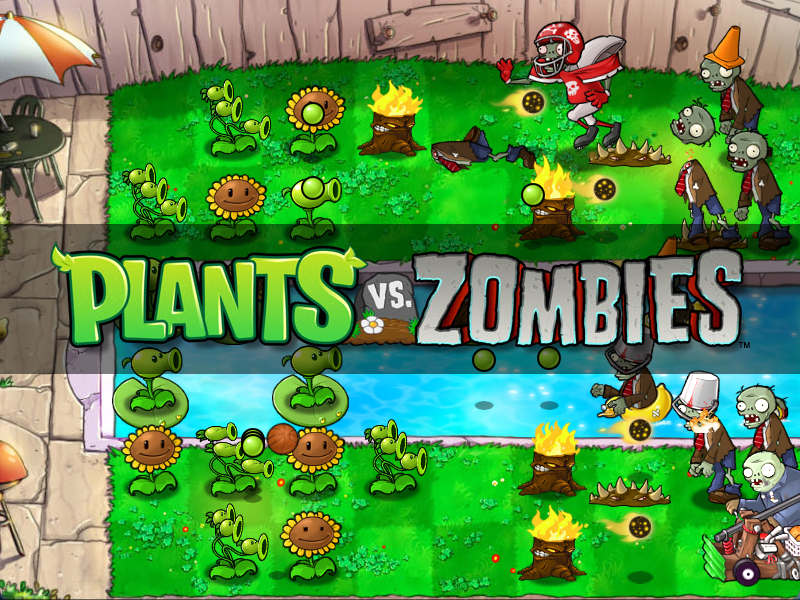}
    \caption{Screenshot of Plant vs. Zombies}
    \label{fig:pvz game title}
\end{wrapfigure}

\textbf{Game Environments.} Plants vs. Zombies is a strategic tower defense game where players must defend a home from incoming waves of enemies by placing plants with specific functions. The environment requires managing a generated resource (sunlight) and spatial positioning. It tests tactical positioning and real-time resource allocation. Agents must recognize enemy types and counter them with the appropriate unit composition while optimizing the economy. This game evaluates the agent's ability to multitask in real-time: collecting resources, monitoring multiple lanes of attack, and deploying units strategically. It combines the pressure of real-time dynamics with the need for thoughtful combinatorial strategy to survive escalating waves of threats.

\noindent\textbf{(1) Game state.} Real-time 2D grid-based defense layout.

\noindent\textbf{(2) Main Semantic action space.} Plant, Collect, and Wait.

\noindent\textbf{(3) Main GUI action space.} Mouse Click (Collect resources, place units).

\noindent\textbf{(4) Evaluation task.}Complete the three levels: 1-1, 1-2, and 1-4. We impose a maximum limit of 50 steps per episode. The agent's performance is quantified by the average number of completed milestones. The milestones are generated from the expert video and are listed in Table~\ref{tab:pvz_milestone}. To obtain a standardized score, the number of completed milestones is normalized by the total number of milestones, and the calculation formula is as follows:

\begin{equation*}
S_{norm} = \frac{N_{\text{completed}}}{N_{\text{total}}} \times 100
\end{equation*}

where \( N_{\text{completed}} \) denotes the number of milestones completed by the agent, and \( N_{\text{total}} \) denotes the total number of milestones (19 in total, from three levels, as listed in Table~\ref{tab:pvz_milestone}). The normalized score ranges from 0 to 100.

\begin{table}[h]
  \centering
  \small
  \caption{Milestones of Plants vs. Zombies (Levels 1-1, 1-2, and 1-4)}
  \label{tab:pvz_milestone}
  \begin{tabular}{rl p{0.2cm} rl p{0.2cm} rl} 
    \toprule
    \multicolumn{2}{c}{\textbf{Level 1-1}} & & \multicolumn{2}{c}{\textbf{Level 1-2}} & & \multicolumn{2}{c}{\textbf{Level 1-4}} \\
    \cmidrule(r){1-2} \cmidrule(lr){4-5} \cmidrule(l){7-8}
    \textbf{No.} & \textbf{Milestone Title} & & \textbf{No.} & \textbf{Milestone Title} & & \textbf{No.} & \textbf{Milestone Title} \\
    \midrule
    1 & Planting First Peashooter     & & 1 & First Economic Setup         & & 1 & Level Start \\
    2 & First Sun Collection          & & 2 & First Enemy Wave Warning     & & 2 & First Economic Setup \\
    3 & Planting Second Peashooter    & & 3 & First Defense Deployment     & & 3 & First Defensive Action \\
    4 & First Zombie Appearance       & & 4 & Huge Wave Alert              & & 4 & Countering Buckethead \\
    5 & Defeating First Zombie        & & 5 & Final Wave Start             & & 5 & Huge Wave Event \\
    6 & Final Wave Triggered          & & 6 & Level Complete \& Unlock     & & 6 & Level Complete \& Shovel \\
    7 & Level Complete \& Sunflower   & &   &                              & &   & \\
    \bottomrule
  \end{tabular}
\end{table}

\noindent\textbf{(5) Expert video content.} The video demonstrates how to pass through three levels, including strategies for plant cultivation and the management of sunlight resources.

\subsubsection*{D.11.2 Game Prompt for Plants vs. Zombies}
\begin{figure}[p]
    \centering
        \includegraphics[width=\linewidth]{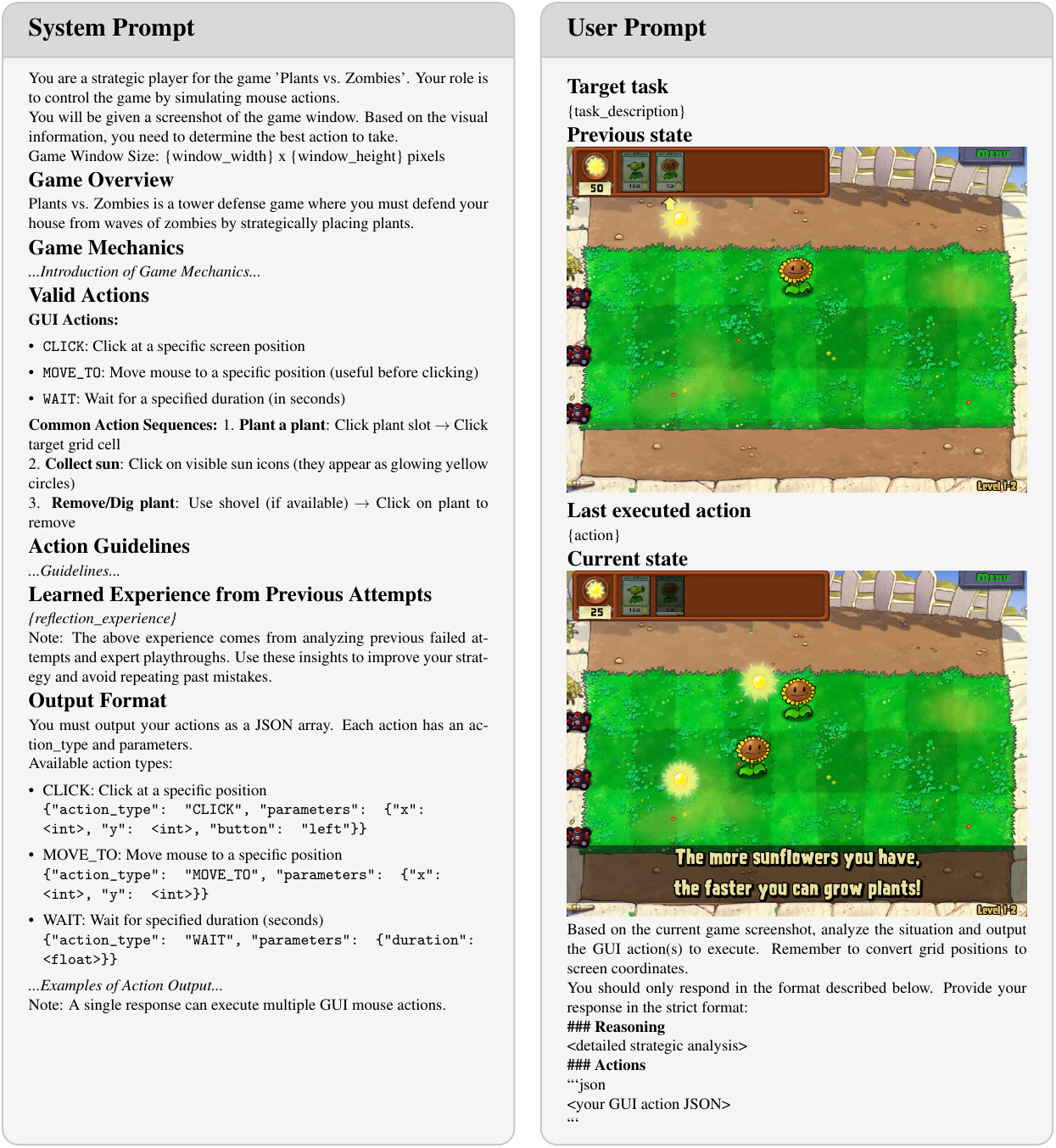}
    \caption{Plants vs. Zombies GUI prompt}
    \label{fig:pvz_gui_prompt}
\end{figure}

\begin{figure}[p]
    \centering
        \includegraphics[width=\linewidth]{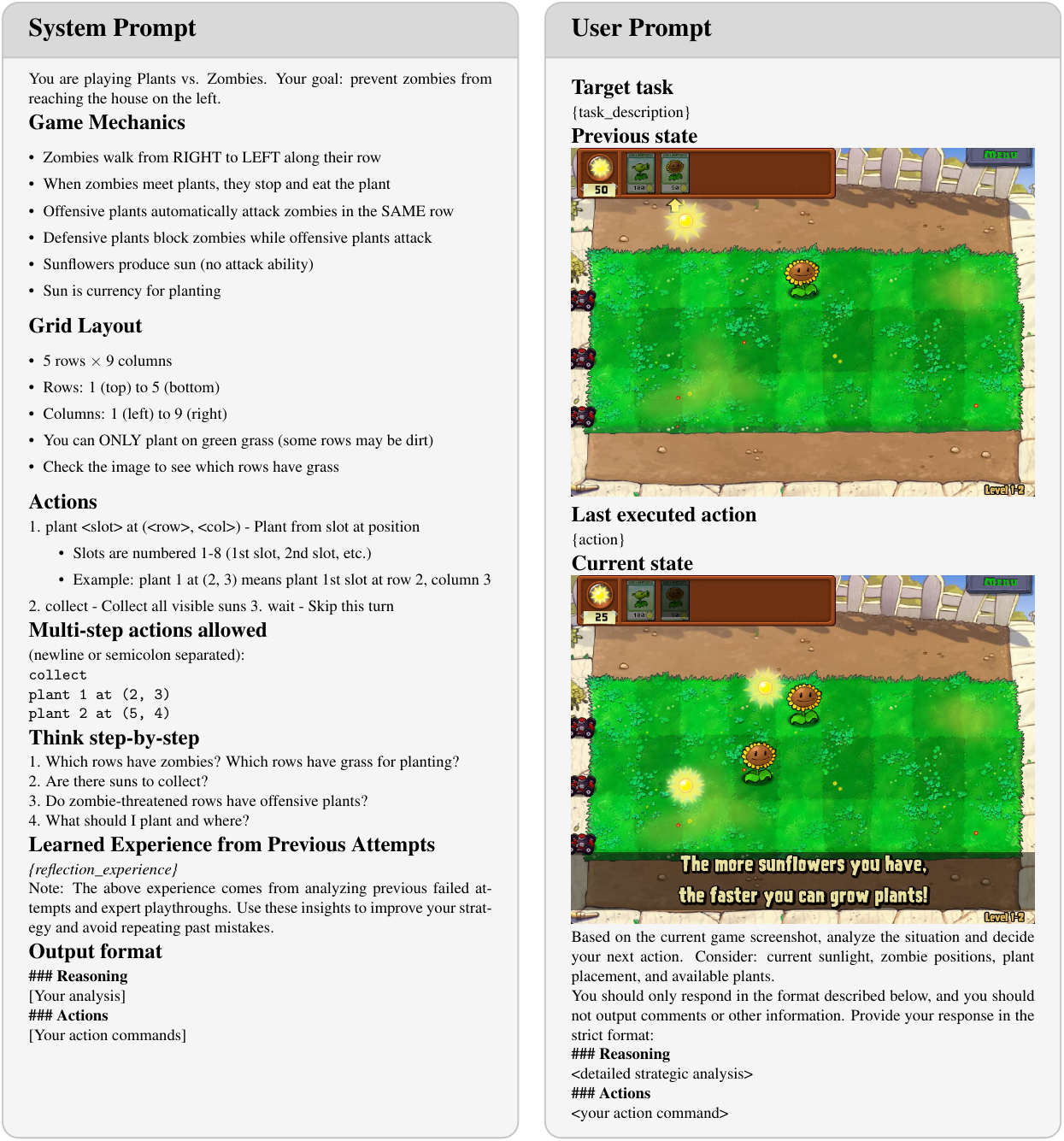}
    \caption{Plants vs. Zombies semantic prompt}
    \label{fig:pvz_semantic_prompt}
\end{figure}

Our implementation of Plant vs. Zombies uses zero-shot agent.
We provide the full structure of our prompts for GUI mode (Figure~\ref{fig:pvz_gui_prompt}) and for semantic mode (Figure~\ref{fig:pvz_semantic_prompt}).

\subsubsection*{D.11.3 Detailed Analysis for Plants vs. Zombies}

\begin{figure}[ht]
    \centering
        \includegraphics[width=\linewidth]{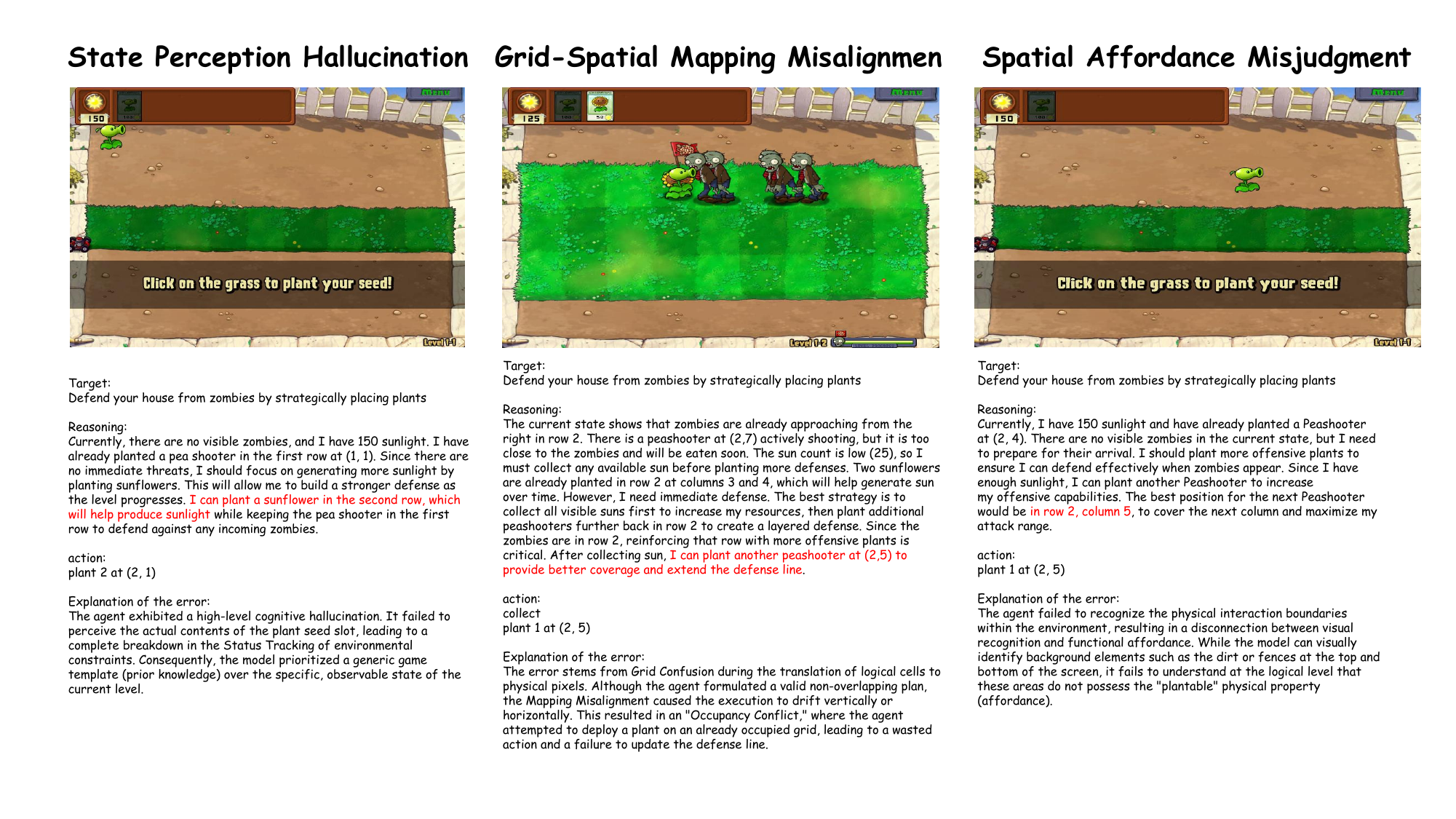}
    \caption{Plant vs. Zombies Typical Errors}
    \label{fig:pvz errors}
\end{figure}

The Plants vs. Zombies (PvZ) environment highlights the distinct challenges VLMs face when integrating resource management with real-time spatial defense. In the \textit{Plants vs. Zombies} experiments (Table~\ref{tab:accuracy_comparison}), strong reactive models such as GPT-4o, Qwen3-VL-32B and Gemini-2.5-Flash perform well. As illustrated in Figure~\ref{fig:pvz errors}, failures predominantly manifest as \textbf{State Perception Hallucination}, \textbf{Grid-Spatial Mapping Misalignment}, and \textbf{Spatial Affordance Misjudgment}.

\begin{table}[htbp]
\centering
\caption{Model Accuracy Comparison: GUI vs. Semantic (with and without VR)}
\label{tab:accuracy_comparison}
\begin{tabular}{lcccc}
\toprule
\textbf{Model} & \textbf{GUI} & \textbf{GUI VR.} & \textbf{Semantic} & \textbf{Semantic VR.} \\ 
\midrule
Qwen3-VL-8B      & 33.4 $\pm$ 4.1  & 51.3 $\pm$ 21.4          & 54.2 $\pm$ 5.8  & 65.2 $\pm$ 11.3          \\
Qwen3-VL-32B     & 41.2 $\pm$ 12.6 & 48.3 $\pm$ 19.0          & -               & -                        \\
GPT-4o-mini      & 27.6 $\pm$ 14.0 & 22.4 $\pm$ 10.4          & -               & -                        \\
GPT-4o           & 32.2 $\pm$ 12.4 & 36.4 $\pm$ 15.2          & 42.7 $\pm$ 16.0 & 53.4 $\pm$ 7.8           \\
Seed-1.8         & 26.1 $\pm$ 8.2  & 22.1 $\pm$ 10.3          & -               & -                        \\
Gemini-2.5-Flash & 51.1 $\pm$ 13.2 & \textbf{55.3 $\pm$ 10.4} & 70.2 $\pm$ 8.5  & \textbf{81.8 $\pm$ 17.0} \\
Gemini-2.5-Pro   & 33.4 $\pm$ 17.2 & 30.3 $\pm$ 16.4          & 45.4 $\pm$ 3.3  & 52.7 $\pm$ 15.2          \\
\bottomrule
\end{tabular}
\end{table}

\textbf{State Perception Hallucinations and Environmental Constraint Violations.} \\
A critical failure mode involves the agent prioritizing parametric knowledge over visual evidence. In the "State Perception Hallucination" instance, the agent attempts to execute a standard opening strategy: "plant a sunflower in the second row". However, this action is technically impossible as the current seed slot only contains a Peashooter card. This error represents a "high-level cognitive hallucination" where the model’s internal prior (that Sunflowers are essential early-game units) overrides the specific, observable state of the interface. The model fails to perform the requisite "Status Tracking" of the seed bank, resulting in a hallucinated action plan that ignores immediate inventory constraints.

\textbf{Grid-Spatial Mapping Misalignment.} \\
Agents struggled to accurately map logical grid coordinates (e.g., Row 2, Col 8) to precise 2D pixel coordinates on the isometric lawn. \textbf{Coordinate Drift:} Significant discrepancies were observed between the intended action and the actual deployment location, often resulting in plants being placed in incorrect lanes or failing to reach the intended defensive position. \textbf{Impact of Perspective:} The inability to account for the game's specific spatial projection meant that even sound strategic intentions failed to translate into effective tactical layouts.

\textbf{Video Reflection Analysis: Strategy Optimization vs. Execution Latency.} \\ 
The introduction of expert-guided reflection led to noticeable improvements in strategic depth, though execution remained a bottleneck. \textbf{Paradigm Shift:} Following exposure to expert experiences, models shifted from random placement to structured layouts, such as prioritizing Sunflowers in the rear ranks and Peashooters in forward positions. \textbf{Execution Latency:} A persistent issue of temporal lag was observed between reasoning and execution. In the dynamic environment of \textit{Plants vs. Zombies}, this latency often rendered commands obsolete by the time they were executed, as the tactical situation had already shifted.

\subsection{Forza Horizon 5}
\subsubsection*{D.12.1.Game Description for Forza Horizon 5}
\begin{wrapfigure}{r}{0.55\textwidth}
\vspace{-10pt}
    \centering
    \includegraphics[width=\linewidth]{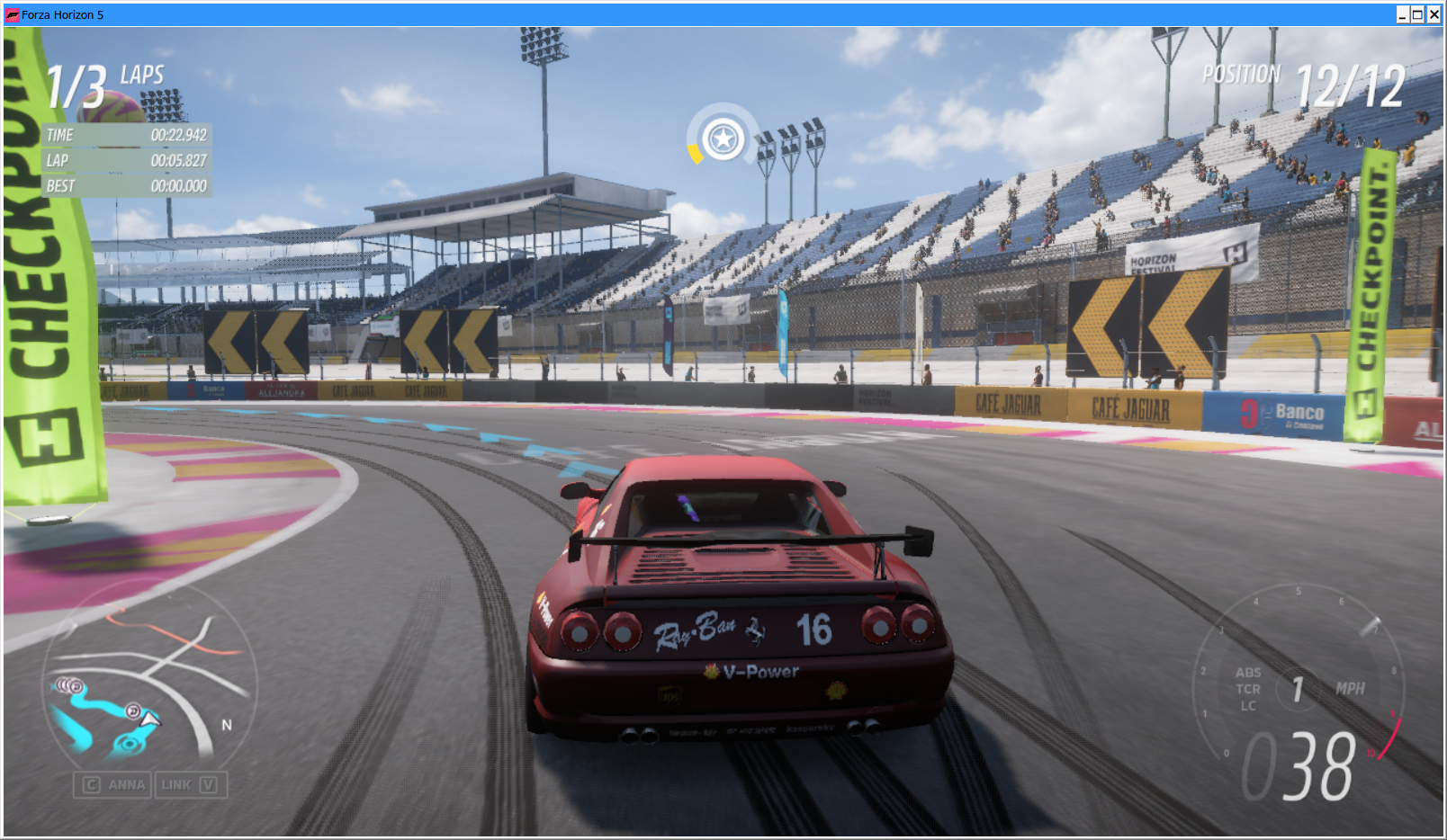}
    \caption{Screenshot of Forza Horizon 5}
    \label{fig:Horizon game title}
\end{wrapfigure}

\textbf{Game Environments.} Forza Horizon 5~\cite{horizon38} serves as a high-fidelity benchmark for continuous control, offering photorealistic 3D visuals and complex vehicle dynamics. The evaluation is conducted within a closed stadium circuit, a confined environment characterized by strict spatial boundaries and distinct track geometry. As a 'Real-Time Linear' task, it rigorously tests visual-motor coordination, requiring the agent to interpret spatial depth and visual driving lines to generate precise steering and throttle commands in real-time. Crucially, this environment assesses failure recovery robustness; beyond merely maintaining high-speed trajectories, the agent must correctly identify collision states (e.g., wall impacts) and execute autonomous rescue maneuvers to regain control.

\noindent\textbf{(1) Game state.} Real-time, High-Fidelity 3D environment.

\noindent\textbf{(2) Main GUI action space.} Keyboard Input (Continuous control for steering and throttle).

\noindent\textbf{(3) Evaluation task} Complete the stadium circuit race as quickly as possible while avoiding crashes. We impose a maximum limit of 3 minutes per episode. The agent’s performance is quantified by the total number of valid checkpoints traversed within a fixed duration of 3 minutes. This time window is calibrated to the approximate duration required for a baseline novice agent to complete three full laps. The circuit comprises 10 checkpoints per lap (including the finish line), establishing a maximum reference score ($C_{max}$) of 30 checkpoints. The final metric is derived by averaging the number of checkpoints passed across $N$ trials and normalizing this value against the maximum reference. The normalized score is defined as follows:

$$S_{norm} = \frac{\frac{1}{N}\sum_{i=1}^N C_{i}}{C_{max}} \times 100, \quad \text{where } C_{max} = 30 \text{ and } C_{i} \text{ is the checkpoints passed in trial } i.$$

\noindent\textbf{(4) Expert video content} The expert video is a skill tutorial for Forza Horizon, specifically illustrating the correlation between vehicle speed and turning radius based on centripetal force. It provides a theoretical and practical analysis of the 'out-in-out' cornering technique, explaining how maximizing the turning radius allows for higher cornering velocities. Furthermore, the tutorial elucidates the limitations of tire traction, emphasizing the necessity of managing longitudinal braking force and lateral steering grip to maintain vehicle stability.

\subsubsection*{D.12.2 Game Prompt For Forza Horizon 5}

\begin{figure}[p]
    \centering
        \includegraphics[height=0.95\textheight]{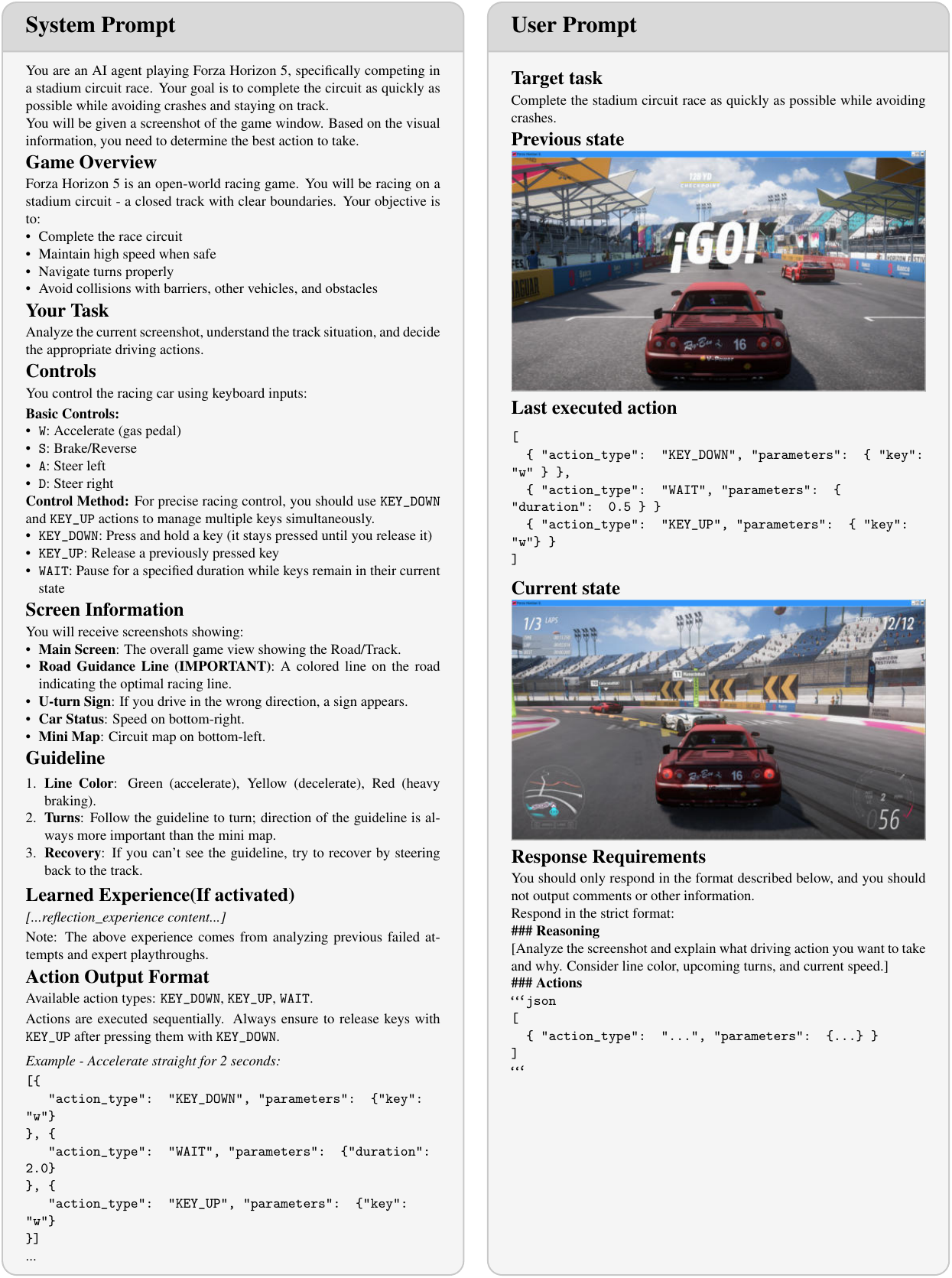}
    \caption{Horizon prompt}
    \label{fig:horizon prompt}
\end{figure}

We provide the full structure of our prompts in Figure~\ref{fig:horizon prompt}.

\newpage

\subsubsection*{D.12.3 Detailed Analysis for Forza Horizon 5}

\begin{wraptable}{r}{0.55\textwidth}
    \vspace{-10pt} 
    \centering
   
    \resizebox{0.45\textwidth}{!}{
        \begin{tabular}{lcc}
            \toprule
            \textbf{Model} & \textbf{GUI } & \textbf{GUI VR.} \\
            \midrule
            Gemini-2.5-Flash & $1.333 \pm 0.577$ & $1.333 \pm 0.58$ \\
            Gemini-2.5-Pro & $1.0 \pm 0.0$ & $1.25 \pm 0.5$ \\
            Qwen3-VL-8B & $1.0 \pm 0.0$ & $1.0 \pm 0.0$ \\
            Qwen3-VL-32B & $1.0 \pm 0.0$ & $1.0 \pm 0.0$ \\
            GPT-4o & $0.75 \pm 0.5$ & $1.0 \pm 0.0$ \\
            GPT-4o-Mini & $1.0 \pm 0.0$ & $1.0 \pm 0.0$ \\
            Seed-1.8 & $1.33 \pm 0.58$ & $0.666 \pm 0.577$ \\
            \bottomrule
        \end{tabular}%
    }
    \vspace{10pt} 
     \caption{Raw Score of Forza Horizon 5}
    \label{tab:Horizon 5 performance}
\end{wraptable}

As detailed in Table~\ref{tab:Horizon 5 performance}, the overall performance across all evaluated models is critically low, bordering on catastrophic failure for this specific domain. The results indicate that current general-purpose Vision-Language Models remain ill-equipped for continuous real-time games that necessitate millisecond-level reaction speed and muscle-memory-like execution. 

The primary bottleneck is the latency mismatch: the inference time of general models significantly exceeds the update frequency of the game environment. Consequently, models are unable to adjust the game state in real-time through rapid, iterative action sequences. This manifests as consistent failures during cornering maneuvers, where models inevitably crash into walls and struggle to navigate back to the track.

\begin{figure}[ht]
    \centering
        \includegraphics[width=\linewidth]{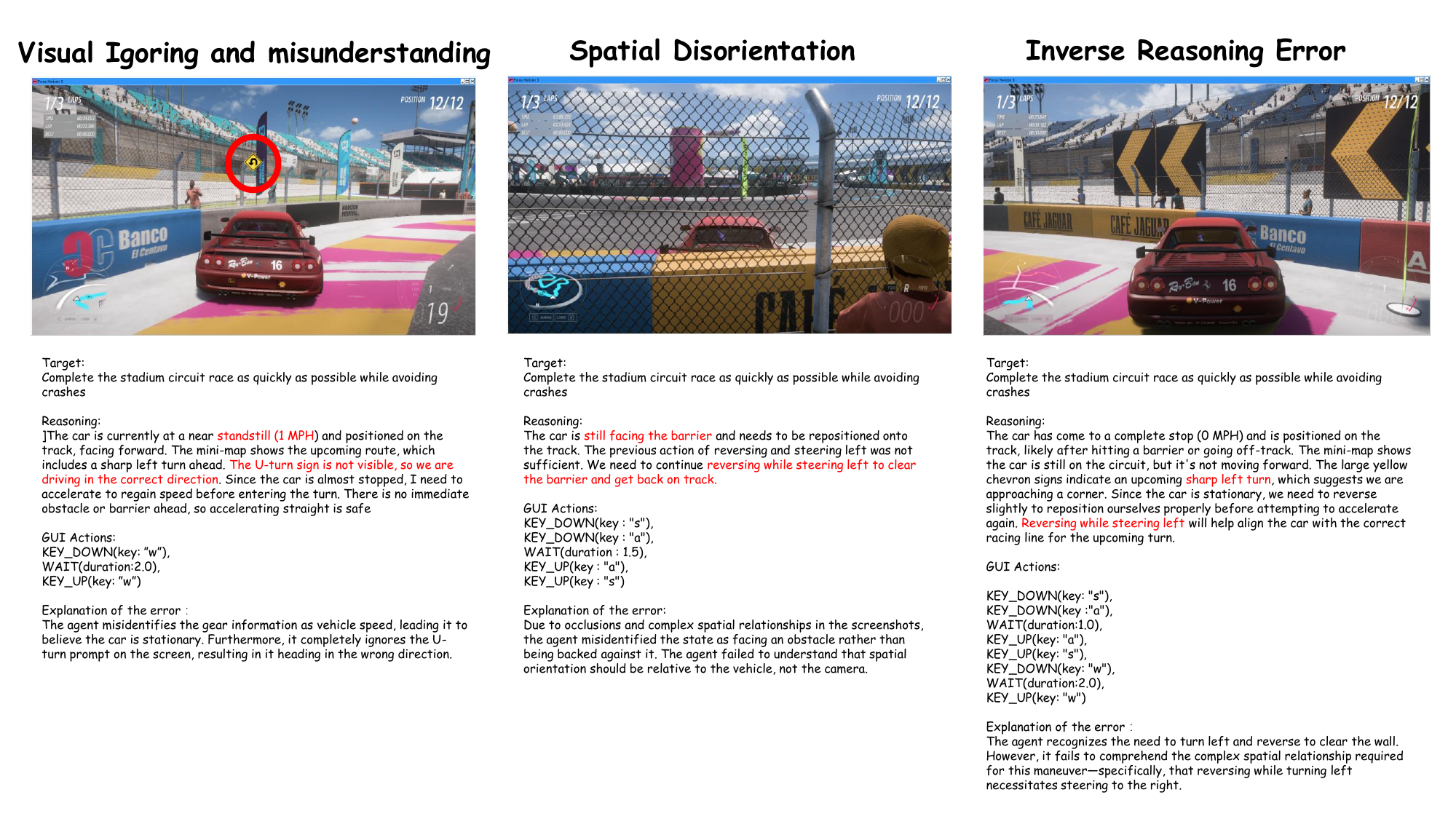}
    \caption{Horizon typical errors}
    \label{fig:Horizon errors}
\end{figure}

\textbf{Spatial Understanding and Visual Perception} \\
Beyond latency constraints, our qualitative analysis reveals significant deficiencies in the models' comprehension of the 3D environment. In high-fidelity 3D scenes, the ability of models to interpret the game state purely from visual inputs degrades substantially.

We identified three primary categories of spatial errors (\cref{fig:Horizon errors}): Visual Information Neglect, where models frequently overlook critical GUI or track information (Left); Spatial Misalignment, characterized by a marked inability to understand relative spatial positions within the 3D depth field (Middle); and Physics and Dynamics Hallucination, in which models lack a grounding in vehicle physics, particularly regarding steering relationships, as evidenced by their inability to determine the correct steering direction during reversing maneuvers (Right).

These deficits in spatial reasoning directly lead to a low success rate in simple recovery tasks, such as returning to the track after a collision, further compounding the poor performance scores.

\textbf{Video Reflection Analysis}\\
Despite the integration of a reflection module, models failed to achieve performance gains. Our analysis suggests that while models can often correctly capture high-level instructional information from teaching materials, they fail to translate this theoretical knowledge (e.g., cornering speed adjustments and trajectory control) into actionable policy updates.

The efficacy of reflection is further hampered by the low frequency of state updates and the fragmentation of failure videos, which prevents the models from accurately localizing the root cause of failure. In conclusion, current models cannot effectively leverage reflection to improve in complex, real-time gaming environments, indicating a need for comprehensive capability enhancements. 

\subsection{Mini Metro}

\subsubsection*{D.13.1 Game Description for Mini Metro}

\begin{wrapfigure}{r}{0.55\textwidth}
\vspace{-10pt}
    \centering
    \includegraphics[width=\linewidth]{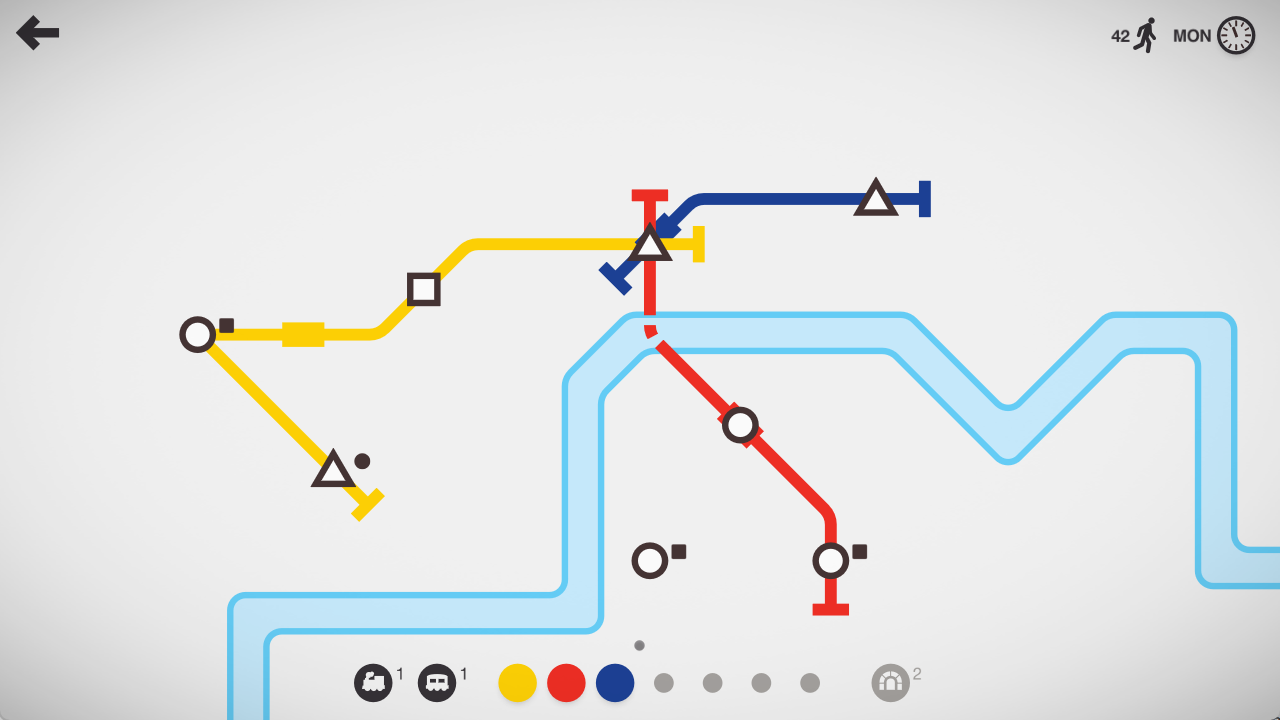}
    \caption{Screenshot of Mini Metro}
    \label{fig:mini metro game title}
\end{wrapfigure}

\textbf{Game Environments.} Mini Metro~\cite{minimetro37} is a dynamic strategy simulation that tasks agents with constructing an efficient subway network on a continuously expanding map.  The environment features stochastic node generation and strict temporal constraints, necessitating rapid, real-time topological optimization to mitigate station overcrowding. Unlike static planning tasks, this game demands fine-grained visual discrimination; the VLM must accurately identify diverse geometric station types and passenger demands within a complex, visually overlapping graph. Success relies on the agent's ability to balance immediate local congestion with global network efficiency. Thus, Mini Metro serves as a rigorous testbed for evaluating a model's capacity to integrate high-speed visual perception with adaptive, long-horizon spatiotemporal reasoning.

\noindent\textbf{(1) Game state.} Abstract, evolving 2D graph.

\noindent\textbf{(2) Main GUI action space.} Mouse Move, Click and Drag (Drawing lines to connect nodes).

\noindent\textbf{(3) Evaluation task} Design an efficient subway network to transport more passengers. The raw metric is defined as the total number of passengers successfully delivered before an episode terminates, when any station exceeds its passenger capacity. We impose a maximum limit of 100 steps per episode. The score is calculated as the mean number of transported passengers over $n$ episodes, normalized by a constant of 500. Formally, it can be defined as follows: 

\begin{equation*}
S_{norm} = \frac{\sum_{i=1}^np_{i}}{500n} \times100,\quad \text{where } p_i \text{ represents the passenger throughput of episode $i$ }
\end{equation*}

\noindent\textbf{(4) Expert video content} The expert video is a skill tutorial, presenting an integrated framework for both topological network design and dynamic resource allocation. It explicitly demonstrates optimal routing heuristics, such as the superior efficiency of loop structures over linear paths and the necessity of avoiding acute angles to mitigate transit deceleration. Parallel to these geometric considerations, the footage illustrates the critical management of finite assets—including locomotives and tunnels—to balance system capacity against stochastic passenger demand. 

\subsubsection*{D.13.2 Game Prompt for Mini Metro}

\begin{figure}[p]
    \centering
        \includegraphics[height=0.95\textheight]{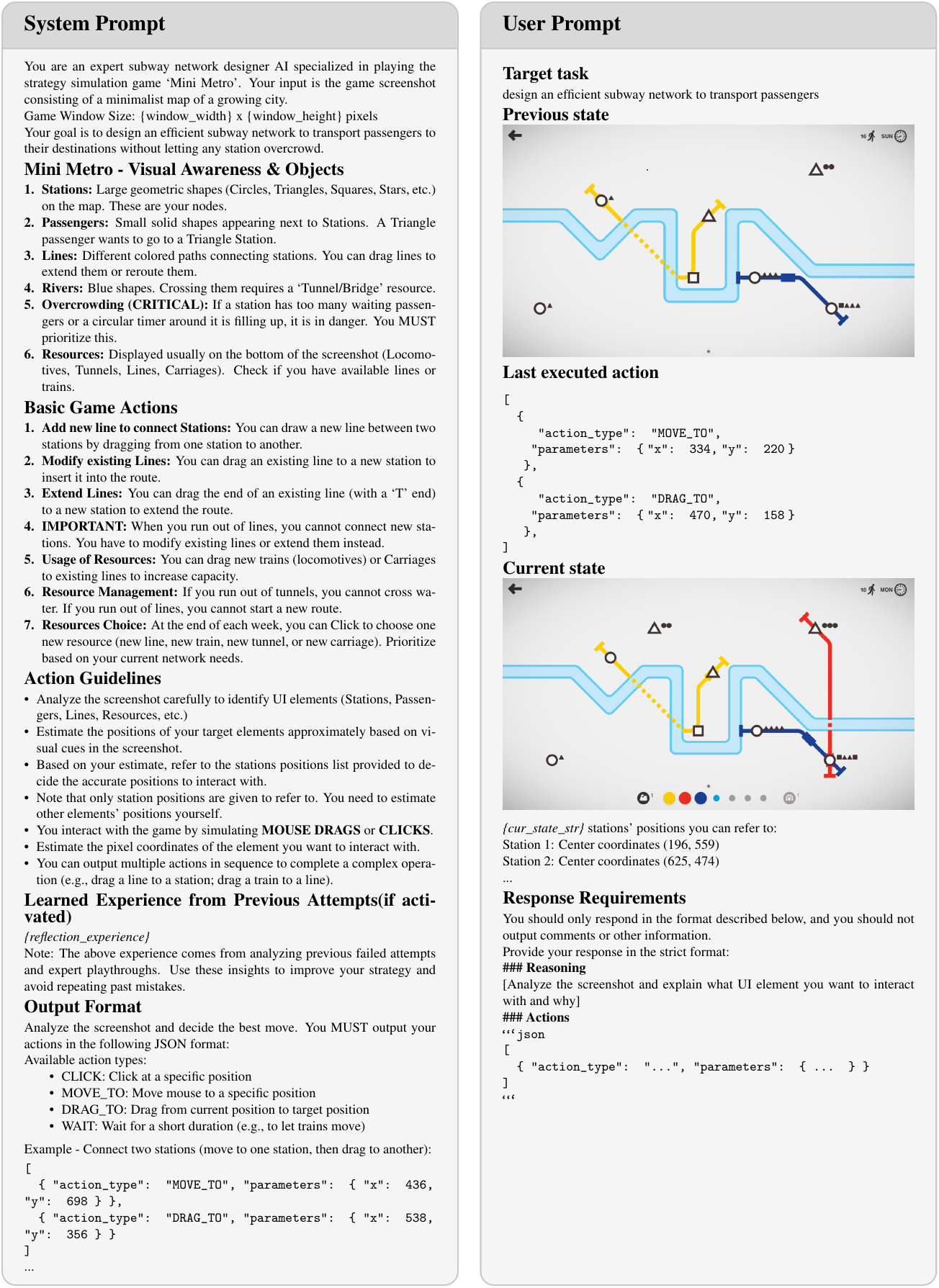}
    \caption{Mini Metro prompt}
    \label{fig:metro prompt}
\end{figure}

Due to the limited precision of VLM's native GUI grounding for the fine-grained operations required in \textit{Mini Metro}, we utilized OpenCV contour detection to provide auxiliary coordinate information in the \texttt{\{cur\_state\_str\}} placeholder. However, the model is still required to correctly map the spatial targets in the screenshot to these provided auxiliary coordinates.
We provide the full structure of our prompts in Figure~\ref{fig:metro prompt}.

\subsubsection*{D.13.3 Detailed Analysis for Metro}

\begin{wraptable}{r}{0.5\textwidth}
    \centering
    \resizebox{\linewidth}{!}{
        \begin{tabular}{lcc}
            \toprule
            \textbf{Model} & \textbf{GUI} & \textbf{GUI VR.} \\
            \midrule
            Gemini-2.5-Flash & $74.5 \pm 19.29$ & $78.0 \pm 46.4$ \\
            Gemini-2.5-Pro & $57.4 \pm 15.09$ & $87.2 \pm 40.62$ \\
            Qwen3-VL-8B & $39.33 \pm 23.51$ & $41.25 \pm 20.09$ \\
            Qwen3-VL-32B & $26.5 \pm 19.15$ & $52.8 \pm 22.28$ \\
            GPT-4o & $58.0 \pm 11.42$ & $68.0 \pm 13.78$ \\
            GPT-4o-mini & $23.0 \pm 15.6$ & $34.0 \pm 12.20$ \\
            Seed-1.8 & $5.0 \pm 9.61$ & $3.42 \pm 6.10$ \\
            \bottomrule
        \end{tabular}
    }
    \caption{Raw Performance on Mini Metro}
    \label{tab:mini_metro_results}
\end{wraptable}

\noindent \textbf{Performance Overview.} \\
The Mini Metro task, which demands a synergy of logical reasoning and real-time responsiveness, remains a significant challenge for current general Vision-Language Models (VLMs). As detailed in Table~\ref{tab:mini_metro_results}, varying bottlenecks were observed across different architectures. Seed-1.8 failed to progress effectively due to excessive single-step inference latency, while GPT-4o-mini yielded lower scores primarily due to weaker reasoning capabilities. Notably, more powerful reasoning models such as Qwen3-VL-32B and Gemini-2.5-Pro were outperformed by their lighter counterparts (Qwen3-VL-8B and Gemini-2.5-Flash) in the standard setting. This inversion indicates that for real-time decision-making tasks, the latency constraints of large-scale reasoning models can negate their logical superiority. Consequently, Mini Metro serves as an effective benchmark for evaluating the trade-off between inference speed and reasoning depth. It is worth noting that nearly all models demonstrated improvement after incorporating video reflection, suggesting that heuristic information derived from visual experience effectively aids model reasoning.

\begin{figure}[h]
    \centering
        \includegraphics[width=\linewidth]{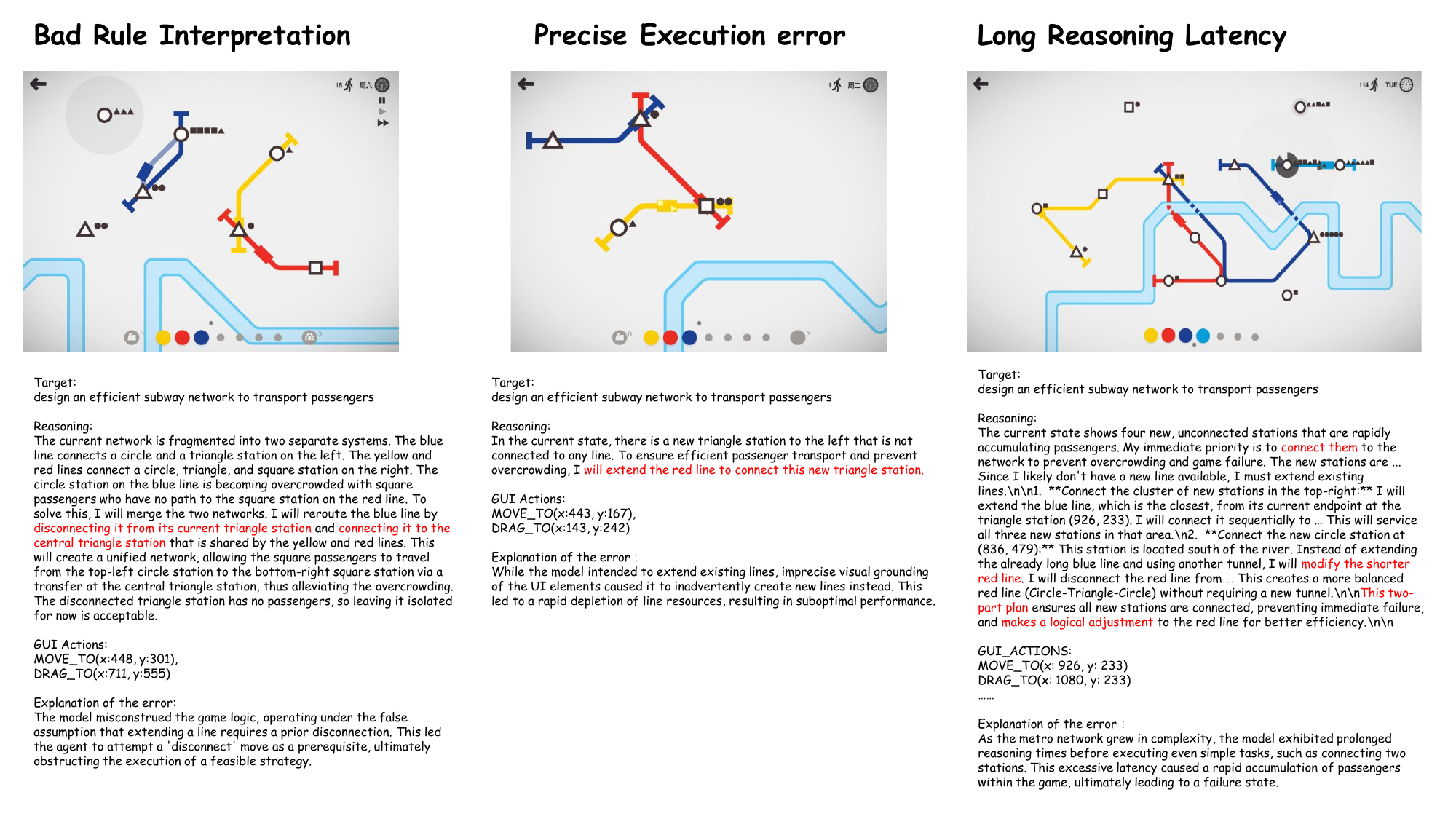}
    \caption{mini metro typical errors}
    \label{fig:Metro Errors}
\end{figure}

\noindent \textbf{Error Analysis.} \\
Three primary error categories were identified (\cref{fig:Metro Errors}). The fundamental reason VLMs underperform compared to humans lies in the execution phase following reasoning. Models struggle to precisely localize minute UI elements such as stations, lines, and extension nodes(\cref{fig:Metro Errors} middle), leading to a frequent "reasoning-execution gap." Even when heuristic prompts explicitly provided station coordinates, execution failures persisted and are often fatal in a real-time environment. Furthermore, current models lack the "active attention" mechanism characteristic of human perception. When facing complex topological networks, models lack the ability to selectively focus only on changing components, resulting in unexpectedly long inference times during the later stages of the game(\cref{fig:Metro Errors} right). Additionally, models occasionally exhibit hallucinations regarding rule interpretation, which unnecessarily inflates reasoning time and directly results in execution errors(\cref{fig:Metro Errors} left)

\noindent \textbf{Video Reflection Analysis} \\
The analysis of reflection results highlights a divergence in learning capabilities based on model scale. For smaller, high-speed models like Qwen3-VL-8B and Gemini-2.5-Flash, video reflection provided limited gains, indicating a deficiency in their ability to reflect and learn from visual history. In contrast, larger models such as Gemini-2.5-Pro and Qwen3-VL-32B successfully extracted critical heuristic experiences from the video data. These learned heuristics significantly reduced inference overhead, thereby enabling substantial performance improvements despite their inherent latency.

\subsection{Genshin Impact}
\subsubsection*{D.14.1 Game Description for Genshin Impact}

\begin{wrapfigure}{r}{0.55\textwidth}
\vspace{-20pt}
    \centering
    \includegraphics[width=\linewidth]{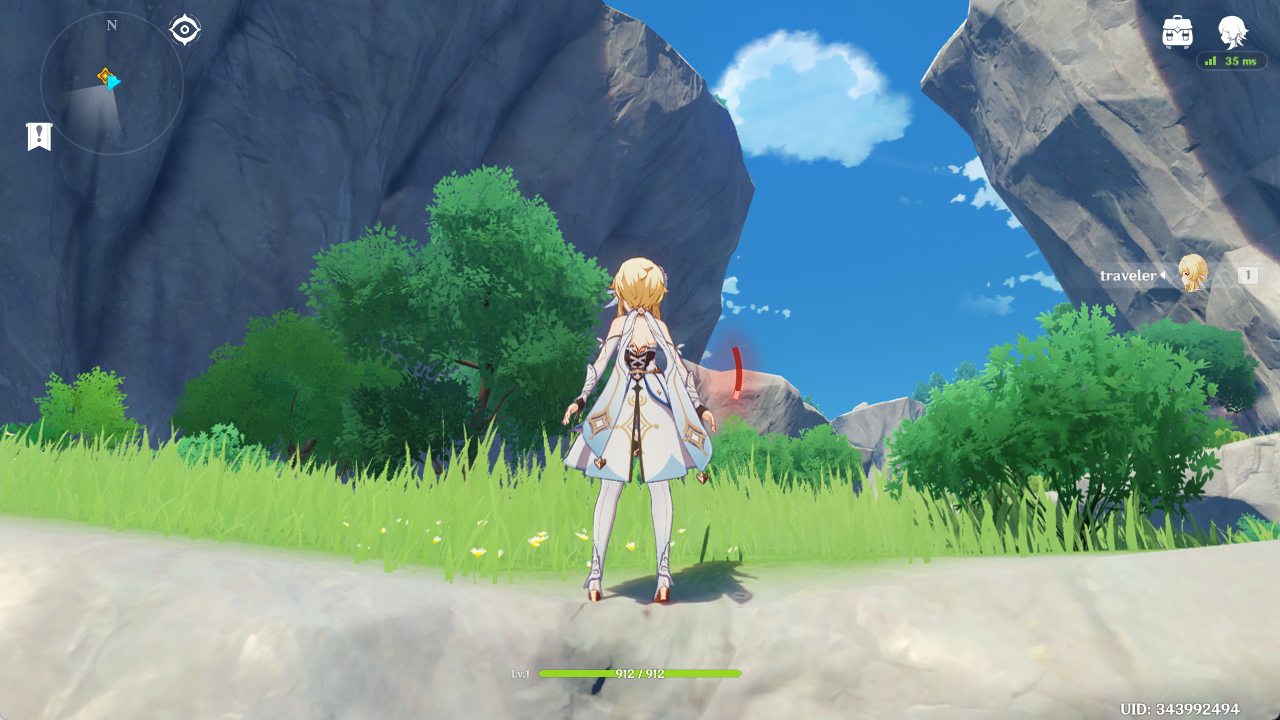}
    \caption{Screenshot of Genshin Impact }
    \label{fig:Genshin game title}
\end{wrapfigure}

\textbf{Game Environments.} 
Genshin Impact~\cite{Genshin36} is a vast open-world action RPG with an elemental combat system. The environment is rich with traversable terrain, puzzles, and enemies. The game tests embodied exploration and multi-tasking. Agents are evaluated on their ability to navigate complex 3D topography, understand elemental interaction mechanics, and follow multi-step quest instructions in a non-linear world. As a "Real-Time Non-Linear" benchmark, it represents one of the most difficult challenges, requiring the synthesis of exploration, combat, and puzzle-solving. The agent must process high-fidelity 3D visuals to identify traversable paths and enemies while managing real-time combat cooldowns and elemental combinations, mimicking a human player's holistic engagement.

\noindent\textbf{(1) Game state.} Non-Linear (Open-Ended), High-Fidelity 3D world.

\noindent\textbf{(2) Main GUI action space.} Keyboard and Mouse (Movement, camera control, combat skills).

\noindent\textbf{(3) Evaluation task.} Finish the prologue. We impose a maximum limit of 100 steps per episode.The agent's performance is quantified by the average number of completed milestones. The milestones are generated from the expert video. To obtain a standardized score, the number of completed milestones is normalized by the total number of milestones, and the calculation formula is as follows:

\begin{equation*}
\text{Score} = \frac{N_{\text{completed}}}{N_{\text{total}}}\times 100
\end{equation*}

where \( N_{\text{completed}} \) denotes the number of milestones completed by the agent, and \( N_{\text{total}} \) denotes the total number of milestones (7 in total, as listed in Table~\ref{tab:genshin_milestone}). The normalized score ranges from 0 to 100.

\begin{table}[h]
  \centering
  \small 
  \begin{tabular}{p{\textwidth}}
    \toprule
    \textbf{Milestones} \\
    \midrule
    \textbf{1. Basic Operation:} Follow Paimon's instructions to learn the basic operations. \\
    \textbf{2. Unlock The Statue of The Seven:} Interact with the first Statue of The Seven at Starfell Lake. \\
    \textbf{3. First Combat Encounter:} The player encounters an enemy and completes the combat tutorial using the newly acquired skills. \\
    \textbf{4. Encounter with Dvalin:} The player enters the Whispering Woods and triggers the plot where the dragon Dvalin interacts with the bard. \\
    \textbf{5. Meet Amber:} The player encounters Amber, a scout knight of the Knights of Favonius, on the way to Mondstadt. \\
    \textbf{6. Clear the Hilichurl camp:} The player clears the nearby Hilichurl camp to proceed further. \\
    \textbf{7. Arrive at Mondstadt:} The player crosses the bridge and enters the gate of Mondstadt, marking the end of the prologue. \\
    \bottomrule
  \end{tabular}
\caption{Milestones of Genshin Impact}
\label{tab:genshin_milestone}
\end{table}

\noindent\textbf{(4) Expert video content.} The expert video is a demonstration video. It demonstrates exploring an open world, collecting resources, navigating to mission locations using maps and mini-map, triggering story events and dialogues, and utilizing skills and teamwork to complete combat.

\subsubsection*{D.14.2 Game Prompt for Genshin Impact}
\begin{figure}[p]
    \centering
        \includegraphics[height=0.8\textheight]{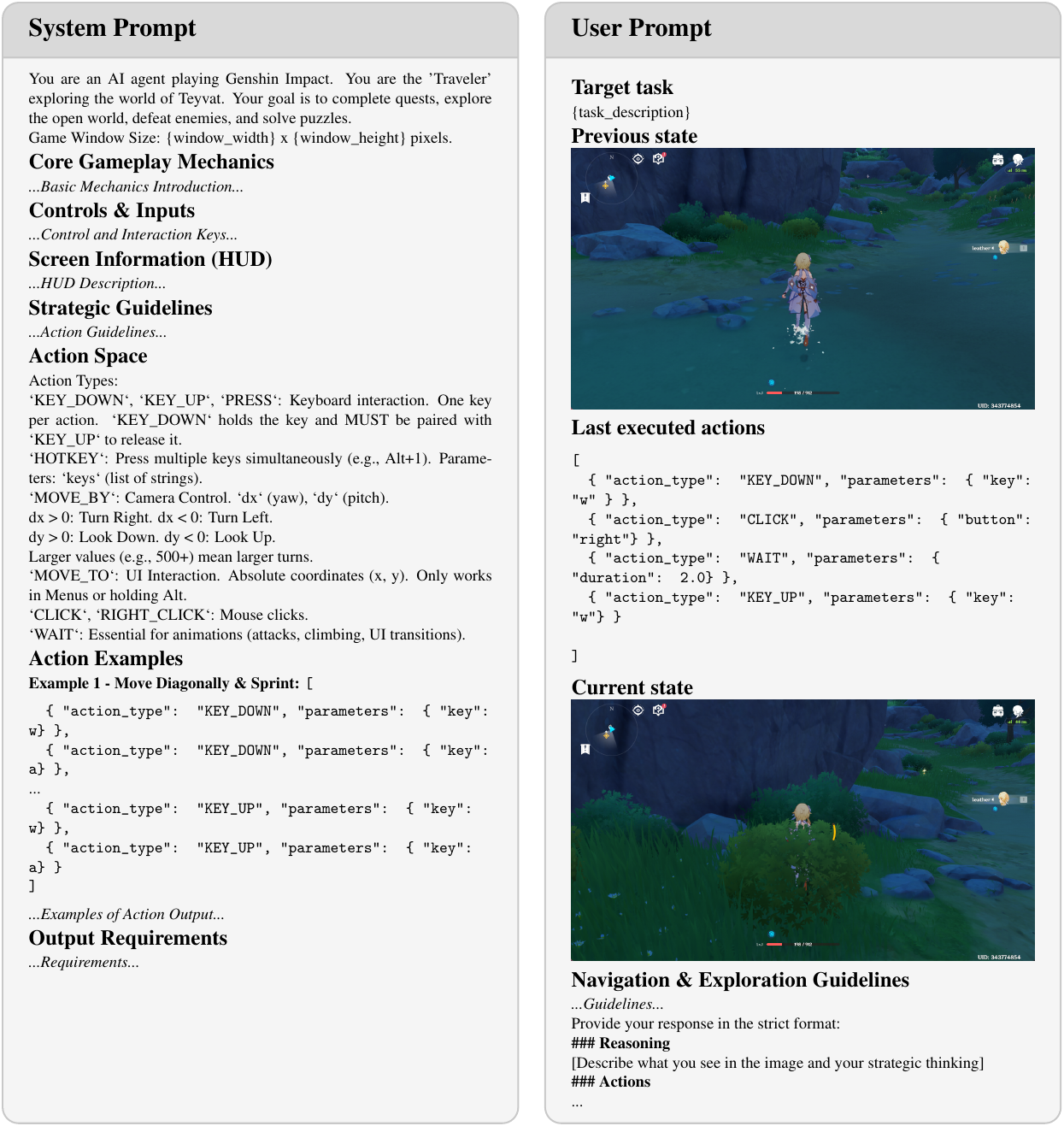}
    \caption{Genshin Impact zero-shot prompt}
    \label{fig:genshin_zeroshot_prompt}
\end{figure}

\begin{figure}[p]
    \centering
        \includegraphics[height=0.8\textheight]{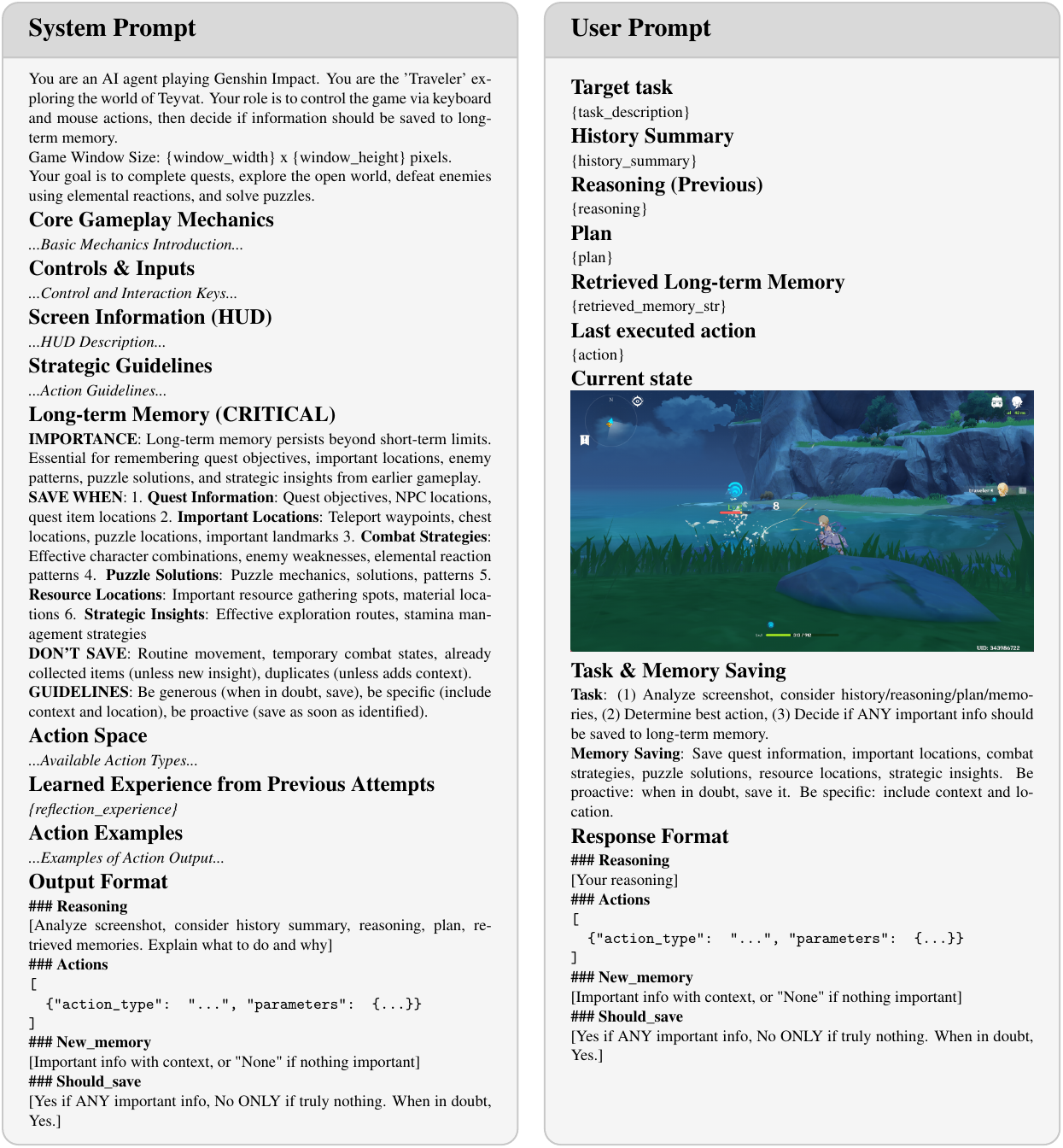}
    \caption{Genshin Impact memory agent prompt}
    \label{fig:genshin_memory_prompt}
\end{figure}

We provide the full structure of our prompts in Figure~\ref{fig:genshin_zeroshot_prompt} and Figure~\ref{fig:genshin_memory_prompt}.

\subsubsection*{D.14.3 Detailed Analysis for Genshin Impact}

The experimental results in \textit{Genshin Impact}, as presented in Table~\ref{tab:genshin_scores}, reveal a uniform performance pattern across all evaluated general VLMs: all models achieved an identical score of $14.3 \pm 0.0$, corresponding to the completion of only the first milestone (\textit{Basic Operation}). This consistent result indicates two key insights: first, mainstream general VLMs possess a baseline capability to execute simple, explicit instructions (e.g., following Paimon’s guidance for basic controls), confirming their fundamental command-following competence. Second, the failure to progress beyond the initial milestone exposes critical deficits in their ability to operate within 3D open-world environments, including deficiencies in spatial distance perception, dynamic environmental awareness, and continuous action execution.

\begin{figure}[ht]
    \centering
        \includegraphics[width=\linewidth]{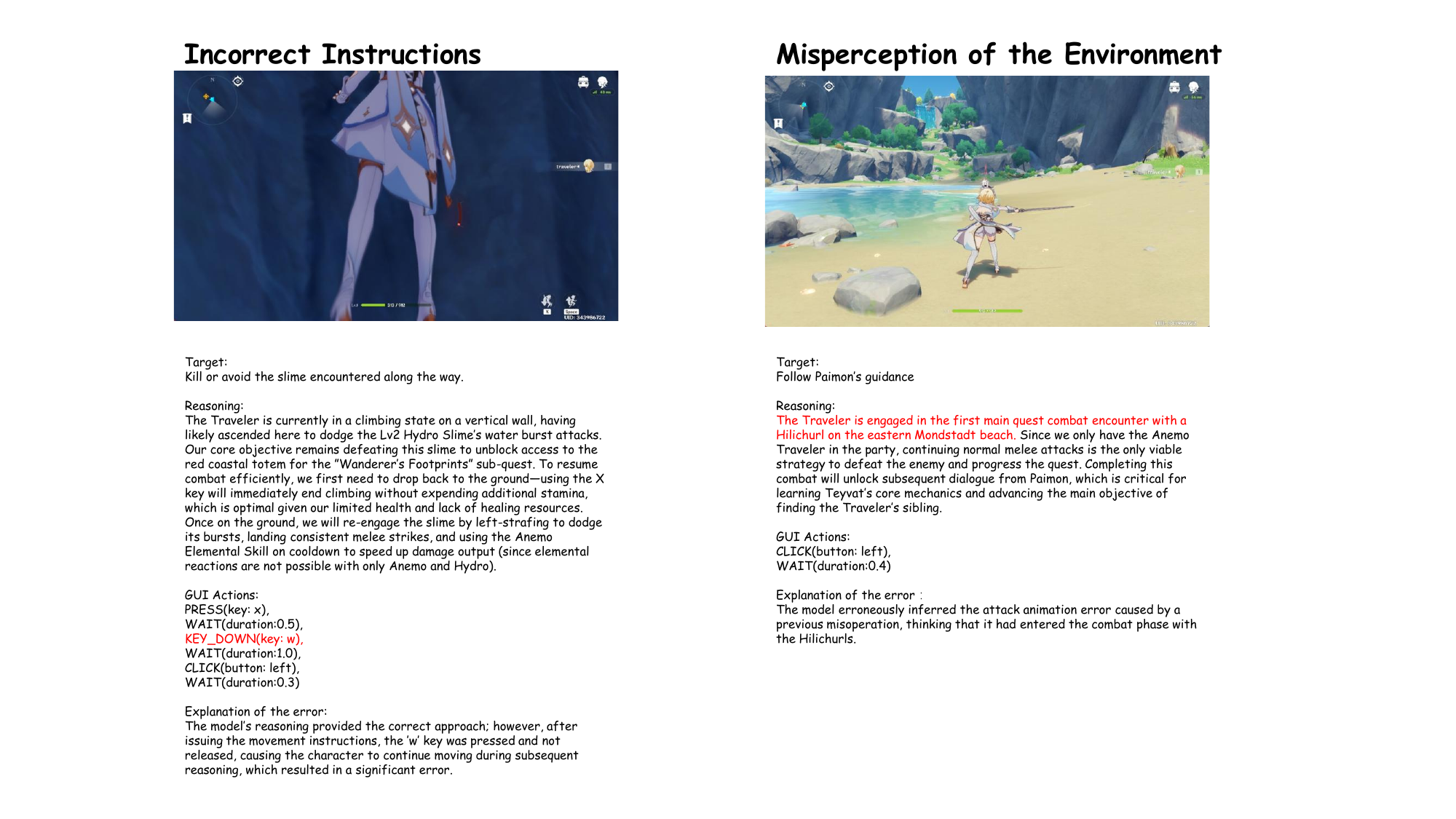}
    \caption{Genshin Impact errors}
    \label{fig:genshin_errors}
\end{figure}

Qualitative analysis of typical errors in Figure~\ref{fig:genshin_errors} further elaborates these limitations. In the left panel, the model demonstrates a failure in sustained action control: after issuing a movement instruction, it failed to release the 'W' key, causing uninterrupted movement that disrupted subsequent reasoning and task progression. This error highlights the inability of general VLMs to maintain precise, context-aware control over continuous in-game actions—a critical requirement for navigating 3D spaces. In the right panel, the model exhibits erroneous environmental perception: it misinterpreted attack animations triggered by prior incorrect operations as initiation of combat with Hilichurls, despite the absence of enemies in the scene. This hallucination reflects a profound weakness in dynamic 3D scene understanding, where the model cannot reliably distinguish between transient visual artifacts and actual environmental events, leading to false strategic judgments.

Collectively, these findings demonstrate that while current general VLMs can handle isolated, 2D-style GUI interactions or simple discrete commands, they lack the robust spatial intelligence and dynamic reasoning capabilities required for 3D open-world tasks. The uniform performance across all models and configurations underscores a systemic limitation: existing general VLMs are not yet equipped to perceive, reason about, and act consistently within the complex, evolving spatial contexts of games like \textit{Genshin Impact}.

\begin{table}[htbp]
    \centering
    \small
    \begin{tabular}{lcccc}
        \toprule 
        \textbf{Model} & \textbf{Zero-shot} & \textbf{Memory} & \textbf{Zero-shot + VR.} & \textbf{Memory + VR.} \\
        \midrule 
        Gemini-2.5-Flash & $14.3 \pm 0.0$ & $14.3 \pm 0.0$ & $14.3 \pm 0.0$ & $14.3 \pm 0.0$ \\

        Gemini-2.5-Pro & $14.3 \pm 0.0$ & $14.3 \pm 0.0$ & $14.3 \pm 0.0$ & $14.3 \pm 0.0$ \\

        Qwen3-VL-8B & $14.3 \pm 0.0$ & $14.3 \pm 0.0$ & $14.3 \pm 0.0$ & $14.3 \pm 0.0$ \\

        Qwen3-VL-32B & $14.3 \pm 0.0$ & $14.3 \pm 0.0$ & $14.3 \pm 0.0$ & $14.3 \pm 0.0$ \\

        GPT-4o & $14.3 \pm 0.0$ & $14.3 \pm 0.0$ & $14.3 \pm 0.0$ & $14.3 \pm 0.0$ \\

        GPT-4o-mini & $14.3 \pm 0.0$ & $14.3 \pm 0.0$ & $14.3 \pm 0.0$ & $14.3 \pm 0.0$ \\

        Seed-1.8 & $14.3 \pm 0.0$ & $14.3 \pm 0.0$ & $14.3 \pm 0.0$ & $14.3 \pm 0.0$ \\
        \bottomrule 
    \end{tabular}
    \caption{Scores in Genshin Impact (Mean $\pm$ Std)}
    \label{tab:genshin_scores}
\end{table}

\newpage

\subsection{Red Dead Redemption 2}
\subsubsection*{D.15.1 Game Description for Red Dead Redemption 2}

\begin{wrapfigure}{r}{0.55\textwidth}
    \centering
    \includegraphics[width=\linewidth]{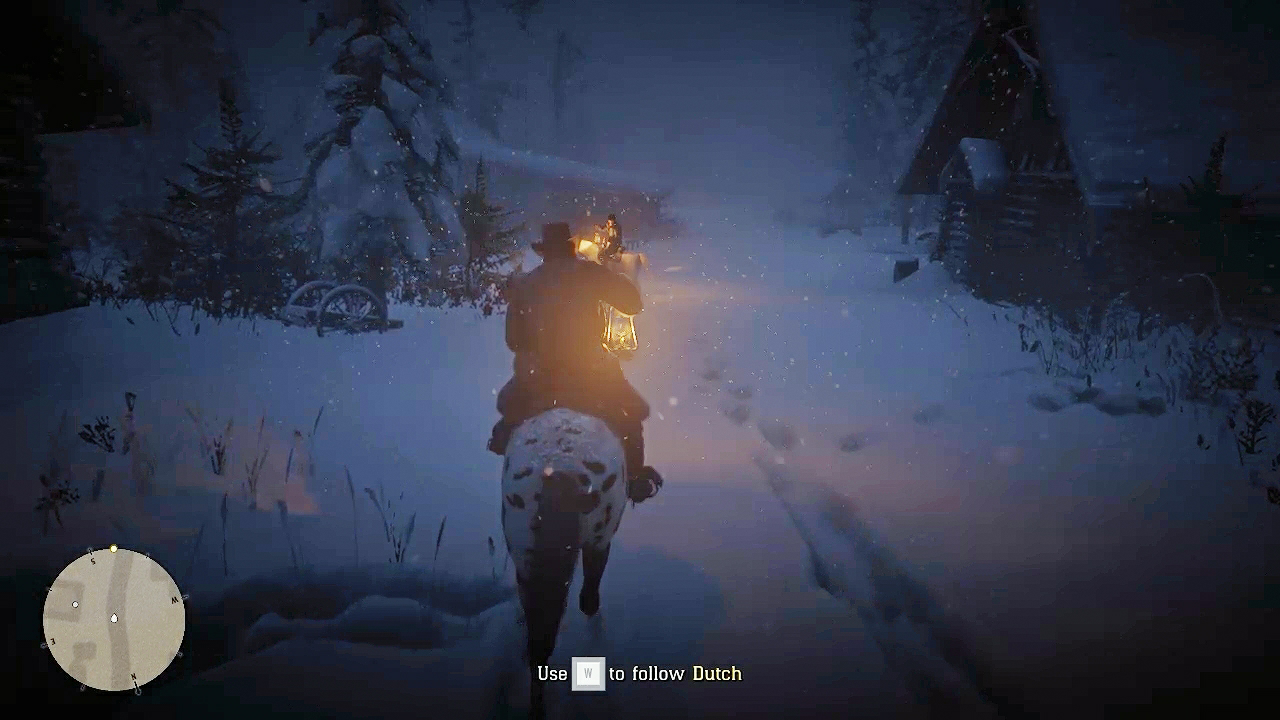}
    \caption{Screenshot of Red Dead Redemption 2}
    \label{fig:rdr2}
\end{wrapfigure}
\textbf{Game Environments.} 
Red Dead Redemption 2~\cite{RDR235}is a highly realistic open-world simulation, known for its narrative depth and environmental realism. The setting is a detailed recreation of the American frontier. This game tests comprehensive general intelligence, including social interactions, realistic physics navigation, and narrative comprehension. 

Agents must operate within a living world where characters and animals react realistically, requiring a high degree of common sense and contextual understanding. It stands as the "Hard" tier of the "Real-Time Non-Linear" category, challenging the agent to handle partially observable, long-horizon tasks that demand not just game mechanics mastery, but an understanding of social cues and realistic environmental constraints.

\noindent\textbf{(1) Game state.} 3D open-world environment with realistic physics, dynamic scenes, and interactive NPCs.

\noindent\textbf{(2) Main GUI action space.} Keyboard and Mouse (Complex interactions, movement, aiming).

\noindent\textbf{(3) Evaluation task.} Complete the milestones of Chapter~1 in \textit{Red Dead Redemption~2}. We impose a maximum limit of 200 steps per episode. The agent’s performance is quantified by the number of successfully completed milestones within Chapter~1. The total set of milestones is extracted from expert demonstration videos. The evaluation score is normalized by the total number of Chapter~1 milestones. The Chapter~1 milestones, in chronological order, are: \textit{First Control: The Journey Begins}, \textit{Arrival at Adler Ranch}, \textit{First Gunfight}, \textit{Looting Supplies}, \textit{Barn Ambush and Interrogation}, \textit{First Encounter with Sadie Adler}, \textit{Mission Complete: Outlaws from the West}, \textit{Mission Start: Enter, Pursued by a Memory}, \textit{Wolf Encounter}, and \textit{Finding John Marston}. The normalized score is defined as follows:
\begin{equation*}
S_{norm} = \frac{N_{agent}}{N_{total}} \times 100,
\end{equation*}
where $S_{agent}$ denotes the number of milestones completed by the agent during evaluation, and $S_{total}$ denotes the total number of milestones in Chapter 1.

\noindent\textbf{(4) Expert video content.} The expert video consists of a full gameplay recording of Chapter~1 completed by an experienced player. The video demonstrates effective long-horizon decision-making in an open-world setting, including navigation, combat engagement, and context-aware interactions with NPCs and the environment. By implicitly modeling task prioritization, situational awareness, and mission-driven action selection under complex and dynamic conditions, the expert gameplay provides a reference trajectory that highlights efficient progression through chapter milestones. It guides agents to align low-level control with high-level narrative objectives and to make decisions that favor robustness and temporal coherence over short-term reactive behaviors.

\subsubsection*{D.15.2 Game Prompt For Red Dead Redemption 2}
\begin{figure}[p]
    \centering
        \includegraphics[height=0.95\textheight]{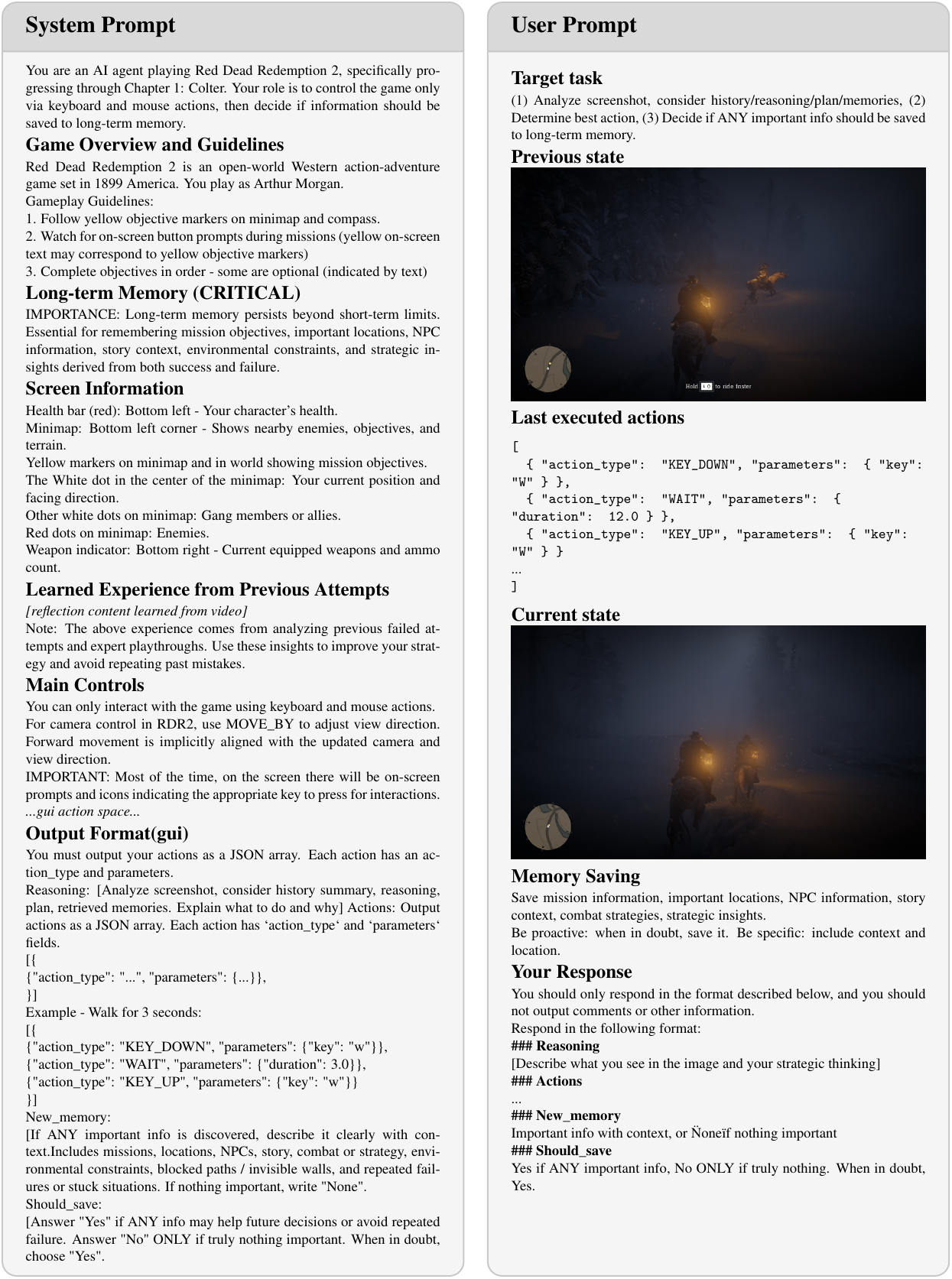}
    \caption{Red Dead Redemption 2 Prompt}
    \label{fig:red dead redemption 2 prompt}
\end{figure}

We provide the full structure of our prompts in Figure~\ref{fig:red dead redemption 2 prompt}.

\subsubsection*{D.15.3 Detailed Analysis For Red Dead Redemption 2}

\begin{wrapfigure}{r}{0.6\textwidth} 
  \centering
  \vspace{0pt} 
  \small 
  \begin{tabular}{lccc}
    \toprule
\textbf{Model} & \textbf{Zeroshot} & \textbf{Memory} & \textbf{Memory VR.} \\
    \midrule
    Gemini-2.5-Flash      & $7.0 \pm 5.0$   & $10.0 \pm 0.0$   & $13.0 \pm 5.0$ \\

    Gemini-2.5-Pro        & $10.0 \pm 0.0$   & $10.0 \pm 0.0$   & $10.0 \pm 0.0$ \\

    Qwen3-VL-8B           & $7.0 \pm 5.0$   & $7.0 \pm 5.0$   & $7.0 \pm 5.0$ \\

    Qwen3-VL-32B          & $10.0 \pm 8.0$   & $7.0 \pm 5.0$   & $10.0 \pm 8.0$ \\

    GPT-4o                & $7.0 \pm 5.0$   & $10.0 \pm 0.0$   & $10.0 \pm 0.0$ \\

    GPT-4o-mini           & $7.0 \pm 5.0$   & $10.0 \pm 0.0$   & $10.0 \pm 0.0$ \\

    Seed-1.8       & $7.0 \pm 5.0$   & $10.0 \pm 0.0$   & $7.0 \pm 5.0$ \\
    \bottomrule
  \end{tabular}
  \vspace{0pt} 
   \caption{Scores in Red Dead Redemption 2}
  \label{tab:red dead redemption 2 raw scores}
\end{wrapfigure}

The experimental results on Red Dead Redemption 2 (RDR2) reveal a pronounced performance ceiling that persists across model scaling and enhanced reasoning strategies. As shown in \cref{tab:red dead redemption 2 raw scores}, the transition from structured 2D interfaces to a dynamic 3D open-world environment poses a substantial challenge for current VLM-based agents. While the reflection mechanism yields marginal improvements for Gemini-2.5-Flash and Qwen3-VL-32B, most frontier models, including GPT-4o and Gemini-2.5-Pro, remain near a baseline score of $10.0$. This stagnation indicates a fundamental limitation in bridging discrete GUI execution with continuous spatial grounding. Notably, Seed-1.8 exhibits degraded performance under reflection, suggesting that iterative reasoning may introduce instability or action loops that interfere with responsiveness to real-time environmental feedback. 
\begin{figure}[h]
    \centering
        \includegraphics[width=\linewidth]{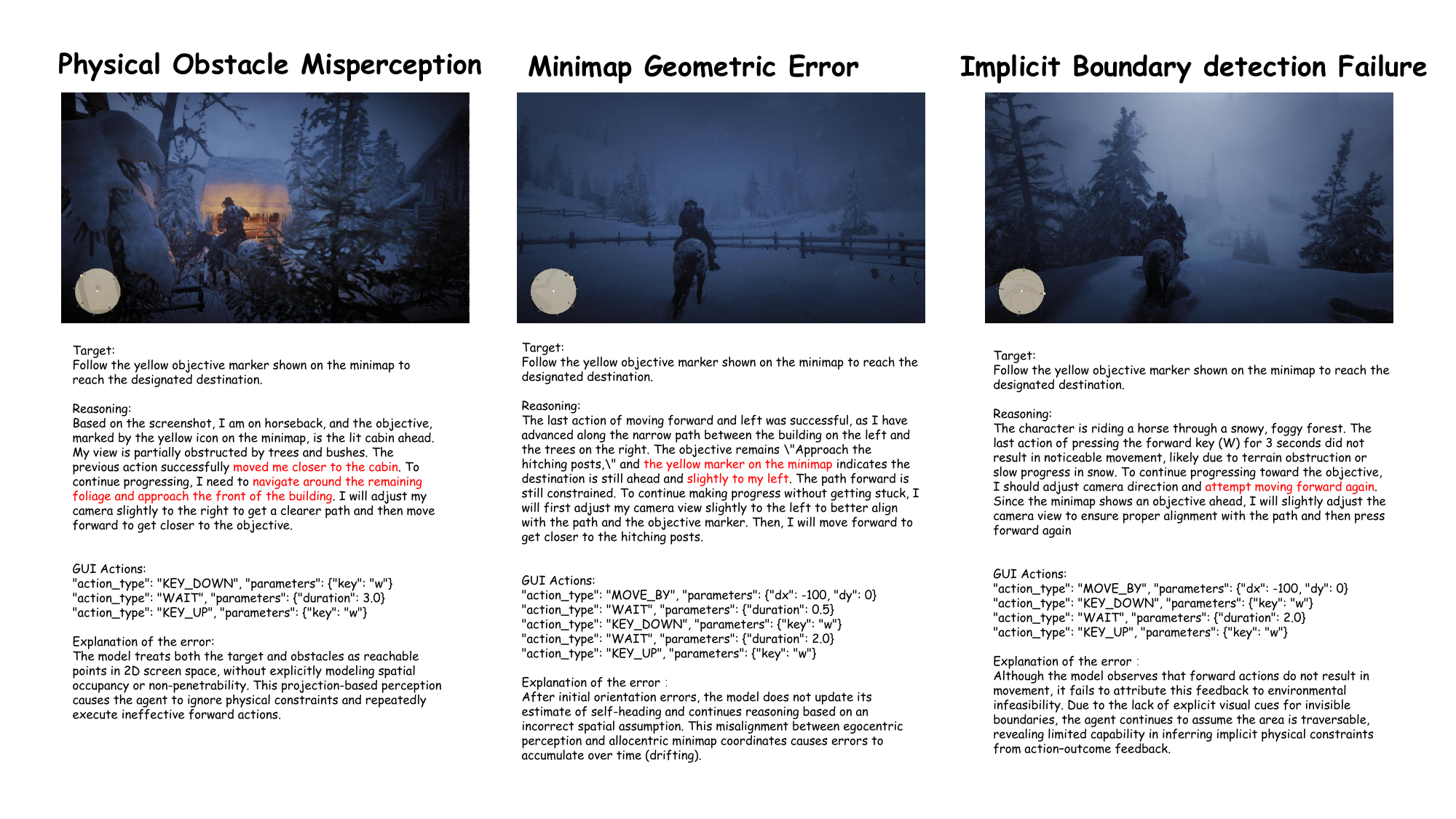}
    \caption{red dead redemption 2 errors}
    \label{fig:red dead redemption 2 typical errors}
\end{figure}

\textbf{Perception–Reasoning Gap. }\\
We observe a dominant Perception–Reasoning Gap in Red Dead Redemption 2, where agents rely on projection-based perception that treats targets and obstacles as reachable points in 2D screen space rather than as volumetric entities with physical constraints. As shown in \cref{fig:red dead redemption 2 typical errors} (left), physical obstacle misperception leads to systematic misclassification of dense foliage, terrain irregularities, and environmental structures as traversable regions, resulting in repeated forward actions despite physical infeasibility.

These perceptual limitations propagate into navigation-level reasoning errors. Although mini-map objectives are correctly identified at the semantic level, agents fail to maintain a stable transformation between allocentric mini-map coordinates and egocentric camera orientation, causing orientation drift over time (\cref{fig:red dead redemption 2 typical errors}, middle). When forward actions yield little or no displacement, agents do not attribute the outcome to environmental constraints but instead persist with forward-biased heuristics and minor camera adjustments, entering repetitive action loops without recovery behaviors (\cref{fig:red dead redemption 2 typical errors}, right). Together, these results suggest that in open-world environments with implicit physical boundaries, reasoning effectiveness is fundamentally constrained by the agent’s ability to infer 3D structure and navigability from visual and action feedback.

\textbf{Video Reflection Analysis.} 

\begin{wrapfigure}{r}{0.5\textwidth} 
    \centering
    \small
    \includegraphics[width=\linewidth]{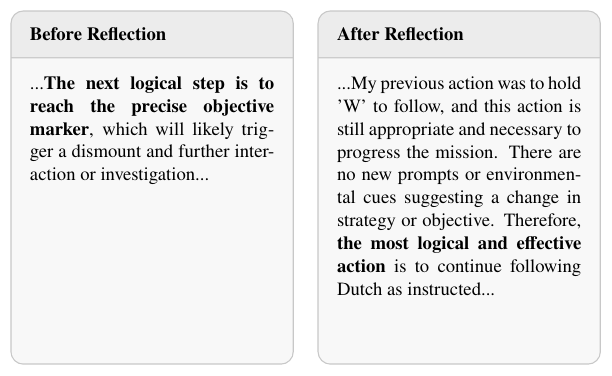}
    \caption{Gemini-2.5-Flash reflection improvement example}
    \label{fig:Gemini_2.5_flash_reflection}
\end{wrapfigure}

A notable observation in Red Dead Redemption 2 is that the effectiveness of reflection is strongly constrained by perceptual grounding rather than high-level reasoning capacity. For powerful models such as Gemini-2.5-Pro and GPT-4o, reflection yields negligible gains, as dominant failures stem from a Perception–Reasoning Gap in 3D spatial understanding. Smaller models similarly fail to benefit, with reflection often amplifying speculative reasoning that is weakly aligned with immediate control signals. In contrast, mid-tier models show limited but consistent improvements, where reflection helps suppress narrative over-interpretation and re-anchor decisions to explicit mission instructions, as evidenced by the shift from exploratory objective inference to instruction-following behavior in the before/after comparison (\cref{fig:Gemini_2.5_flash_reflection}). 

However, reflection does not correct core perceptual errors such as obstacle misclassification or mini-map camera misalignment, leading agents to persist in forward-biased action loops. Overall, these results suggest that in open-world environments with implicit physical constraints, reflection is beneficial only when perceptual grounding is sufficiently reliable; otherwise, it may reinforce misaligned reasoning rather than improve execution.


\end{document}